\pdfoutput=1
\pdftrailerid{}
\documentclass[11pt,a4paper]{article}
\usepackage[a4paper,margin=20mm]{geometry}
\usepackage[T1]{fontenc}
\usepackage[utf8]{inputenc}
\usepackage{lmodern,microtype,amsmath,amssymb,booktabs,longtable,array,graphicx}
\usepackage[table]{xcolor}
\usepackage[round]{natbib}
\usepackage{fvextra,upquote,seqsplit,xurl}
\usepackage[hypertexnames=false,colorlinks=true,linkcolor=blue!60!black,citecolor=blue!60!black,urlcolor=blue!60!black]{hyperref}
\DeclareUnicodeCharacter{00D7}{\ensuremath{\times}}
\DeclareUnicodeCharacter{2013}{--}
\DeclareUnicodeCharacter{2212}{\ensuremath{-}}
\begin{document}

% BEGIN M:1 text

\title{Multi-\allowbreak{}Estimate Elicitation Does Not Uniformly Debias Language Model Judgments}\author{An-Sheng Jhang\\Dongguan University of Technology}\date{}\maketitle

% END M:1

% BEGIN M:6 text

\section*{Abstract}

% END M:6

% BEGIN M:8 text

Multi-\allowbreak{}Candidate Verbalization (MCV) is a task-\allowbreak{} and model-\allowbreak{}dependent output protocol:\allowbreak{} repairing format does not guarantee better posterior accuracy,\allowbreak{} and requesting weights differs from reweighting an already generated candidate set. Across 31,\allowbreak{}672 planned/\allowbreak{}admitted evaluation units and 32,\allowbreak{}184 planned/\allowbreak{}admitted calls (excluding retries),\allowbreak{} we connect a frozen six-\allowbreak{}configuration cohort to exploratory scale/\allowbreak{}organization experiments and two independently preregistered post-\allowbreak{}cohort formal suites. Stage 1 retains 11 directional confirmations among 45 frozen tests; only increased post-\allowbreak{}outcome confirmation error on gpt-\allowbreak{}oss-\allowbreak{}120b exceeds its SESOI. P1 completes 1,\allowbreak{}920 terminal units:\allowbreak{} smaller response scales sharply lower operational string-\allowbreak{}match rates but increase Bayesian error. P2 completes 2,\allowbreak{}560 calls and 2,\allowbreak{}048 units:\allowbreak{} eliciting weights raises three errors by +2.75 /\allowbreak{} +2.05 /\allowbreak{} +3.88 pp,\allowbreak{} while within-\allowbreak{}generation weighting changes them by only -\allowbreak{}0.052 /\allowbreak{} -\allowbreak{}0.049 /\allowbreak{} -\allowbreak{}0.067 pp,\allowbreak{} a 42\ensuremath{\times}--58\ensuremath{\times} magnitude difference. Unconstrained joint generation with post-\allowbreak{}hoc normalization offers a local normalized-\allowbreak{}error improvement for three exhaustive options,\allowbreak{} without confirmed accuracy improvement. These results distinguish format validity,\allowbreak{} numerical accuracy and psychological susceptibility; a descriptive three-\allowbreak{}class behavioral spectrum organizes heterogeneity without identifying internal causes.\par

% END M:8

% BEGIN M:10 text

\section{Introduction}

% END M:10

% BEGIN M:12 text

Multi-\allowbreak{}Candidate Verbalization (MCV) is conditional on task structure and model configuration:\allowbreak{} format repair need not improve accuracy; asking for weights and changing the readout are different interventions; near-\allowbreak{}perfect arithmetic does not ensure immunity to contextual bias.\par

% END M:12

% BEGIN M:14 text

\textbf{Solid Claim.} P2-\allowbreak{}A independently manipulates weight elicitation:\allowbreak{} E3(M\_\allowbreak{}W) minus E2(M\_\allowbreak{}U) increases Bayesian MAE,\allowbreak{} attribute-\allowbreak{}framing MAE and FrameGap by +2.75,\allowbreak{} +2.05 and +3.88 pp. Same-\allowbreak{}generation M\_\allowbreak{}W minus M\_\allowbreak{}U changes are -\allowbreak{}0.052,\allowbreak{} -\allowbreak{}0.049 and -\allowbreak{}0.067 pp. These are 42\ensuremath{\times}--58\ensuremath{\times} apart in magnitude (Table 5),\allowbreak{} so evidence about aggregation alone cannot identify a prompt-\allowbreak{}level weight effect.\par

% END M:14

% BEGIN M:16 text

\textbf{Unusual Insight.} Better formatting,\allowbreak{} lower operational cue matching,\allowbreak{} lower objective probability error and reduced contextual shifts are separate outcomes. In P1,\allowbreak{} much lower string matching coexists with larger MAE; in P2,\allowbreak{} short arithmetic decomposition approaches the error floor,\allowbreak{} while requesting additional candidates and weights increases error. This changes the prediction from ``more candidates should help'' to a task-\allowbreak{}by-\allowbreak{}configuration choice requiring separate endpoints and interventions.\par

% END M:16

% BEGIN M:18 text

We use MCV as the family name. Direct (D0; \texttt{single}) returns one point; Direct-\allowbreak{}CoT (D\_\allowbreak{}CoT; \texttt{posterior\_\allowbreak{}focus} in Stage 2C) explicitly decomposes the update; MCU (V2; M\_\allowbreak{}U) takes an equal-\allowbreak{}weight mean; MCW (V0; M\_\allowbreak{}W) uses self-\allowbreak{}reported weights. Baseline MCW uses \texttt{multi\_\allowbreak{}repeat}; Upgraded MCW-\allowbreak{}Asc uses \texttt{1p5\_\allowbreak{}asc\_\allowbreak{}repeat},\allowbreak{} a point-\allowbreak{}first response followed by five non-\allowbreak{}decreasing,\allowbreak{} repeat-\allowbreak{}allowed weighted candidates. These mappings align readouts,\allowbreak{} not identical prompt wording across stages. The Stage 1 legacy term ``multi-\allowbreak{}estimate'' is retained in the frozen report below.\par

% END M:18

% BEGIN M:20 text

The study spans five stages. Stage 1 contains 3,\allowbreak{}708 Dev units (1 missing) and 22,\allowbreak{}212 Formal units (21 missing),\allowbreak{} or 25,\allowbreak{}920 planned/\allowbreak{}admitted units with 22 missing overall. Stage 2B contributes 384 units; Stage 2C contributes 1,\allowbreak{}400,\allowbreak{} including 1,\allowbreak{}396 valid. P1 contributes 1,\allowbreak{}920 (Block A 1,\allowbreak{}152 + Block B 768). P2 contributes 2,\allowbreak{}048 (P2-\allowbreak{}A 1,\allowbreak{}280 + P2-\allowbreak{}B 768),\allowbreak{} requiring 2,\allowbreak{}560 calls because separate-\allowbreak{}option generation uses multiple calls per evaluation unit. The totals are 31,\allowbreak{}672 units and 32,\allowbreak{}184 calls,\allowbreak{} excluding retry requests. Stage 1 local/\allowbreak{}CLI configurations and later direct API configurations must not be treated as identical backends.\par

% END M:20

% BEGIN M:22 table

Table 1. Study-\allowbreak{}wide execution and evaluation accounting across all five stages (31,\allowbreak{}672 planned/\allowbreak{}admitted evaluation units; 32,\allowbreak{}184 planned/\allowbreak{}admitted calls excluding retries).\par
\begingroup\scriptsize\setlength{\tabcolsep}{1.3mm}\setlength{\LTpre}{4pt}\setlength{\LTpost}{6pt}
\begin{longtable}{>{\raggedright\arraybackslash}p{39.079mm}>{\raggedright\arraybackslash}p{53.733mm}>{\raggedright\arraybackslash}p{68.388mm}}
\toprule
Stage /\allowbreak{} accounting label & Planned/\allowbreak{}admitted evaluation units & Planned/\allowbreak{}admitted calls (excluding retries) \\
\midrule\endfirsthead
\toprule
Stage /\allowbreak{} accounting label & Planned/\allowbreak{}admitted evaluation units & Planned/\allowbreak{}admitted calls (excluding retries) \\
\midrule\endhead
\bottomrule\endfoot
Stage 1 Dev & 3708 & 3708 \\
Stage 1 Formal & 22212 & 22212 \\
Stage 2B & 384 & 384 \\
Stage 2C & 1400 & 1400 \\
P1 Block A & 1152 & 1152 \\
P1 Block B & 768 & 768 \\
P1 total & 1920 & 1920 \\
P2-\allowbreak{}A & 1280 & 1280 \\
P2-\allowbreak{}B & 768 & 1280 \\
P2 total & 2048 & 2560 \\
Entire study & 31672 & 32184 \\
P2 first-\allowbreak{}attempt parsed calls & 2556 & 99.84\% \\
\end{longtable}
\endgroup

% END M:22

% BEGIN M:37 text

Public Stage 1 preregistration and v2 study materials: \url{https://github.com/asj252/llm-judgment-studies}. \hyperref[sec:appendices]{Appendices A--I} (included in this document) preserve original evidence and provide Stage 2 and P1/P2 reproduction details. The \hyperref[sec:references]{References} section retains all 70 bibliography entries.

% END M:37

% BEGIN M:39 text

\section{Related Work}

% END M:39

% BEGIN M:41 text

\textbf{Judgment Biases in Language Models.} Prior studies of classic cognitive biases in LLMs report mixed patterns across model families and prompting protocols. In anchoring (Experiment A),\allowbreak{} Huang et al. \citep{huang2025understanding} reported lower anchor susceptibility in larger and reasoning models within their 11-\allowbreak{}model SynAnchors panel,\allowbreak{} and Takenami et al. \citep{takenami2025anchoring} similarly observed reduced anchoring in reasoning models during price negotiations. By contrast,\allowbreak{} Lou and Sun \citep{lou2026anchoring} documented persistent anchoring in stronger GPT models that resisted chain-\allowbreak{}of-\allowbreak{}thought and ignore instructions. Most relevant to our design,\allowbreak{} Borjigin et al.'s AnchorBench \citep{borjigin2026anchorbench} separates plausible from irrelevant anchors across prompt pathways. Drawing on human anchoring paradigms \citep{jacowitz1995measures,mussweiler2000overcoming},\allowbreak{} we pair identical numbers with random-\allowbreak{}generator versus reliable-\allowbreak{}source framing and adapt the anchoring index \citep{jacowitz1995measures} to the natural-\allowbreak{}log scale item by item.\par

% END M:41

% BEGIN M:43 text

In hindsight (Experiment B),\allowbreak{} human judges systematically inflate retrospective probability estimates after learning an outcome \citep{fischhoff1975hindsight,arkes1988eliminating,roese2012hindsight},\allowbreak{} a phenomenon examined in recent retrospective-\allowbreak{}judgment research \citep{chen2021retrospective}. In LLMs,\allowbreak{} Malberg et al. \citep{malberg2025comprehensive} included hindsight in a 30-\allowbreak{}bias battery across 20 models,\allowbreak{} and Matsuura et al. \citep{matsuura2026hindsight} reported outcome-\allowbreak{}conditioned shifts consistent with hindsight in comparisons of clinical timelines. Related work reports limitations in simulated pre-\allowbreak{}cutoff ignorance for forecasting \citep{li2026simulated}. Separately,\allowbreak{} self-\allowbreak{}blinding work on discrimination and sycophancy distinguishes instructions to ignore information from preventing access to it \citep{christian2026selfblinding}. Using ProbCOPA items \citep{kamath2026humans,roemmele2011choice},\allowbreak{} we separate prospective foresight,\allowbreak{} retrospective reconstruction with and without an explicit ignore instruction,\allowbreak{} and post-\allowbreak{}outcome confirmation.\par

% END M:43

% BEGIN M:45 text

In base-\allowbreak{}rate tasks (Experiment C),\allowbreak{} humans often neglect base rates under probability framing \citep{kahneman1973psychology,eddy1982probabilistic} but improve under natural frequencies \citep{gigerenzer1995improve}. Prior work examines judgment inconsistencies \citep{macmillanscott2024irrationality},\allowbreak{} coexisting Bayesian and representativeness patterns \citep{li2024dual},\allowbreak{} and models' use of natural-\allowbreak{}frequency representations \citep{araya2025chains}. Early work explored ChatGPT on structured Bayesian problems \citep{krohling2023bayesian}; the ceiling reported here is an empirical finding of our own development screening. In conjunction tasks (Experiment D),\allowbreak{} early choice-\allowbreak{}based work found that GPT-\allowbreak{}3 committed the Linda fallacy \citep{binz2023using}. Related studies examine representativeness and token-\allowbreak{}pattern sensitivity \citep{wang2024real,jiang2024peek},\allowbreak{} including reduced violations after hashing trigger words in a modified Linda task \citep{chadimova2024meaningless}. Numerical probability evaluations are rarer:\allowbreak{} Zhu and Griffiths \citep{zhu2024incoherent} found probability-\allowbreak{}identity incoherence that narrowed in larger models,\allowbreak{} and Imannezhad et al. \citep{imannezhad2026divergent} reported lower violation rates among GPT-\allowbreak{}5 judgments ($18.8\%$) than among human judgments ($58.5\%$). We author new person profiles following \citep{tversky1983extensional} and event pairs following \citep{costello2014surprisingly} to compare separate and joint numerical elicitation. In status quo and default tasks (Experiment E),\allowbreak{} human preferences for current states and preselected defaults are well established \citep{samuelson1988status,johnson2003defaults,jachimowicz2019defaults},\allowbreak{} whereas LLM findings conflict in direction:\allowbreak{} Malberg et al.'s models preferred change on average \citep{malberg2025comprehensive},\allowbreak{} whereas Horton's model-\allowbreak{}based experiments \citep{horton2023homo} (arXiv v2) found status quo preferences under their prompts. Because prior prompts often mix status quo labels,\allowbreak{} prefilled defaults,\allowbreak{} and implied endorsement,\allowbreak{} Experiment E separates these three manipulations on a probability scale.\par

% END M:45

% BEGIN M:47 text

\textbf{Inner-\allowbreak{}Crowd Aggregation and Elicitation Formats.} Human studies have found that aggregating multiple estimates from a single judge ("the crowd within") can improve accuracy under the studied protocols \citep{vul2008crowd,steegen2014measuring,herzog2009wisdom,litvinova2020inner},\allowbreak{} and eliciting full distributions has been studied for overprecision \citep{haran2010simple} and for jointly measuring base-\allowbreak{}rate neglect and conservatism \citep{howe2022base}. In LLMs,\allowbreak{} Verbalized Sampling requests multiple candidates with probabilities in one response to counter mode collapse \citep{zhang2025verbalized},\allowbreak{} and verbalized confidence has been compared with token-\allowbreak{}based confidence \citep{tian2023just}. Other studies examine response consistency and confidence-\allowbreak{}elicitation strategies \citep{xiong2024llms},\allowbreak{} or distributional verbalization for confidence estimation \citep{wang2025distribute}. DiNCo estimates confidence for incompatible claims separately and then normalizes them within its method \citep{wang2026dinco}; this does not make arbitrary continuous numeric candidates mutually exclusive. Forecast aggregation gains can depend on sample diversity and correlation \citep{aitchison2026diversity},\allowbreak{} and Bayesian-\allowbreak{}sampler theory models systematic deviations in human probability judgments using finite samples and a generic prior \citep{zhu2020bayesian}. Related work examines input certainty markers and question-\allowbreak{}answering accuracy \citep{zhou2023navigating},\allowbreak{} integer clustering \citep{dai2026beyond},\allowbreak{} model--human reasoning comparisons \citep{yax2024studying},\allowbreak{} confidence-\allowbreak{}measurement protocols \citep{kim2026same},\allowbreak{} service-\allowbreak{}version drift \citep{chen2024chatgpt},\allowbreak{} and hard--easy patterns in confidence \citep{michael2026confidence}. Unlike temporally separated human inner-\allowbreak{}crowd estimates,\allowbreak{} our multi-\allowbreak{}estimate format generates five interdependent candidates in a single autoregressive call; we test its effects empirically across paradigms without assuming that it replicates human inner-\allowbreak{}crowd mechanisms.\par

% END M:47

% BEGIN M:49 text

\setcounter{subsection}{2}\subsection{Numerical pragmatics,\allowbreak{} scale granularity and output organization}

% END M:49

% BEGIN M:51 text

Number frequency,\allowbreak{} rounding conventions and numeral tokenization motivate scale experiments \citep{dehaene1992cross,krifka2007approximate,razeghi2022impact,singh2024tokenization}. Work on confidence elicitation studies subjective metacognitive reports \citep{dai2026beyond},\allowbreak{} whereas our Bayesian and matched-\allowbreak{}framing tasks have item-\allowbreak{}level normative probability targets. Results from interpretable judge models \citep{feldhus2026judge} do not establish the same internal process in our tested endpoints. Surface percentage-\allowbreak{}token matches are observable outcomes. Whether a reduction reflects less psychological copying or changed grid accessibility is unresolved; the former is a hypothesis,\allowbreak{} not an inference licensed by matching alone.\par

% END M:51

% BEGIN M:53 text

\section{Methods}

% END M:53

% BEGIN M:55 text

\subsection{Preregistration,\allowbreak{} completion,\allowbreak{} and configurations}

% END M:55

% BEGIN M:57 text

The plan,\allowbreak{} hypothesis definitions,\allowbreak{} sidedness,\allowbreak{} smallest effect sizes of interest (SESOIs),\allowbreak{} missingness rules,\allowbreak{} screening criteria and configuration-\allowbreak{}specific Holm lists were frozen before formal collection and published on October 22,\allowbreak{} 2026 (commit \texttt{802d760}). Addendum 1 (\texttt{940aee5},\allowbreak{} October 24) changed the codex executable binding before its formal collection. Addendum 2 recorded stopping and disposition decisions,\allowbreak{} including a post-\allowbreak{}hoc session closure. This v2 completes the original six configurations without expanding their frozen lists or creating new cross-\allowbreak{}model confirmatory tests.\par

% END M:57

% BEGIN M:59 text

\textbf{Table 2. Frozen execution identities and confirmatory scope}\par

% END M:59

% BEGIN M:61 text

Identifiers are transcribed from frozen metadata. They identify requested configurations; unobserved provider-\allowbreak{}side routing or service changes are not independently verified.\par

% END M:61

% BEGIN M:63 table

\begingroup\scriptsize\setlength{\tabcolsep}{1.3mm}\setlength{\LTpre}{4pt}\setlength{\LTpost}{6pt}
\begin{longtable}{>{\raggedright\arraybackslash}p{18.615mm}>{\raggedright\arraybackslash}p{68.503mm}>{\raggedright\arraybackslash}p{56.590mm}>{\raggedright\arraybackslash}p{14.892mm}}
\toprule
Alias & Frozen model identifier & Interface /\allowbreak{} setting & Tests \\
\midrule\endfirsthead
\toprule
Alias & Frozen model identifier & Interface /\allowbreak{} setting & Tests \\
\midrule\endhead
\bottomrule\endfoot
codex & gpt-\allowbreak{}6.1-\allowbreak{}sol & codex 0.163.0-\allowbreak{}alpha.10.2; medium & 11 \\
Sonnet 5.5 & claude-\allowbreak{}sonnet-\allowbreak{}5-\allowbreak{}5-\allowbreak{}medium & agy 1.4.0; medium & 12 \\
Qwen3 & Qwen/\allowbreak{}Qwen3-\allowbreak{}8B & local GGUF Q8\_\allowbreak{}0; non-\allowbreak{}thinking & 0 \\
gpt-\allowbreak{}oss-\allowbreak{}120b & gpt-\allowbreak{}oss-\allowbreak{}120b-\allowbreak{}medium & agy 1.4.0; medium & 3 \\
Flash low & gemini-\allowbreak{}3.8-\allowbreak{}flash-\allowbreak{}low & agy 1.4.0; low & 9 \\
Flash high & gemini-\allowbreak{}3.8-\allowbreak{}flash-\allowbreak{}high & agy 1.4.0; high & 10 \\
\end{longtable}
\endgroup

% END M:63

% BEGIN M:72 text

For codex and agy,\allowbreak{} provider system prompts,\allowbreak{} sampling controls and server seeds are not fully observable. Calls excluded tools and web search and used no researcher-\allowbreak{}written system prompt. Five candidates generated in one autoregressive call are interdependent. Requested model identities are preserved from the freeze rather than replaced retrospectively by names from another task or session.\par

% END M:72

% BEGIN M:74 text

Flash high required a separately bound execution-\allowbreak{}window continuation. F0 contributes 3,\allowbreak{}049 admitted formal units (2,\allowbreak{}687 r0 and 362 r1; 3,\allowbreak{}048 OK,\allowbreak{} one missing); F1 contributes 653 (531 r0 and 122 r1; all OK). Their disjoint union is exactly 3,\allowbreak{}702. The inherited development set is used once. F1 follows \texttt{window-\allowbreak{}addendum-\allowbreak{}high-\allowbreak{}20261104/\allowbreak{}protocol.json},\allowbreak{} whose start boundary is \texttt{2026-\allowbreak{}11-\allowbreak{}05T00:\allowbreak{}00:\allowbreak{}00+08:\allowbreak{}00},\allowbreak{} after the earlier study-\allowbreak{}window cutoff. The two batches retain their own sessions and fingerprints,\allowbreak{} while the union uses the original ten-\allowbreak{}hypothesis Flash-\allowbreak{}high list. Tables F1 and H11 expose the execution-\allowbreak{}window partition. Cluster D /\allowbreak{} w01 spans F0 (7 r0 calls) and F1 (3 r0 calls) across both response formats. There are zero cross-\allowbreak{}batch r0/\allowbreak{}r1 repeat pairs:\allowbreak{} every repeated unit has its r0 and r1 calls in the same batch.\par

% END M:74

% BEGIN M:76 text

F0 and F1 are non-\allowbreak{}randomized execution windows confounded with item schedule order and collection time. Differences are exploratory and do not identify a causal execution-\allowbreak{}window or model effect.\par

% END M:76

% BEGIN M:78 text

\subsection{Response Formats,\allowbreak{} Call Protocol,\allowbreak{} and Real Example}

% END M:78

% BEGIN M:80 text

Each unit (item $\times$ condition $\times$ response format $\times$ round) was executed in an independent,\allowbreak{} stateless call with seeded cluster-\allowbreak{}then-\allowbreak{}unit shuffling. Parsing or contamination failures allowed one retry,\allowbreak{} retaining the first valid attempt. The \textbf{direct format} (\texttt{ASK\_\allowbreak{}DIRECT}) requests a single final-\allowbreak{}line number. The \textbf{multi-\allowbreak{}estimate format} (\texttt{ASK\_\allowbreak{}MULTI}) requests five estimates with weights in $[0, 1]$ summing to 1,\allowbreak{} aggregated by the parser as $\bar{y}_{\text{multi}} = \sum_{m=1}^5 \tilde{w}_m y_m$ (normalizing weights by their sum whenever $\sum w_m > 0$). Experiments B,\allowbreak{} C,\allowbreak{} E,\allowbreak{} and separate D statements use the verbatim $0$--$100\%$ suffixes from $\delta$-\allowbreak{}NLI \citep{jhang2026hearsay}; Experiment A requests positive finite quantities; joint D requests four probabilities per line.\par

% END M:80

% BEGIN M:82 text

\subsubsection*{Box 1. From a recorded answer to a weighted estimate}

% END M:82

% BEGIN M:84 text

This post hoc display example is not a preregistered analysis or statistical evidence for a format effect. We fixed the completed codex configuration,\allowbreak{} formal round r0,\allowbreak{} and the separately elicited conjunction condition,\allowbreak{} then selected the first self-\allowbreak{}written person-\allowbreak{}profile item in lexicographic ID order with valid admitted direct and multi-\allowbreak{}estimate records. The selected item is p01; both records retain their first valid attempt (attempt 0).\par

% END M:84

% BEGIN M:86 text

\textbf{Shared question stem (verbatim).}\par

% END M:86

% BEGIN M:88 code

\begin{Verbatim}[fontsize=\scriptsize,breaklines=true,breakanywhere=true]
Here is a situation: Tomasz is 34 and spends nearly every weekend hiking in the hills near his home, carrying a camera and a field guide. He keeps a notebook listing every bird species he has seen and often gets up before dawn to reach his favourite lookout.

Using common sense, how likely is it that the following statement is true?
Statement: Tomasz drives a long-haul truck for a living and belongs to a local birdwatching club.
\end{Verbatim}

% END M:88

% BEGIN M:95 text

\textbf{Direct-\allowbreak{}format request suffix (verbatim).}\par

% END M:95

% BEGIN M:97 code

\begin{Verbatim}[fontsize=\scriptsize,breaklines=true,breakanywhere=true]
What is the probability that the statement is true?

On the last line of your answer, give the answer strictly in the following format:
Probability: NN%
where NN is a number from 0 to 100 (decimals allowed), and it must be followed by a percent sign.

Do not use any tools; answer directly.
\end{Verbatim}

% END M:97

% BEGIN M:107 text

\textbf{Observed direct final output (numeric line excerpted verbatim).}\par

% END M:107

% BEGIN M:109 code

\begin{Verbatim}[fontsize=\scriptsize,breaklines=true,breakanywhere=true]
Probability: 2%
\end{Verbatim}

% END M:109

% BEGIN M:113 text

\textbf{Multi-\allowbreak{}estimate request suffix (verbatim).}\par

% END M:113

% BEGIN M:115 code

\begin{Verbatim}[fontsize=\scriptsize,breaklines=true,breakanywhere=true]
Give 5 probability estimates that the statement is true, each with a weight for how likely you think that estimate is; the weights sum to 1.

At the end of your answer, give the 5 estimates strictly in the following format, one per line:
Estimate 1: NN% | Weight: W
Estimate 2: NN% | Weight: W
Estimate 3: NN% | Weight: W
Estimate 4: NN% | Weight: W
Estimate 5: NN% | Weight: W
where each NN is a number from 0 to 100 (decimals allowed) followed by a percent sign, and each W is a number from 0 to 1 (decimals allowed); the five weights sum to 1.

Do not use any tools; answer directly.
\end{Verbatim}

% END M:115

% BEGIN M:129 text

\textbf{Observed multi-\allowbreak{}estimate final output (numeric lines excerpted verbatim).}\par

% END M:129

% BEGIN M:131 code

\begin{Verbatim}[fontsize=\scriptsize,breaklines=true,breakanywhere=true]
Estimate 1: 0.5% | Weight: 0.15
Estimate 2: 1% | Weight: 0.30
Estimate 3: 2% | Weight: 0.30
Estimate 4: 3% | Weight: 0.15
Estimate 5: 5% | Weight: 0.10
\end{Verbatim}

% END M:131

% BEGIN M:139 text

The weights sum to 1. The parser's recorded weighted mean is 0.15 \ensuremath{\times} 0.5 + 0.30 \ensuremath{\times} 1 + 0.30 \ensuremath{\times} 2 + 0.15 \ensuremath{\times} 3 + 0.10 \ensuremath{\times} 5 = \textbf{1.925\%},\allowbreak{} compared with the separate direct response of \textbf{2\%}. The displayed outputs omit the answers' introductory prose; no output wording is repaired or synthesized. These are two calls to the same item under different response formats,\allowbreak{} while the five candidates come from one call and are not independent samples. This example shows the requested format and parsing operation,\allowbreak{} without identifying an internal mechanism or establishing an effect.\par

% END M:139

% BEGIN M:141 text

Source units are \texttt{D:\allowbreak{}p01\_\allowbreak{}S\_\allowbreak{}conj\_\allowbreak{}direct|r0} and \texttt{D:\allowbreak{}p01\_\allowbreak{}S\_\allowbreak{}conj\_\allowbreak{}multi|r0},\allowbreak{} both in the admitted codex session \texttt{11df552506824d299603df5db02361e7}. The source prompts,\allowbreak{} final-\allowbreak{}answer text and hashes are retained in the editorial evidence record.\par

% END M:141

% BEGIN M:143 text

\subsection{Experimental Designs and Operational Definitions (A--E)}

% END M:143

% BEGIN M:145 text

\textbf{Experiment A (Anchoring,\allowbreak{} 10 dev /\allowbreak{} 50 formal items,\allowbreak{} 500 formal units).} We selected 36 eligible items with standalone,\allowbreak{} positive factual truths from the 60-\allowbreak{}item SynAnchors semantic subset \citep{huang2025understanding} (\texttt{TimTargaryen/\allowbreak{}SynAnchors},\allowbreak{} rev. \texttt{e6df428}) and wrote 24 verifiable factual items (formal split:\allowbreak{} 30 SynAnchors,\allowbreak{} 20 self-\allowbreak{}written). High and low anchors were frozen at $\text{truth} \times 1.5$ and $\text{truth} \div 1.5$,\allowbreak{} with rounding at the prespecified item precision. Each item has 10 units:\allowbreak{} unanchored baseline,\allowbreak{} \textbf{irrelevant anchors} (high/\allowbreak{}low,\allowbreak{} framed as outputs of an unrelated random-\allowbreak{}number generator with a higher/\allowbreak{}lower comparison question),\allowbreak{} and \textbf{informative anchors} (high/\allowbreak{}low,\allowbreak{} identical numbers framed as a "reliable reference source"),\allowbreak{} crossed with both response formats. We define the item-\allowbreak{}level log anchoring index as $\text{AI}_c = [\ln(\text{estimate}_{c,\text{high}}) - \ln(\text{estimate}_{c,\text{low}})] / [\ln(\text{anchor}_{\text{high}}) - \ln(\text{anchor}_{\text{low}})]$ for $c \in \{\text{irrelevant}, \text{informative}\}$,\allowbreak{} and source selectivity as $S = \text{AI}_{\text{informative}} - \text{AI}_{\text{irrelevant}}$. Because quantity estimates have no upper bound,\allowbreak{} Experiment A uses complete-\allowbreak{}item subsets without worst-\allowbreak{}case missingness substitution. \emph{Local validity condition:\allowbreak{}} Because informative anchors deviate from truth by a factor of $1.5$,\allowbreak{} $\text{AI}_{\text{informative}}$ and $S$ measure selectivity to source framing rather than factual accuracy; a model that knows the exact answer may reasonably ignore both anchors.\par

% END M:145

% BEGIN M:147 text

\textbf{Experiment B (Hindsight,\allowbreak{} 15 dev /\allowbreak{} 90 formal items,\allowbreak{} 1,\allowbreak{}440 formal units).} All 105 COPA effect items \citep{roemmele2011choice} in ProbCOPA \citep{kamath2026humans} (\texttt{McGill-\allowbreak{}NLP/\allowbreak{}probabilistic-\allowbreak{}reasoning},\allowbreak{} rev. \texttt{b983b12}) were matched after removing \texttt{canary} entries. Each item has two outcomes $O_1, O_2$ (not necessarily mutually exclusive or exhaustive) elicited across 16 units:\allowbreak{} prospective foresight ($F$); outcome disclosure with an instruction to set the outcome aside and reconstruct a prediction "from the situation alone" ($\text{HI}$); outcome disclosure with the same retrospective reconstruction question but without the ignore instruction ($\text{HP}$); and post-\allowbreak{}outcome confirmation disclosing $O_k$ before repeating the $F$ question ($\text{PO}$),\allowbreak{} crossed with $k \in \{1, 2\}$ and both response formats. We define reconstructed prospective shift $b = \frac{1}{2}\sum_{k=1}^2 [\text{HI}(O_k) - F(O_k)]$,\allowbreak{} uninstructed shift $b_{\text{HP}} = \frac{1}{2}\sum_{k=1}^2 [\text{HP}(O_k) - F(O_k)]$,\allowbreak{} post-\allowbreak{}outcome confirmation error $e_{\text{PO}} = \frac{1}{2}\sum_{k=1}^2 [100 - \text{PO}(O_k)]$,\allowbreak{} and input-\allowbreak{}condition contrast $c = \frac{1}{2}\sum_{k=1}^2 [\text{PO}(O_k) - F(O_k)]$. \emph{Local validity condition:\allowbreak{}} Given COPA's public availability,\allowbreak{} $b$ measures reconstructed prospective shifts after outcome disclosure rather than guaranteed memory-\allowbreak{}free hindsight bias,\allowbreak{} and $b_{\text{HP}} - b$ isolates the wording effect of the ignore instruction. Human ratings are descriptive references,\allowbreak{} not true probabilities or calibration targets.\par

% END M:147

% BEGIN M:149 text

\textbf{Experiment C (Base Rates,\allowbreak{} 4 dev /\allowbreak{} 20 formal scenario clusters = 40 formal items,\allowbreak{} 480 formal units).} Twenty-\allowbreak{}four newly written scenarios were crossed with two test accuracy sets (Set A:\allowbreak{} $90\%$ hit /\allowbreak{} $10\%$ false-\allowbreak{}positive rate,\allowbreak{} posteriors $8.33\%, 50.00\%, 79.41\%$; Set B:\allowbreak{} $80\%$ /\allowbreak{} $5\%$,\allowbreak{} posteriors $13.91\%, 64.00\%, 87.27\%$). Each item has 12 units:\allowbreak{} 3 base rates ($1\%, 10\%, 30\%$) $\times$ 2 presentations (probabilities vs. natural frequencies out of $10,000$,\allowbreak{} which also reduce calculation steps) $\times$ 2 response formats. Responses clipped to $[0.5\%, 99.5\%]$ are logit-\allowbreak{}transformed and regressed on $\text{logit}(\text{base rate})$ by 3-\allowbreak{}point OLS to obtain the \textbf{slope} (normative $= 1$,\allowbreak{} interpreted together with the intercept and probability error); \textbf{Bayesian error} is the mean unclipped absolute deviation from the Bayes posterior across the three base rates,\allowbreak{} equally weighted across accuracy sets and scenarios.\par

% END M:149

% BEGIN M:151 text

\textbf{Experiment D (Conjunction,\allowbreak{} 7 dev /\allowbreak{} 35 formal items:\allowbreak{} 20 person items $P$,\allowbreak{} 15 event pairs $W$; 350 formal units).} Newly written person items $P$ pair a character sketch with a fitting event $A$ and a less likely event $B$ \citep{tversky1983extensional}; event pairs $W$ span positive,\allowbreak{} independent,\allowbreak{} and negative dependence strata (5 formal items each) \citep{costello2014surprisingly}. Each item has 10 units:\allowbreak{} statements $A$,\allowbreak{} $B$,\allowbreak{} "$B$ and $A$,\allowbreak{}" and "$B$ or $A$ (or both)" elicited \textbf{separately} ($\text{S}$,\allowbreak{} 4 independent calls) or \textbf{jointly} ($\text{J}$,\allowbreak{} 4 values in 1 call in a seeded item-\allowbreak{}fixed order),\allowbreak{} crossed with both formats. In joint multi-\allowbreak{}estimate calls,\allowbreak{} one weight vector aggregates the four statements before computing $\Delta = P(A \text{ and } B) - P(B)$,\allowbreak{} $\Delta_{\min} = P(A \text{ and } B) - \min(P(A), P(B))$,\allowbreak{} and additive-\allowbreak{}identity deviation $X = P(A) + P(B) - P(A \text{ and } B) - P(A \text{ or } B)$. \emph{Local validity condition:\allowbreak{}} $\Delta_S$ includes between-\allowbreak{}call variance,\allowbreak{} $\Delta_J$ makes the extension constraint ($\Delta \le 0$) visible within the prompt,\allowbreak{} and D2 ($\Delta_S - \Delta_J > 0$) tests mean signed $\Delta$ rather than binary violation rates.\par

% END M:151

% BEGIN M:153 text

\textbf{Experiment E (Status Quo and Defaults,\allowbreak{} 8 dev /\allowbreak{} 32 formal items,\allowbreak{} 448 formal units).} Newly written binary everyday decision items ask the probability that target option $T$ (listed first on 17 formal items,\allowbreak{} second on 15) is better for the protagonist across 14 units:\allowbreak{} neutral baseline,\allowbreak{} plus a \textbf{status quo label} (explicitly randomly assigned),\allowbreak{} a \textbf{prefilled default} (explicitly randomly set),\allowbreak{} and \textbf{informed advice} (from an independent adviser who reviewed the person's circumstances),\allowbreak{} each assigned to $T$ or to the other option $O$,\allowbreak{} with zero switching cost,\allowbreak{} crossed with both formats. We define $E_m = y(m \text{ assigned to } T) - y(m \text{ assigned to } O)$ for $m \in \{\text{status\_quo}, \text{default}, \text{advice}\}$ and $S_E = E_{\text{advice}} - E_{\text{status\_quo}}$. $E_{\text{advice}}$ serves as a prespecified positive-\allowbreak{}control diagnostic outside Holm (threshold $\ge 5\text{ pp}$). \emph{Local validity condition:\allowbreak{}} Experiment E measures third-\allowbreak{}party probability judgments rather than behavioral choices; loss framing and omission bias were not manipulated.\par

% END M:153

% BEGIN M:155 text

\subsection{Hypotheses,\allowbreak{} Screening,\allowbreak{} Inference,\allowbreak{} and Execution Accounting}

% END M:155

% BEGIN M:157 text

We preregistered four hypotheses per experiment (20 total; SESOI $= 0.10$ in A,\allowbreak{} $0.20$ for slopes in C1--C3,\allowbreak{} and $5\text{ pp}$ in B,\allowbreak{} C4,\allowbreak{} D,\allowbreak{} E):\allowbreak{}
-\allowbreak{} \textbf{Direct-\allowbreak{}format baseline tests:\allowbreak{}} One-\allowbreak{}sided $>0$ for \textbf{A1} ($\text{AI}_{\text{irrelevant}}$),\allowbreak{} \textbf{A2} ($S$),\allowbreak{} \textbf{B1} ($b$),\allowbreak{} \textbf{B2} ($b_{\text{HP}} - b$),\allowbreak{} \textbf{C1} ($1 - \text{probability slope}$),\allowbreak{} \textbf{C2} ($\text{frequency} - \text{probability slope}$),\allowbreak{} \textbf{D1} ($\Delta_S$ on $P$),\allowbreak{} and \textbf{D2} ($\Delta_S - \Delta_J$ on $P$); two-\allowbreak{}sided for \textbf{E1} ($E_{\text{status\_quo}}$) and \textbf{E2} ($E_{\text{default}}$).
-\allowbreak{} \textbf{Multi-\allowbreak{}estimate minus direct format-\allowbreak{}difference tests (all two-\allowbreak{}sided):\allowbreak{}} \textbf{A3} ($\text{AI}_{\text{irrelevant}}$),\allowbreak{} \textbf{A4} ($S$),\allowbreak{} \textbf{B3} ($b$),\allowbreak{} \textbf{B4} ($e_{\text{PO}}$),\allowbreak{} \textbf{C3} (probability slope),\allowbreak{} \textbf{C4} (probability Bayes error),\allowbreak{} \textbf{D3} ($\Delta_S$ on $P$),\allowbreak{} \textbf{D4} ($\Delta_J$ on $P$),\allowbreak{} \textbf{E3} ($E_{\text{status\_quo}}$),\allowbreak{} and \textbf{E4} ($E_{\text{default}}$).\par

% END M:157

% BEGIN M:161 text

All six development sets passed the original acceptance rule. C's ceiling gate triggered across all configurations; A's known-\allowbreak{}truth gate triggered in codex and both Flash settings. Development power simulations and the E no-\allowbreak{}expansion exception determined the remaining lists (Table H1). Qwen3 was retained as a descriptive benchmark with zero confirmatory hypotheses. No hypothesis was rescued by rescreening after family reduction.\par

% END M:161

% BEGIN M:163 text

Inference uses 5,\allowbreak{}000 whole-\allowbreak{}cluster bootstrap draws (seed 2026101903),\allowbreak{} non-\allowbreak{}interpolated quantiles and step-\allowbreak{}down Holm adjustment within each configuration by experiment family at alpha 0.05. Rank-\allowbreak{}specific intervals are not fixed 95\% simultaneous intervals. The strongest retained tier requires an adjusted bound beyond its SESOI; weaker directional tiers distinguish whether the point reaches the threshold. Reverse one-\allowbreak{}sided bounds below zero at the same Holm-\allowbreak{}step alpha are descriptive opposite-\allowbreak{}direction diagnostics,\allowbreak{} not extra confirmatory rejections. Finite-\allowbreak{}bootstrap tail proportions displayed as zero are reported as p < .001,\allowbreak{} not zero population probabilities.\par

% END M:163

% BEGIN M:165 text

For bounded probability outcomes,\allowbreak{} the prespecified sensitivity analysis retains observed values and assigns missing cells feasible worst-\allowbreak{}case 0/\allowbreak{}100 endpoints. It has its own Holm ordering and full planned denominator. The complete-\allowbreak{}case rejected flag remains visible,\allowbreak{} but final directional strength follows the protection rule. A uses positive unbounded quantities and complete-\allowbreak{}item subsets without finite worst-\allowbreak{}case protection. E recommendation-\allowbreak{}control interpretation caveats remain separate from development eligibility.\par

% END M:165

% BEGIN M:167 text

Each configuration contributes 3,\allowbreak{}218 primary r0 and 484 repeat r1 formal units. Across six configurations,\allowbreak{} 22,\allowbreak{}294 formal JSONL lines yield 22,\allowbreak{}212 admitted units (22,\allowbreak{}191 OK and 21 missing),\allowbreak{} excluding 82 historical lines. Primary r0 has 19,\allowbreak{}308 admitted units (19,\allowbreak{}288 OK,\allowbreak{} 20 missing); r1 has 2,\allowbreak{}904 (2,\allowbreak{}903 OK,\allowbreak{} one missing). Development adds 3,\allowbreak{}708 admitted units (3,\allowbreak{}707 OK,\allowbreak{} one missing),\allowbreak{} giving 25,\allowbreak{}920 admitted units overall. Admitted is an accounting category,\allowbreak{} not a synonym for usable numeric answers. Table A1 separates stages,\allowbreak{} rounds,\allowbreak{} exclusions and the two Flash-\allowbreak{}high batches.\par

% END M:167

% BEGIN M:169 text

The earlier codex post-\allowbreak{}hoc closure retained 359 completed units after integrity checks following an append permission error; it remains a disclosed inclusion deviation. Minor frozen descriptive resampling departures for B's tied-\allowbreak{}F stratum and C's marginal item proportions also remain. The audited v2 supplementary inventory now exposes all 20 endpoint expressions per configuration,\allowbreak{} including B2 and C1-\allowbreak{}C2 when the older descriptive export lacked a value. This adds descriptive transparency without changing confirmatory code or family membership.\par

% END M:169

% BEGIN M:171 text

\section{Results}

% END M:171

% BEGIN M:173 text

\subsection{Confirmatory decisions and missingness protection}

% END M:173

% BEGIN M:175 text

Of 45 frozen tests in five configurations,\allowbreak{} 12 reject on complete cases and 11 retain directional confirmation; 33 complete-\allowbreak{}case tests do not reject. One retained test has an adjusted bound beyond its SESOI,\allowbreak{} while ten are point\_\allowbreak{}below. None is point\_\allowbreak{}reaches. These outcomes do not establish equivalence,\allowbreak{} and not all contrasts test a debiasing benefit.\par

% END M:175

% BEGIN M:177 text

The strongest effect is gpt-\allowbreak{}oss-\allowbreak{}120b B4:\allowbreak{} multi-\allowbreak{}estimate minus direct post-\allowbreak{}outcome confirmation error is +10.68 pp [7.61,\allowbreak{} 13.81] complete-\allowbreak{}case and +10.04 pp [6.78,\allowbreak{} 14.32] worst-\allowbreak{}case (both p < .001). Larger e\_\allowbreak{}PO means greater error:\allowbreak{} this is deterioration,\allowbreak{} not an anchoring reduction. Sonnet's B4 instead decreases error (-\allowbreak{}4.19 pp [-\allowbreak{}7.35,\allowbreak{} -\allowbreak{}1.08]),\allowbreak{} with a sub-\allowbreak{}5-\allowbreak{}pp absolute point estimate.\par

% END M:177

% BEGIN M:179 text

Flash-\allowbreak{}low B3 illustrates the distinction between rejection and retained confirmation:\allowbreak{} +1.64 pp [0.09,\allowbreak{} 3.24],\allowbreak{} p = .0108 at alpha .0167 with 89 clusters,\allowbreak{} becomes +1.14 pp [-\allowbreak{}0.79,\allowbreak{} 3.18],\allowbreak{} p = .1724 with 90 clusters under protection. Its complete-\allowbreak{}case rejected flag remains true,\allowbreak{} but missing\_\allowbreak{}protected and final.directional are false. Flash-\allowbreak{}high B3 retains direction (+4.53 pp [2.76,\allowbreak{} 6.45] complete-\allowbreak{}case; +4.11 pp [1.98,\allowbreak{} 6.58] worst-\allowbreak{}case),\allowbreak{} without a bound exceeding 5 pp.\par

% END M:179

% BEGIN M:183 text

\clearpage\begin{center}\includegraphics[width=\linewidth,height=.70\textheight,keepaspectratio]{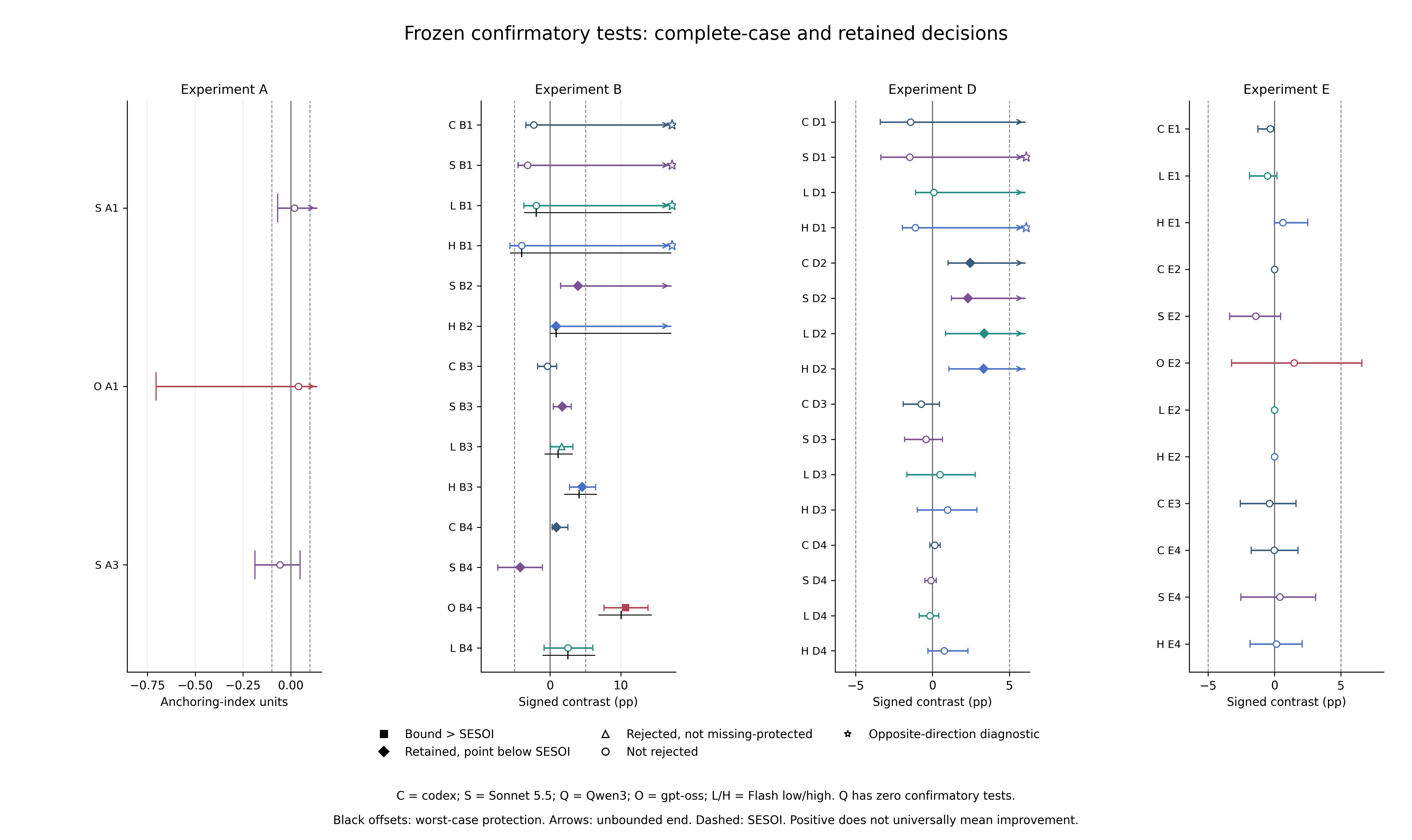}\end{center}
\textbf{Figure 1. All 45 frozen confirmatory tests.} Squares:\allowbreak{} final bound exceeds SESOI; diamonds:\allowbreak{} retained directional point below SESOI; triangles:\allowbreak{} complete-\allowbreak{}case rejection not retained under missingness protection; circles:\allowbreak{} no complete-\allowbreak{}case rejection. Hollow stars mark six opposite-\allowbreak{}direction diagnostic bounds,\allowbreak{} not additional rejections. Colored intervals show complete-\allowbreak{}case Holm-\allowbreak{}step bounds; black offset intervals show protection when required. Arrows indicate unbounded ends. A uses native anchoring-\allowbreak{}index units,\allowbreak{} not factual percentage-\allowbreak{}error reduction; B/\allowbreak{}D/\allowbreak{}E use signed operational contrasts in pp. A positive sign does not universally indicate improvement. Qwen3 forwarded zero hypotheses. Table H2 reports complete numeric results.\par\clearpage

% END M:183

% BEGIN M:185 text

Six reverse-\allowbreak{}bound diagnostics qualify:\allowbreak{} B1 in codex,\allowbreak{} Sonnet,\allowbreak{} Flash low and Flash high,\allowbreak{} and D1 in Sonnet and Flash high. They use rank-\allowbreak{}specific alpha,\allowbreak{} not a fixed 95\% cutoff. They are not six new tests.\par

% END M:185

% BEGIN M:187 text

\subsection{Anchoring and source framing}

% END M:187

% BEGIN M:189 text

The configurations do not exhibit a uniform response to random versus informative-\allowbreak{}source framing. Codex and Sonnet have small mean direct irrelevant-\allowbreak{}anchor indices. Qwen3's paired direct index is approximately 0.946 and falls to 0.338 with multi-\allowbreak{}estimate elicitation; the gpt-\allowbreak{}oss paired index instead rises from 0.039 to 0.197. These are descriptive paired means; single-\allowbreak{}format endpoint denominators can differ.\par

% END M:189

% BEGIN M:191 text

Sonnet A1/\allowbreak{}A3 and gpt-\allowbreak{}oss A1 do not reject. Codex and both Flash A families were downgraded after development known-\allowbreak{}truth screening. Source sensitivity does not measure factual accuracy:\allowbreak{} a model knowing the answer may appropriately ignore an inaccurate informative anchor. Tables H3 and F2 retain the distinction.\par

% END M:191

% BEGIN M:193 text

\subsection{Reconstructed hindsight,\allowbreak{} wording and confirmation error}

% END M:193

% BEGIN M:195 text

Direct b is negative in the four configurations that forwarded B1; reverse adjusted upper bounds are -\allowbreak{}1.23,\allowbreak{} -\allowbreak{}1.84,\allowbreak{} -\allowbreak{}0.29 and -\allowbreak{}2.42 pp in codex,\allowbreak{} Sonnet,\allowbreak{} Flash low and Flash high. This is the operational HI-\allowbreak{}minus-\allowbreak{}F contrast. Public-\allowbreak{}item exposure,\allowbreak{} reconstruction wording and conditioning on noisy prospective ratings remain possible influences; none is identified as the cause of the overall negative means.\par

% END M:195

% BEGIN M:197 text

Removing the ignore instruction while retaining the retrospective question produces retained B2 shifts in Sonnet (+3.93 pp) and Flash high (+0.85 pp). The multi-\allowbreak{}minus-\allowbreak{}direct b shift B3 retains direction in Sonnet and Flash high but not in Flash low after protection. B4 varies by configuration,\allowbreak{} including gpt-\allowbreak{}oss deterioration and Sonnet improvement. Differences in b and e\_\allowbreak{}PO cannot be pooled as a universal attenuation factor. Human ProbCOPA ratings remain descriptive references,\allowbreak{} not normative probability targets.\par

% END M:197

% BEGIN M:199 text

\subsection{Base rates and the direct-\allowbreak{}response ceiling}

% END M:199

% BEGIN M:201 text

All C endpoints are descriptive after the development ceiling. Direct slopes are near one and posterior errors small; displayed 1.000 and 0.000 values are rounded. A slope near one alone is insufficient:\allowbreak{} intercepts and probability error also matter. Multi-\allowbreak{}estimate responses increase paired posterior error in Sonnet by approximately 1.006 pp and in Qwen3 by 0.647 pp,\allowbreak{} while codex error slightly decreases. Elicitation and aggregation change together,\allowbreak{} so increased error does not identify aggregation arithmetic as an isolated mechanism.\par

% END M:201

% BEGIN M:203 text

\subsection{Conjunction presentation and call structure}

% END M:203

% BEGIN M:205 text

Direct separate-\allowbreak{}minus-\allowbreak{}joint signed excess D2 retains confirmation in codex (+2.46 pp),\allowbreak{} Sonnet (+2.30 pp),\allowbreak{} Flash low (+3.37 pp) and Flash high (+3.34 pp),\allowbreak{} all with sub-\allowbreak{}SESOI points. Qwen3 and gpt-\allowbreak{}oss D2 remain descriptive because they were not forwarded. Joint prompts make constraints visible while placing related values in the same call; separate prompts involve independent calls. D2 therefore does not identify constraint visibility alone or confirm a reduction in binary violation rates. D3/\allowbreak{}D4 format contrasts do not yield retained confirmation.\par

% END M:205

% BEGIN M:207 text

\subsection{Status quo,\allowbreak{} defaults and informed advice}

% END M:207

% BEGIN M:209 text

No E test yields retained confirmation. This does not establish absence of a framing effect or equivalence. Several status/\allowbreak{}default means are small under explicit random-\allowbreak{}label provenance and zero switching cost,\allowbreak{} while advice effects remain large. Advice effects decline under multi-\allowbreak{}estimate elicitation in gpt-\allowbreak{}oss and both Flash settings,\allowbreak{} with a different pattern in codex. All formal recommendation controls remain above the 5-\allowbreak{}pp diagnostic threshold; this diagnostic does not remove frozen tests.\par

% END M:209

% BEGIN M:211 text

Formal neutral summaries use 32 distinct items. The development neutral diagnostic uses eight original manifest IDs,\allowbreak{} not E01-\allowbreak{}E08 renumbering; E12 did not cause formal-\allowbreak{}item screening. Tables H3-\allowbreak{}H4 report complete paired subsets,\allowbreak{} controls and supplemental endpoints.\par

% END M:211

% BEGIN M:215 text

\clearpage\begin{center}\includegraphics[width=\linewidth,height=.70\textheight,keepaspectratio]{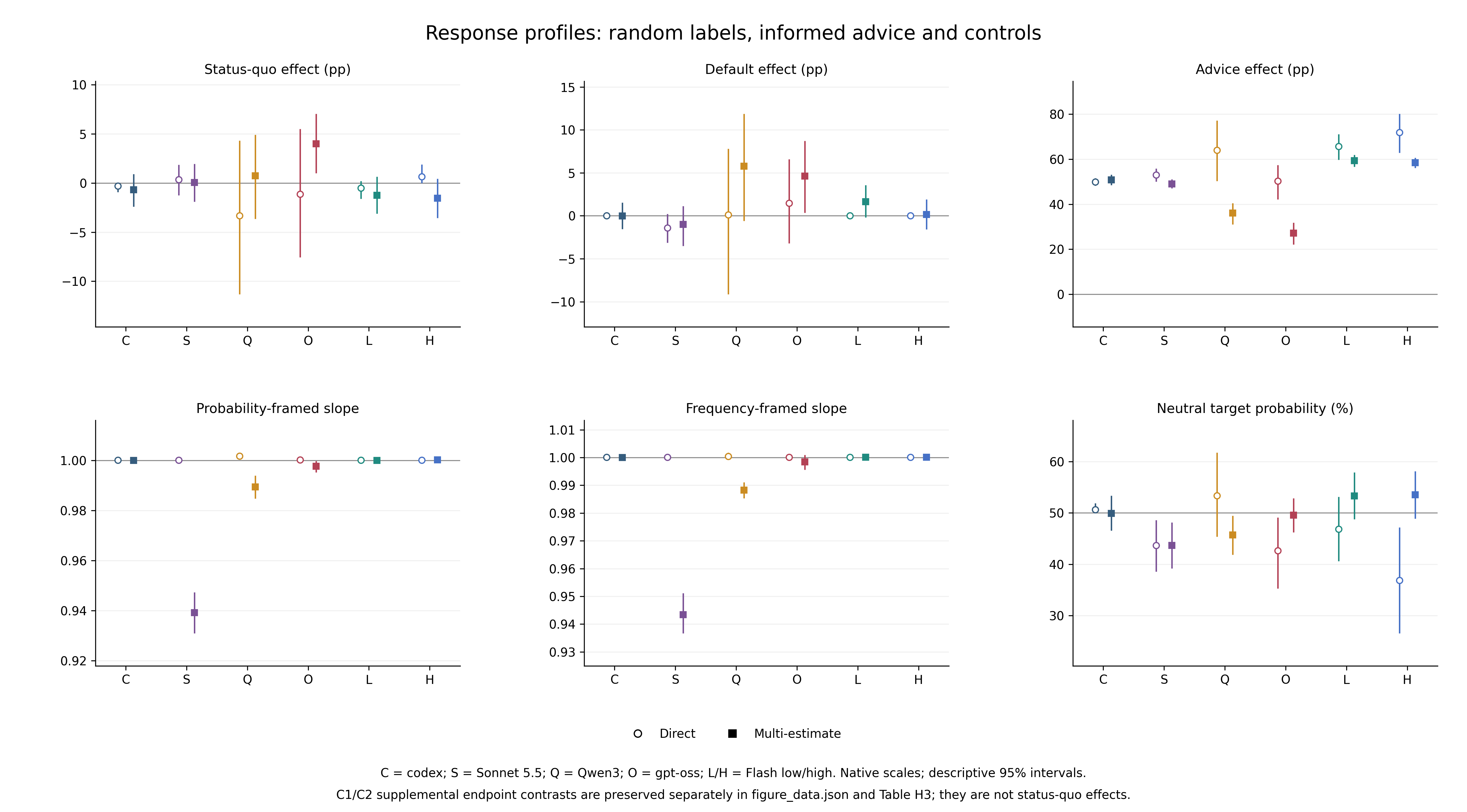}\end{center}
\textbf{Figure 2. Direct and multi-\allowbreak{}estimate descriptive profiles across six configurations.} Status-\allowbreak{}quo,\allowbreak{} default and advice effects occupy the top panels; C probability/\allowbreak{}frequency slopes and E neutral judgments occupy the lower panels. Intervals are unadjusted 95\% cluster-\allowbreak{}bootstrap intervals. C1 is 1 minus direct probability slope and C2 is direct frequency-\allowbreak{}minus-\allowbreak{}probability slope; they are not status-\allowbreak{}quo endpoints or the underlying slope means. Supplemental C1/\allowbreak{}C2 values are reported separately in H3.\par\clearpage

% END M:215

% BEGIN M:219 text

\clearpage\begin{center}\includegraphics[width=\linewidth,height=.70\textheight,keepaspectratio]{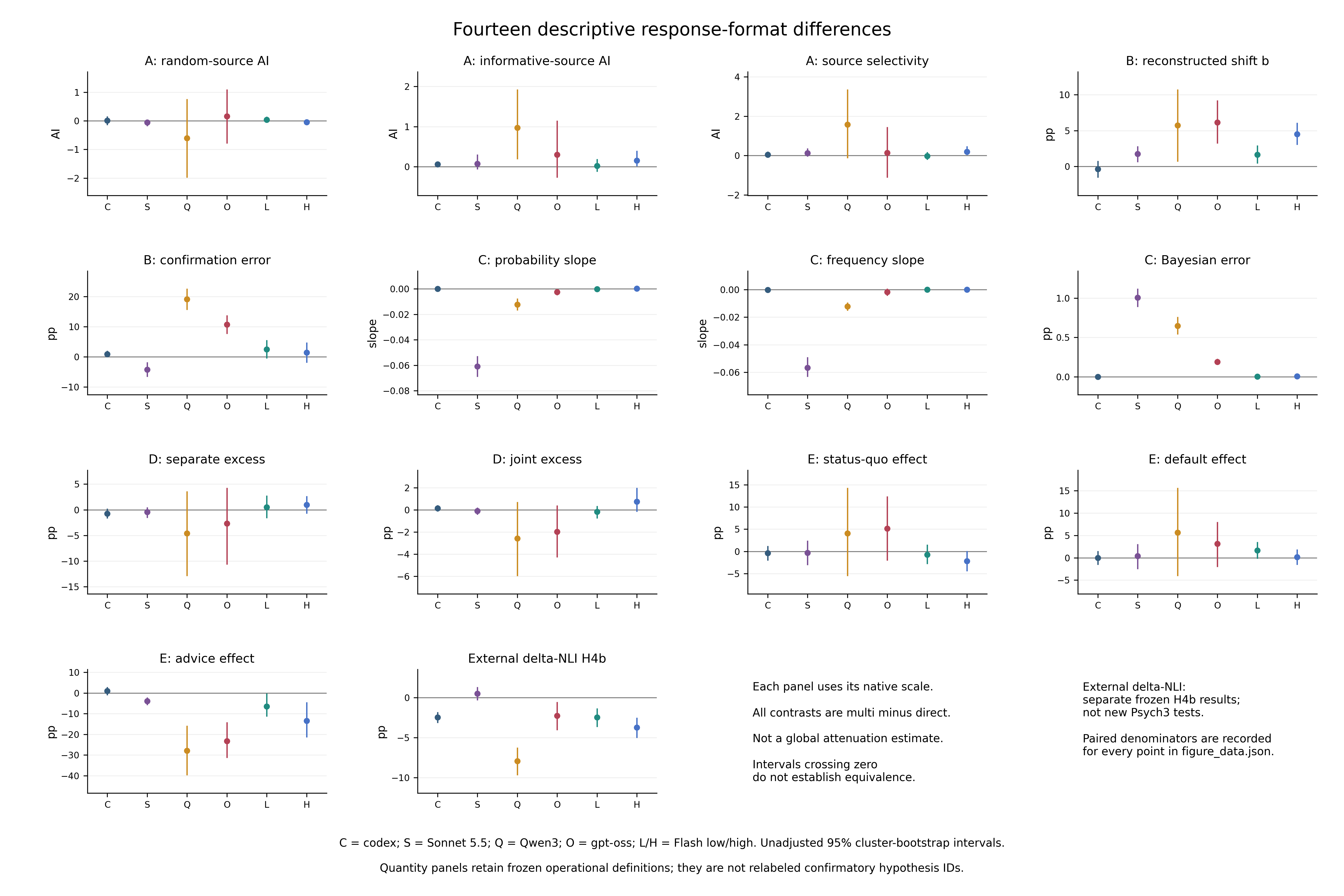}\end{center}
\textbf{Figure 3. Fourteen descriptive multi-\allowbreak{}estimate-\allowbreak{}minus-\allowbreak{}direct quantities.} Panels use the physical quantities from the frozen cross-\allowbreak{}experiment export:\allowbreak{} A's irrelevant/\allowbreak{}informative anchoring indices and selectivity; B's b and e\_\allowbreak{}PO; C's probability/\allowbreak{}frequency slopes and Bayesian error; D's separate/\allowbreak{}joint signed excess; E's status/\allowbreak{}default/\allowbreak{}advice effects; and external delta-\allowbreak{}NLI H4b. Each has its own native scale and paired subset. These quantity panels are not identical to confirmatory hypothesis IDs. The external comparison comes from a separate frozen study,\allowbreak{} with a different codex CLI version,\allowbreak{} and is not a new cross-\allowbreak{}task confirmatory test. Intervals crossing zero do not establish equivalence.\par\clearpage

% END M:219

% BEGIN M:221 text

\subsection{Flash-\allowbreak{}high execution windows}

% END M:221

% BEGIN M:223 text

The 3,\allowbreak{}049 F0 and 653 F1 units are disjoint. F0 retains one missing B710 unit; F1 has none. Both inherit one development set of 618 admitted OK units. The combined formal export retains the frozen nine-\allowbreak{}key schema (3,\allowbreak{}718 lines; 3,\allowbreak{}702 admitted; 16 excluded). Material scheduling and execution windows are associated. Cluster D /\allowbreak{} w01 spans F0 (7 r0 calls) and F1 (3 r0 calls) across both response formats. There are zero cross-\allowbreak{}batch r0/\allowbreak{}r1 repeat pairs:\allowbreak{} every repeated unit has its r0 and r1 calls in the same batch. Tables F1 and H11 expose actual frozen condition codes without invented plausible/\allowbreak{}implausible or HI/\allowbreak{}LO renaming.\par

% END M:223

% BEGIN M:225 text

F0 and F1 are non-\allowbreak{}randomized execution windows confounded with item schedule order and collection time. Differences are exploratory and do not identify a causal execution-\allowbreak{}window or model effect.\par

% END M:225

% BEGIN M:227 text

\section{Exploratory and Post-\allowbreak{}Cohort Mechanistic Analyses}

% END M:227

% BEGIN M:229 text

\subsection{Exploratory candidate spread in the six-\allowbreak{}configuration cohort}

% END M:229

% BEGIN M:231 text

Candidate-\allowbreak{}spread analysis was proposed after viewing formal results and is not preregistered. All six configurations use the original pure analyse\_\allowbreak{}items function,\allowbreak{} 5,\allowbreak{}000 cluster-\allowbreak{}bootstrap draws and seed 20261029 on admitted r0 items. The three v1 configurations reproduce their earlier exploratory families with exact JSON value equality. This manuscript-\allowbreak{}generation sandbox reads registered result JSON without opening raw attempt files or the protected original freeze; strict raw admission belongs to the audited host bundle,\allowbreak{} not a new claim by this builder.\par

% END M:231

% BEGIN M:233 text

A spread is weighted candidate log-\allowbreak{}quantity SD,\allowbreak{} while extracted estimates retain the log of the weighted arithmetic mean. B-\allowbreak{}E use probability points. D joint-\allowbreak{}call spread averages the conjunction and B component spreads,\allowbreak{} not four independent calls. Correlations aggregate within item and resample whole clusters. Undefined correlations remain undefined,\allowbreak{} including constant-\allowbreak{}input cases; zero valid resamples are not failed experiments.\par

% END M:233

% BEGIN M:235 text

Tables G1 and H7-\allowbreak{}H10 provide all six distributions,\allowbreak{} pairing denominators,\allowbreak{} spread-\allowbreak{}effect associations,\allowbreak{} A spread-\allowbreak{}error associations and D P/\allowbreak{}W strata. Within-\allowbreak{}sample median splits are descriptions,\allowbreak{} not deployment thresholds. Shared arithmetic inputs,\allowbreak{} difficulty/\allowbreak{}source composition,\allowbreak{} framing and cell availability limit interpretation. Independent task splits,\allowbreak{} targets and difficulty/\allowbreak{}scale baselines are required before claiming predictive utility.\par

% END M:235

% BEGIN M:237 text

\subsection{Post-\allowbreak{}cohort mechanistic follow-\allowbreak{}up:\allowbreak{} Scale granularity and within-\allowbreak{}call elicitation organization}\label{sec:5.2}

% END M:237

% BEGIN M:239 text

\subsubsection{Motivation,\allowbreak{} Stage 2B diagnostic pilot,\allowbreak{} and Stage 2C factorial crossover design}

% END M:239

% BEGIN M:241 text

In the primary six-\allowbreak{}configuration cohort (Section 4),\allowbreak{} Experiment C used structured numerical tables and reached a near-\allowbreak{}ceiling in direct probability mode,\allowbreak{} preventing confirmatory tests of whether multi-\allowbreak{}estimate elicitation mitigates base-\allowbreak{}rate neglect when diagnostic cues are embedded in prose narratives. Moreover,\allowbreak{} across Experiments A--E,\allowbreak{} all probabilities were elicited on the standard $0\text{--}100\%$ scale using an unguided five-\allowbreak{}estimate format (\texttt{ASK\_\allowbreak{}MULTI}) without constraining generation order or permitting repeated candidate values on coarser grids.\par

% END M:241

% BEGIN M:243 text

To examine why unguided \texttt{ASK\_\allowbreak{}MULTI} fails to uniformly debias judgments---and under what representational conditions structured elicitation reduces estimation error---we conducted a two-\allowbreak{}stage post-\allowbreak{}cohort mechanistic investigation via direct REST APIs with thinking disabled (except on the explicit reasoning benchmark \texttt{groq:\allowbreak{}gpt-\allowbreak{}oss-\allowbreak{}120b}). First,\allowbreak{} a four-\allowbreak{}block diagnostic pilot (\textbf{Stage 2B},\allowbreak{} $192 / 192$ valid units per model across \texttt{openai:\allowbreak{}gpt-\allowbreak{}4o-\allowbreak{}mini} and \texttt{sf:\allowbreak{}qwen3-\allowbreak{}8b}) varied within-\allowbreak{}call generation order,\allowbreak{} candidate-\allowbreak{}repeat constraints,\allowbreak{} and scale endpoints ($M \in \{7, 13, 14, 15, 23, 100\}$). Stage 2B identified three specific failure modes of unguided \texttt{ASK\_\allowbreak{}MULTI}:\allowbreak{}
1. \textbf{Unordered Multi-\allowbreak{}Candidate Degradation versus Point-\allowbreak{}First Stabilization (\texttt{Block A})}:\allowbreak{} In narrative Bayesian problems ($n=4$ per cell),\allowbreak{} unguided \texttt{multi\_\allowbreak{}free} increased mean absolute error (MAE) relative to \texttt{direct} on \texttt{openai:\allowbreak{}gpt-\allowbreak{}4o-\allowbreak{}mini} from $10.94\text{ pp}$ to $19.32\text{ pp}$,\allowbreak{} whereas committing to a single point estimate on Line 1 before generating five ordered candidates reduced MAE to $7.34\text{ pp}$ (\texttt{1p5\_\allowbreak{}asc}) and $6.31\text{ pp}$ (\texttt{1p5\_\allowbreak{}desc}),\allowbreak{} and explicit posterior focus reduced MAE to $6.37\text{ pp}$. On \texttt{sf:\allowbreak{}qwen3-\allowbreak{}8b},\allowbreak{} \texttt{1p5\_\allowbreak{}asc} similarly reduced narrative Bayesian MAE from $29.94\text{ pp}$ (\texttt{direct}) and $18.28\text{ pp}$ (\texttt{multi\_\allowbreak{}free}) to $14.76\text{ pp}$.
2. \textbf{Coarse-\allowbreak{}Grid Repulsion and Non-\allowbreak{}Decimal Sweet Spots (\texttt{Blocks B \& D})}:\allowbreak{} On an 8-\allowbreak{}point integer grid ($M=7$,\allowbreak{} integers $\{0, \dots, 7\}$,\allowbreak{} $n=6$ per cell in Block D),\allowbreak{} requiring five distinct candidates (\texttt{multi}) inflated MAE from $6.09\text{ pp}$ (\texttt{direct}) to $14.14\text{ pp}$ on \texttt{openai:\allowbreak{}gpt-\allowbreak{}4o-\allowbreak{}mini} ($3.71\text{ pp}$ to $7.33\text{ pp}$ on the corresponding 8-\allowbreak{}point percentage grid),\allowbreak{} while on \texttt{sf:\allowbreak{}qwen3-\allowbreak{}8b} ($n=4$ in Block B),\allowbreak{} allowing repeated grid values (\texttt{multi\_\allowbreak{}repeat}) reduced $M=7$ MAE from $27.65\text{ pp}$ (\texttt{multi}) to $21.73\text{ pp}$ ($8.61\text{ pp}$ to $6.99\text{ pp}$ on the Block D percentage grid). Meanwhile,\allowbreak{} on scale $M=15$ (Block B,\allowbreak{} $n=4$),\allowbreak{} \texttt{sf:\allowbreak{}qwen3-\allowbreak{}8b} spontaneously used \texttt{.0}/\allowbreak{}\texttt{.5} half-\allowbreak{}integer steps under \texttt{multi},\allowbreak{} lowering MAE from $18.46\text{ pp}$ (\texttt{direct}) to $5.14\text{ pp}$.
3. \textbf{Scale-\allowbreak{}Associated Anchor-\allowbreak{}Shift Reduction (\texttt{Block C})}:\allowbreak{} On two-\allowbreak{}group mixture problems with irrelevant high/\allowbreak{}low anchors ($n=4$ per cell),\allowbreak{} replacing $M=100$ with prime scale $M=13$ reduced the high-\allowbreak{}minus-\allowbreak{}low multi-\allowbreak{}estimate anchoring shift on \texttt{openai:\allowbreak{}gpt-\allowbreak{}4o-\allowbreak{}mini} from $+3.27\text{ pp}$ (\texttt{M100/\allowbreak{}multi}) to $+0.87\text{ pp}$ (\texttt{M13/\allowbreak{}multi}),\allowbreak{} while multi-\allowbreak{}estimate elicitation on $M=13$ lowered anchored MAE relative to \texttt{M13/\allowbreak{}direct} on both \texttt{openai:\allowbreak{}gpt-\allowbreak{}4o-\allowbreak{}mini} ($5.14\text{ pp}$ vs. $11.08\text{ pp}$ under high anchors; $9.70\text{ pp}$ vs. $15.62\text{ pp}$ under low anchors) and \texttt{sf:\allowbreak{}qwen3-\allowbreak{}8b} ($6.36\text{ pp}$ vs. $11.06\text{ pp}$ under high anchors; $5.19\text{ pp}$ vs. $10.87\text{ pp}$ under low anchors).\par

% END M:243

% BEGIN M:248 text

To test whether these observed patterns generalize when scale endpoints and elicitation organization are crossed orthogonally,\allowbreak{} we executed a full \textbf{$5 \times 4 \times 2$ factorial crossover experiment (Stage 2C,\allowbreak{} $280$ units per model; $1,396 / 1,400 = 99.7\%$ valid units across five model families)} crossing:\allowbreak{}
-\allowbreak{} \textbf{Five Bounded Response-\allowbreak{}Scale Ranges ($[0, M]$ with $M \in \{7, 13, 15, 23, 100\}$)}:\allowbreak{} unlike Stage 2B Blocks B and D (which enforced discrete integer grid points),\allowbreak{} Stage 2C prompts specify bounded coordinate ranges $[0, M]$ with decimals permitted (\texttt{Decimals are allowed.}),\allowbreak{} comparing a compact endpoint ($M=7$),\allowbreak{} two prime endpoints ($M=13, 23$),\allowbreak{} a composite non-\allowbreak{}decimal endpoint ($M=15$),\allowbreak{} and the decimal percentage range ($M=100$). All estimates on $[0, M]$ are linearly rescaled to percentage points ($\hat{y}_{\text{pct}} = 100 \cdot \hat{y}_M / M$) for uniform evaluation.
-\allowbreak{} \textbf{Four Elicitation Organizations (\texttt{org})}:\allowbreak{}
  1. \texttt{single}:\allowbreak{} A single direct estimate (\texttt{estimate:\allowbreak{} <0..M>}).
  2. \texttt{multi\_\allowbreak{}repeat}:\allowbreak{} Five weighted candidates on $[0, M]$ summing to weight $1$,\allowbreak{} explicitly permitting repeated candidate values so compact scales are not penalized by distinct-\allowbreak{}point repulsion.
  3. \texttt{1p5\_\allowbreak{}asc\_\allowbreak{}repeat}:\allowbreak{} A point estimate followed by five non-\allowbreak{}decreasing,\allowbreak{} repeat-\allowbreak{}allowed weighted candidates on [0,\allowbreak{}M]; evaluate the five-\allowbreak{}candidate weighted mean. Loose ascending compliance is 329/\allowbreak{}350 (94.0\%); strict five-\allowbreak{}candidate ascending compliance is 324/\allowbreak{}350 (92.6\%),\allowbreak{} including Nemotron 65/\allowbreak{}70. Noncompliant responses remain unsorted and are retained.
  4. \texttt{posterior\_\allowbreak{}focus}:\allowbreak{} An explicit three-\allowbreak{}line belief-\allowbreak{}updating scaffold (\texttt{prior\_\allowbreak{}anchor},\allowbreak{} \texttt{evidence\_\allowbreak{}shift},\allowbreak{} \texttt{final\_\allowbreak{}estimate}) paired with an instruction not to copy raw background percentages as the final estimate; responses that end before emitting an explicit \texttt{final\_\allowbreak{}estimate:\allowbreak{}} line are marked invalid (\texttt{no\_\allowbreak{}final\_\allowbreak{}estimate\_\allowbreak{}label}).
-\allowbreak{} \textbf{Two Objective Ground-\allowbreak{}Truth Task Families ($14$ items per scale $\times$ organization cell)}:\allowbreak{}
  1. \emph{Task 1 --- Narrative Bayesian Updating (\texttt{bayes\_\allowbreak{}narrative},\allowbreak{} $8$ items)}:\allowbreak{} Prose scenarios spanning four prior/\allowbreak{}likelihood-\allowbreak{}ratio regimes (\texttt{low\_\allowbreak{}moderate},\allowbreak{} \texttt{verylow\_\allowbreak{}strong},\allowbreak{} \texttt{high\_\allowbreak{}reverse},\allowbreak{} \texttt{middle\_\allowbreak{}mixed}),\allowbreak{} evaluated by mean absolute error ($\text{MAE} = |\hat{y}_{\text{pct}} - p_{\text{Bayes}}|$) and \textbf{Surface Input-\allowbreak{}Number Echo Rate (\texttt{echo\_\allowbreak{}rate})},\allowbreak{} defined as the proportion of responses whose rescaled percentage (within $\pm 0.75\text{ pp}$) or raw scale value (within $\pm 0.25$ units on $M \ne 100$) reproduces one of the three raw percentages in the narrative (\texttt{prior},\allowbreak{} \texttt{hit\_\allowbreak{}rate},\allowbreak{} or \texttt{false\_\allowbreak{}alarm\_\allowbreak{}rate}).
  2. \emph{Task 3 --- Matched Positive/\allowbreak{}Negative Attribute Framing (\texttt{attribute\_\allowbreak{}framing},\allowbreak{} $3$ pairs $\times 2$ frames $= 6$ items)}:\allowbreak{} Two-\allowbreak{}group weighted mixtures ($60\%$ Group A,\allowbreak{} $40\%$ Group B) with non-\allowbreak{}round normative ground truths $y^* \in \{43\%, 46\%, 57\%\}$ \citep{levin1998all}. In each pair,\allowbreak{} the positive frame (\texttt{pos}) states group success rates $(a\%, b\%)$ and the negative frame (\texttt{neg}) states identical group failure rates $((100-a)\%, (100-b)\%)$,\allowbreak{} while \textbf{both frames ask for the combined probability of the positive outcome}. This design simultaneously measures \textbf{Objective Error ($\text{MAE} = |\hat{y}_{\text{pct}} - y^*|$)} and \textbf{Framing Asymmetry ($\text{FA} = |\hat{y}_{\text{pos}} - \hat{y}_{\text{neg}}|$)} on identical normative targets.\par

% END M:248

% BEGIN M:259 text

\subsubsection{Empirical findings across five model families ($1,396 / 1,400 = 99.7\%$ valid)}

% END M:259

% BEGIN M:261 text

We evaluated five diverse model families spanning three capability regimes:\allowbreak{} (i) an explicit reasoning model (\texttt{groq:\allowbreak{}gpt-\allowbreak{}oss-\allowbreak{}120b},\allowbreak{} configured for OpenAI GPT-\allowbreak{}OSS 120B with low reasoning effort and automatic HTTP-\allowbreak{}429 failover to Cloudflare Workers AI \texttt{@cf/\allowbreak{}openai/\allowbreak{}gpt-\allowbreak{}oss-\allowbreak{}120b} in \texttt{run\_\allowbreak{}stage2c\_\allowbreak{}suite.py}; because per-\allowbreak{}unit backend routing and provider-\allowbreak{}side quantization were not logged in the JSONL output,\allowbreak{} we treat this configuration as a composite API endpoint rather than claiming verified checkpoint equivalence across providers),\allowbreak{} (ii) two frontier non-\allowbreak{}thinking MoE/\allowbreak{}compact models (\texttt{aistudio:\allowbreak{}gemma-\allowbreak{}4-\allowbreak{}26b-\allowbreak{}a4b-\allowbreak{}it},\allowbreak{} Google Gemma 4 26B-\allowbreak{}A4B MoE; and \texttt{openai:\allowbreak{}gpt-\allowbreak{}4o-\allowbreak{}mini}),\allowbreak{} and (iii) two open-\allowbreak{}weight non-\allowbreak{}thinking models spanning 8B to 120B parameters (\texttt{sf:\allowbreak{}qwen3-\allowbreak{}8b},\allowbreak{} Alibaba Qwen 3 8B; and \texttt{nv:\allowbreak{}nemotron-\allowbreak{}3.5-\allowbreak{}super-\allowbreak{}vl-\allowbreak{}120b},\allowbreak{} NVIDIA Nemotron 3.5 120B-\allowbreak{}A12B MoE). All four non-\allowbreak{}thinking models achieved $280 / 280$ ($100.0\%$) valid parses,\allowbreak{} and \texttt{groq:\allowbreak{}gpt-\allowbreak{}oss-\allowbreak{}120b} achieved $276 / 280$ ($98.6\%$) valid parses ($4$ \texttt{posterior\_\allowbreak{}focus} trials on non-\allowbreak{}decimal scales $M \in \{13, 15, 23\}$ ended mid-\allowbreak{}calculation during verbose fraction arithmetic without emitting a \texttt{final\_\allowbreak{}estimate:\allowbreak{}} line; because \texttt{finish\_\allowbreak{}reason} and per-\allowbreak{}request token usage were not logged in the JSONL records,\allowbreak{} these $4$ incomplete responses are marked invalid),\allowbreak{} yielding $1,396 / 1,400$ ($99.7\%$) valid units overall (Table 3,\allowbreak{} Figure 4).\par

% END M:261

% BEGIN M:265 text

\clearpage\begin{center}\includegraphics[width=\linewidth,height=.70\textheight,keepaspectratio]{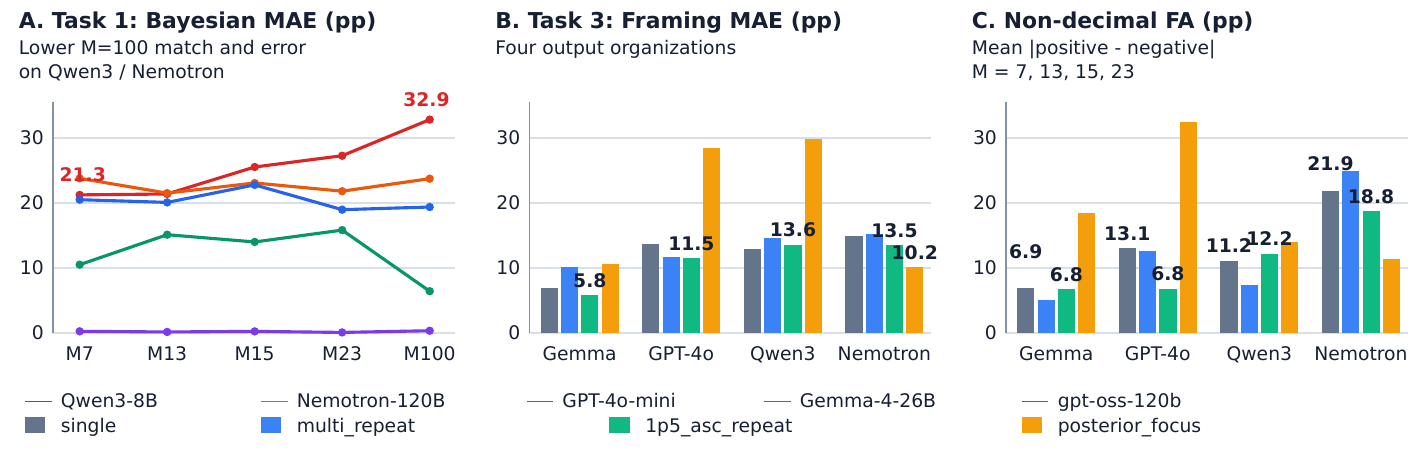}\end{center}
\textbf{Figure 4. Stage 2C $5 \times 4 \times 2$ factorial crossover across five model families ($1,396 / 1,400$ valid units).} Panel A shows Narrative Bayesian Updating MAE (\texttt{pp},\allowbreak{} Task 1,\allowbreak{} $n=8$ items/\allowbreak{}cell) and surface percentage-\allowbreak{}token echo rates across response scales $M \in \{7, 13, 15, 23, 100\}$. Panel B shows Attribute Framing objective MAE (\texttt{pp},\allowbreak{} Task 3,\allowbreak{} $n=6$ items/\allowbreak{}cell) across all four elicitation organizations. Panel C shows mean Positive/\allowbreak{}Negative Framing Asymmetry ($\text{FA} = |\hat{y}_{\text{pos}} - \hat{y}_{\text{neg}}|$,\allowbreak{} \texttt{pp}) averaged across the four non-\allowbreak{}decimal scales $M \in \{7, 13, 15, 23\}$.\par\clearpage

% END M:265

% BEGIN M:267 text

\textbf{Table 3. Stage 2C marginal performance across response scales ($M$) and elicitation organizations (\texttt{org}) for all five model families ($1,396 / 1,400$ valid units)}\par

% END M:267

% BEGIN M:269 text

Each scale row pools all valid units across the 4 organizations ($n=32$ units for Task 1 MAE,\allowbreak{} $n=24$ for Task 3 MAE,\allowbreak{} $n=56$ across Task 1 + Task 3 for overall \texttt{echo\_\allowbreak{}rate}; for \texttt{groq:\allowbreak{}gpt-\allowbreak{}oss-\allowbreak{}120b},\allowbreak{} $n=31$ Task 1 on $M \in \{13, 15, 23\}$ and $n=23$ Task 3 on $M=23$). Each organization row pools all valid units across the 5 scales ($n=40$ for Task 1 MAE,\allowbreak{} $n=30$ for Task 3 MAE,\allowbreak{} $n=12$ matched pairs across the 4 non-\allowbreak{}decimal scales for Task 3 Non-\allowbreak{}Decimal $\text{FA}$,\allowbreak{} and $n=70$ across Task 1 + Task 3 for overall \texttt{echo\_\allowbreak{}rate}; for \texttt{groq:\allowbreak{}gpt-\allowbreak{}oss-\allowbreak{}120b} \texttt{posterior\_\allowbreak{}focus},\allowbreak{} $n=37$ Task 1,\allowbreak{} $n=29$ Task 3,\allowbreak{} $n=11$ non-\allowbreak{}decimal $\text{FA}$ pairs,\allowbreak{} and $n=66$ overall). Bold indicates the lowest (best) value within each model $\times$ factor block.\par

% END M:269

% BEGIN M:271 table

\begingroup\scriptsize\setlength{\tabcolsep}{1.3mm}\setlength{\LTpre}{4pt}\setlength{\LTpost}{6pt}
\begin{longtable}{>{\raggedright\arraybackslash}p{14.429mm}>{\raggedright\arraybackslash}p{12.910mm}>{\raggedright\arraybackslash}p{25.820mm}>{\raggedright\arraybackslash}p{25.820mm}>{\raggedright\arraybackslash}p{44.805mm}>{\raggedright\arraybackslash}p{29.617mm}}
\toprule
Model Configuration & Slice /\allowbreak{} Condition & Task 1:\allowbreak{} Bayesian Updating MAE (pp) & Task 3:\allowbreak{} Attribute Framing MAE (pp) & Task 3:\allowbreak{} Non-\allowbreak{}Decimal Mean Framing Asymmetry $\text{FA}$ (pp) & Overall Surface Echo Rate (\texttt{echo\_\allowbreak{}rate}) \\
\midrule\endfirsthead
\toprule
Model Configuration & Slice /\allowbreak{} Condition & Task 1:\allowbreak{} Bayesian Updating MAE (pp) & Task 3:\allowbreak{} Attribute Framing MAE (pp) & Task 3:\allowbreak{} Non-\allowbreak{}Decimal Mean Framing Asymmetry $\text{FA}$ (pp) & Overall Surface Echo Rate (\texttt{echo\_\allowbreak{}rate}) \\
\midrule\endhead
\bottomrule\endfoot
\textbf{\texttt{groq:\allowbreak{}gpt-\allowbreak{}oss-\allowbreak{}120b}} & Scale $M=7$ & 0.23 & 0.15 & --- & \textbf{0.0\%} \\
\emph{(Reasoning model; 276 /\allowbreak{} 280)} & Scale $M=13$ (prime) & 0.13 & 0.19 & --- & 1.8\% \\
 & Scale $M=15$ & 0.22 & 0.04 & --- & \textbf{0.0\%} \\
 & Scale $M=23$ (prime) & \textbf{0.06} & 0.05 & --- & \textbf{0.0\%} \\
 & Scale $M=100$ (decimal) & 0.32 & \textbf{0.02} & --- & \textbf{0.0\%} \\
 & Org:\allowbreak{} \texttt{single} & 0.07 & \textbf{0.01} & \textbf{0.00} & \textbf{0.0\%} \\
 & Org:\allowbreak{} \texttt{multi\_\allowbreak{}repeat} & 0.16 & 0.12 & 0.21 & \textbf{0.0\%} \\
 & Org:\allowbreak{} \texttt{1p5\_\allowbreak{}asc\_\allowbreak{}repeat} & 0.49 & 0.13 & 0.17 & \textbf{0.0\%} \\
 & Org:\allowbreak{} \texttt{posterior\_\allowbreak{}focus} \emph{(66/\allowbreak{}70)} & \textbf{0.03} & 0.11 & 0.23 & 1.5\% \\
\textbf{\texttt{aistudio:\allowbreak{}gemma-\allowbreak{}4-\allowbreak{}26b-\allowbreak{}a4b-\allowbreak{}it}} & Scale $M=7$ & 10.51 & 13.04 & --- & 17.9\% \\
\emph{(Frontier MoE; 280 /\allowbreak{} 280)} & Scale $M=13$ (prime) & 15.13 & 8.58 & --- & 10.7\% \\
 & Scale $M=15$ & 14.01 & 8.39 & --- & 7.1\% \\
 & Scale $M=23$ (prime) & 15.84 & 6.78 & --- & 7.1\% \\
 & Scale $M=100$ (decimal) & \textbf{6.42} & \textbf{5.25} & --- & \textbf{1.8\%} \\
 & Org:\allowbreak{} \texttt{single} & 11.09 & 6.98 & 6.94 & 7.1\% \\
 & Org:\allowbreak{} \texttt{multi\_\allowbreak{}repeat} & 13.17 & 10.22 & \textbf{4.99} & 10.0\% \\
 & Org:\allowbreak{} \textbf{\texttt{1p5\_\allowbreak{}asc\_\allowbreak{}repeat}} & \textbf{9.88} & \textbf{5.84} & 6.78 & \textbf{1.4\%} \\
 & Org:\allowbreak{} \texttt{posterior\_\allowbreak{}focus} & 15.39 & 10.60 & 18.41 & 17.1\% \\
\textbf{\texttt{openai:\allowbreak{}gpt-\allowbreak{}4o-\allowbreak{}mini}} & Scale $M=7$ & 20.53 & 20.28 & --- & \textbf{3.6\%} \\
\emph{(Compact commercial; 280 /\allowbreak{} 280)} & Scale $M=13$ (prime) & 20.11 & 15.13 & --- & 7.1\% \\
 & Scale $M=15$ & 22.81 & 18.84 & --- & 14.3\% \\
 & Scale $M=23$ (prime) & \textbf{18.99} & 19.28 & --- & \textbf{3.6\%} \\
 & Scale $M=100$ (decimal) & 19.40 & \textbf{8.18} & --- & 14.3\% \\
 & Org:\allowbreak{} \texttt{single} & \textbf{15.70} & 13.71 & 13.07 & 11.4\% \\
 & Org:\allowbreak{} \texttt{multi\_\allowbreak{}repeat} & 21.07 & 11.69 & 12.58 & \textbf{5.7\%} \\
 & Org:\allowbreak{} \textbf{\texttt{1p5\_\allowbreak{}asc\_\allowbreak{}repeat}} & 19.33 & \textbf{11.49} & \textbf{6.79} & 7.1\% \\
 & Org:\allowbreak{} \texttt{posterior\_\allowbreak{}focus} & 25.36 & 28.48 & 32.55 & 10.0\% \\
\textbf{\texttt{nv:\allowbreak{}nemotron-\allowbreak{}3.5-\allowbreak{}super-\allowbreak{}vl-\allowbreak{}120b}} & Scale $M=7$ & 23.82 & 20.15 & --- & 14.3\% \\
\emph{(120B-\allowbreak{}A12B MoE non-\allowbreak{}think; 280 /\allowbreak{} 280)} & Scale $M=13$ (prime) & \textbf{21.54} & 11.29 & --- & 10.7\% \\
 & Scale $M=15$ & 23.11 & 14.47 & --- & \textbf{8.9\%} \\
 & Scale $M=23$ (prime) & 21.84 & 13.82 & --- & 10.7\% \\
 & Scale $M=100$ (decimal) & 23.77 & \textbf{7.63} & --- & 10.7\% \\
 & Org:\allowbreak{} \texttt{single} & 24.05 & 14.92 & 21.93 & 8.6\% \\
 & Org:\allowbreak{} \texttt{multi\_\allowbreak{}repeat} & 24.80 & 15.31 & 24.93 & 15.7\% \\
 & Org:\allowbreak{} \texttt{1p5\_\allowbreak{}asc\_\allowbreak{}repeat} & 25.68 & 13.51 & 18.75 & 12.9\% \\
 & Org:\allowbreak{} \textbf{\texttt{posterior\_\allowbreak{}focus}} & \textbf{16.73} & \textbf{10.15} & \textbf{11.42} & \textbf{7.1\%} \\
\textbf{\texttt{sf:\allowbreak{}qwen3-\allowbreak{}8b}} & Scale $M=7$ & \textbf{21.27} & 21.24 & --- & 8.9\% \\
\emph{(8B Dense non-\allowbreak{}think; 280 /\allowbreak{} 280)} & Scale $M=13$ (prime) & 21.41 & 14.36 & --- & 7.1\% \\
 & Scale $M=15$ & 25.57 & 20.35 & --- & 14.3\% \\
 & Scale $M=23$ (prime) & 27.31 & 22.06 & --- & \textbf{3.6\%} \\
 & Scale $M=100$ (decimal) & 32.88 & \textbf{10.72} & --- & 32.1\% \\
 & Org:\allowbreak{} \texttt{single} & 26.25 & \textbf{12.98} & 11.15 & 18.6\% \\
 & Org:\allowbreak{} \texttt{multi\_\allowbreak{}repeat} & \textbf{23.34} & 14.59 & \textbf{7.36} & 11.4\% \\
 & Org:\allowbreak{} \texttt{1p5\_\allowbreak{}asc\_\allowbreak{}repeat} & 25.66 & 13.59 & 12.22 & \textbf{10.0\%} \\
 & Org:\allowbreak{} \texttt{posterior\_\allowbreak{}focus} & 27.49 & 29.83 & 14.02 & 12.9\% \\
\end{longtable}
\endgroup

% END M:271

% BEGIN M:319 text

\textbf{1. Observable surface percentage-\allowbreak{}token matches and error.} On Qwen3-\allowbreak{}8B,\allowbreak{} M=100 is associated with high surface percentage-\allowbreak{}token match rates:\allowbreak{} 43.8\% (14/\allowbreak{}32) in narrative Bayesian trials and 32.1\% (18/\allowbreak{}56) overall,\allowbreak{} with Bayesian MAE 32.88 pp (37.3 pp under \texttt{single}). Non-\allowbreak{}decimal endpoints lower surface percentage-\allowbreak{}token match rates:\allowbreak{} Bayesian rates are 6.2\% at M=7,\allowbreak{} 9.4\% at M=13 and 0.0\% at M=23; overall rates range 3.6\%--8.9\%. Bayesian MAE is 21.27,\allowbreak{} 21.41 and 27.31 pp,\allowbreak{} improvements of 11.61,\allowbreak{} 11.47 and 5.57 pp relative to M=100. Cell-\allowbreak{}level errors reach 17.6 pp at M=7 \texttt{single} and 16.7 pp at M=13 \texttt{posterior\_\allowbreak{}focus}. Nemotron achieves 21.54 /\allowbreak{} 21.84 pp at M=13 /\allowbreak{} 23 versus 23.77 pp at M=100; GPT-\allowbreak{}4o-\allowbreak{}mini achieves 18.99 pp at M=23 versus 19.40 pp at M=100. Attribute-\allowbreak{}framing MAE,\allowbreak{} however,\allowbreak{} is lowest at M=100 for every model. A match-\allowbreak{}rate change cannot alone distinguish reduced psychological copying (a hypothesis) from changed scale-\allowbreak{}grid accessibility. P1 independently tests this distinction below.\par

% END M:319

% BEGIN M:321 text

\textbf{2. \texttt{1p5\_\allowbreak{}asc\_\allowbreak{}repeat} as a Structural Stabilizer Relative to Unguided \texttt{multi\_\allowbreak{}repeat}.}  
Across the four non-\allowbreak{}ceiling models,\allowbreak{} unguided \texttt{multi\_\allowbreak{}repeat} exhibits a trade-\allowbreak{}off between smoothing framing noise in Task 3 and suffering from candidate dispersion in Task 1. By contrast,\allowbreak{} \textbf{\texttt{1p5\_\allowbreak{}asc\_\allowbreak{}repeat}}---which commits the model to a single point estimate on Line 1 before eliciting five non-\allowbreak{}decreasing repeat-\allowbreak{}allowed candidates---outperforms \texttt{multi\_\allowbreak{}repeat} on Attribute Framing MAE across all four non-\allowbreak{}ceiling models ($5.84$ vs. $10.22\text{ pp}$ on \texttt{gemma-\allowbreak{}4-\allowbreak{}26b-\allowbreak{}a4b-\allowbreak{}it},\allowbreak{} $11.49$ vs. $11.69\text{ pp}$ on \texttt{gpt-\allowbreak{}4o-\allowbreak{}mini},\allowbreak{} $13.51$ vs. $15.31\text{ pp}$ on \texttt{nemotron-\allowbreak{}120b},\allowbreak{} and $13.59$ vs. $14.59\text{ pp}$ on \texttt{qwen3-\allowbreak{}8b}),\allowbreak{} while on Task 1 (\texttt{bayes\_\allowbreak{}narrative}) it improves over \texttt{multi\_\allowbreak{}repeat} on \texttt{gemma-\allowbreak{}4-\allowbreak{}26b-\allowbreak{}a4b-\allowbreak{}it} ($9.88$ vs. $13.17\text{ pp}$) and \texttt{gpt-\allowbreak{}4o-\allowbreak{}mini} ($19.33$ vs. $21.07\text{ pp}$,\allowbreak{} where \texttt{single} remains lowest at $15.70\text{ pp}$),\allowbreak{} whereas \texttt{multi\_\allowbreak{}repeat} stays slightly lower on \texttt{qwen3-\allowbreak{}8b} ($23.34$ vs. $25.66\text{ pp}$) and \texttt{nemotron-\allowbreak{}120b} ($24.80$ vs. $25.68\text{ pp}$):\allowbreak{}
-\allowbreak{} On \textbf{\texttt{aistudio:\allowbreak{}gemma-\allowbreak{}4-\allowbreak{}26b-\allowbreak{}a4b-\allowbreak{}it}},\allowbreak{} \texttt{1p5\_\allowbreak{}asc\_\allowbreak{}repeat} achieves the lowest marginal Bayesian Updating MAE (\textbf{$9.88\text{ pp}$} vs. $11.09\text{ pp}$ \texttt{single} and $13.17\text{ pp}$ \texttt{multi\_\allowbreak{}repeat}),\allowbreak{} lowest Attribute Framing MAE (\textbf{$5.84\text{ pp}$} vs. $6.98\text{ pp}$ \texttt{single} and $10.22\text{ pp}$ \texttt{multi\_\allowbreak{}repeat}),\allowbreak{} and lowest surface echo rate (\textbf{$1.4\%$}). At compact scale $M=7$,\allowbreak{} \texttt{1p5\_\allowbreak{}asc\_\allowbreak{}repeat} reduces Gemma-\allowbreak{}4's Bayesian MAE to \textbf{$6.5\text{ pp}$} (compared with $10.5\text{ pp}$ under \texttt{single} and $15.95\text{ pp}$ under \texttt{multi\_\allowbreak{}repeat}); on non-\allowbreak{}decimal scales $M=13, 15, 23$,\allowbreak{} it reduces Attribute Framing MAE to \textbf{$5.0\text{ pp}$},\allowbreak{} \textbf{$3.7\text{ pp}$},\allowbreak{} and \textbf{$3.8\text{ pp}$},\allowbreak{} matching the decimal $M=100$ baseline ($3.6\text{ pp}$).
-\allowbreak{} On \textbf{\texttt{openai:\allowbreak{}gpt-\allowbreak{}4o-\allowbreak{}mini}},\allowbreak{} \texttt{1p5\_\allowbreak{}asc\_\allowbreak{}repeat} achieves the lowest marginal Attribute Framing MAE (\textbf{$11.49\text{ pp}$} vs. $13.71\text{ pp}$ \texttt{single}) and \textbf{cuts Positive/\allowbreak{}Negative Framing Asymmetry ($\text{FA}$) nearly in half across the four non-\allowbreak{}decimal scales $M \in \{7, 13, 15, 23\}$},\allowbreak{} reducing mean non-\allowbreak{}decimal $\text{FA}$ from \textbf{$13.07\text{ pp}$} (\texttt{single}) and \textbf{$12.58\text{ pp}$} (\texttt{multi\_\allowbreak{}repeat}) to \textbf{$6.79\text{ pp}$} (at $M=7$,\allowbreak{} $\text{FA}$ drops from $20.95\text{ pp}$ to \textbf{$5.1\text{ pp}$},\allowbreak{} while objective MAE drops from $21.3\text{ pp}$ to \textbf{$13.2\text{ pp}$}).
-\allowbreak{} On \textbf{\texttt{nv:\allowbreak{}nemotron-\allowbreak{}3.5-\allowbreak{}super-\allowbreak{}vl-\allowbreak{}120b}},\allowbreak{} \texttt{1p5\_\allowbreak{}asc\_\allowbreak{}repeat} reduces Attribute Framing MAE from $14.92\text{ pp}$ (\texttt{single}) and $15.31\text{ pp}$ (\texttt{multi\_\allowbreak{}repeat}) to \textbf{$13.51\text{ pp}$} and reduces non-\allowbreak{}decimal Framing Asymmetry from $21.93\text{ pp}$ (\texttt{single}) and $24.93\text{ pp}$ (\texttt{multi\_\allowbreak{}repeat}) to \textbf{$18.75\text{ pp}$} (with prime scales $M=13$ and $M=23$ improving from $12.8\text{ pp}$ and $16.5\text{ pp}$ MAE under \texttt{single} to \textbf{$9.9\text{ pp}$} and \textbf{$10.7\text{ pp}$} under \texttt{1p5\_\allowbreak{}asc\_\allowbreak{}repeat}).\par

% END M:321

% BEGIN M:327 text

\textbf{3. Three-\allowbreak{}Tier Capability Spectrum and the Role of \texttt{posterior\_\allowbreak{}focus}.}  
Comparing \texttt{posterior\_\allowbreak{}focus} across all five models reveals how explicit decomposition interacts with model scale and configuration:\allowbreak{}
-\allowbreak{} \emph{Tier 1 (Reasoning Ceiling and Incomplete Decomposition Responses --- \texttt{groq:\allowbreak{}gpt-\allowbreak{}oss-\allowbreak{}120b})}:\allowbreak{} On this explicit reasoning configuration,\allowbreak{} completed outputs achieve near-\allowbreak{}zero error across \textbf{all five scales and all four organizations} on completed responses (marginal $\text{MAE} \le 0.49\text{ pp}$,\allowbreak{} cell-\allowbreak{}level $\text{MAE} \le 0.97\text{ pp}$,\allowbreak{} and non-\allowbreak{}decimal $\text{FA} \le 0.23\text{ pp}$,\allowbreak{} with \texttt{posterior\_\allowbreak{}focus} achieving $0.03\text{ pp}$ Bayesian MAE and $0.11\text{ pp}$ Attribute Framing MAE across its $66 / 70$ valid responses). On $4 / 70$ \texttt{posterior\_\allowbreak{}focus} trials on non-\allowbreak{}decimal scales ($M \in \{13, 15, 23\}$),\allowbreak{} the response ended mid-\allowbreak{}calculation during verbose fraction arithmetic without emitting a \texttt{final\_\allowbreak{}estimate:\allowbreak{}} line (consistent with output truncation,\allowbreak{} though \texttt{finish\_\allowbreak{}reason} and per-\allowbreak{}request token usage were not logged in the JSONL records).
-\allowbreak{} \emph{Tier 2 (Large MoE with Decomposition-\allowbreak{}Associated Gains --- \texttt{nv:\allowbreak{}nemotron-\allowbreak{}3.5-\allowbreak{}super-\allowbreak{}vl-\allowbreak{}120b})}:\allowbreak{} When a high-\allowbreak{}capacity 120B MoE model operates with thinking disabled,\allowbreak{} it achieves substantially lower error once prompted to decompose the update,\allowbreak{} and \textbf{\texttt{posterior\_\allowbreak{}focus} achieves the best marginal performance across every metric}:\allowbreak{} cutting Bayesian MAE from $24.05\text{ pp}$ (\texttt{single}) to \textbf{$16.73\text{ pp}$} ($12.7\text{ pp}$ at $M=7$,\allowbreak{} $14.9\text{ pp}$ at $M=13$,\allowbreak{} $15.5\text{ pp}$ at $M=23$),\allowbreak{} cutting Attribute Framing MAE from $14.92\text{ pp}$ to \textbf{$10.15\text{ pp}$},\allowbreak{} and cutting non-\allowbreak{}decimal Framing Asymmetry from $21.93\text{ pp}$ to \textbf{$11.42\text{ pp}$}.
-\allowbreak{} \emph{Tier 3 (Compact Non-\allowbreak{}Thinking Models --- \texttt{openai:\allowbreak{}gpt-\allowbreak{}4o-\allowbreak{}mini} and \texttt{sf:\allowbreak{}qwen3-\allowbreak{}8b})}:\allowbreak{} On smaller non-\allowbreak{}thinking models,\allowbreak{} \texttt{posterior\_\allowbreak{}focus} improves \textbf{non-\allowbreak{}linear Bayesian updating on prime scale $M=13$} ($23.6 \to 16.6\text{ pp}$ on \texttt{gpt-\allowbreak{}4o-\allowbreak{}mini}; $23.0 \to 16.7\text{ pp}$ on \texttt{qwen3-\allowbreak{}8b}),\allowbreak{} but induces \textbf{over-\allowbreak{}shifting on linear weighted mixtures} (Task 3),\allowbreak{} where \texttt{1p5\_\allowbreak{}asc\_\allowbreak{}repeat} provides more stable domain-\allowbreak{}general performance.\par

% END M:327

% BEGIN M:334 text

\textbf{Format validity and model heterogeneity.} Nemotron single-\allowbreak{}line collapse falls from 55/\allowbreak{}70 (78.6\%) under unordered \texttt{multi\_\allowbreak{}repeat} to 5/\allowbreak{}70 (7.1\%) under \texttt{1p5\_\allowbreak{}asc\_\allowbreak{}repeat}; this comparison concerns two multi-\allowbreak{}candidate protocols. Across five models,\allowbreak{} fewer-\allowbreak{}than-\allowbreak{}five candidate responses fall from 58/\allowbreak{}350 (16.6\%) to 6/\allowbreak{}350 (1.7\%). Pooled attribute-\allowbreak{}framing MAE is 8.91 pp for Upgraded MCW-\allowbreak{}Asc,\allowbreak{} versus Direct 9.72,\allowbreak{} Direct-\allowbreak{}CoT 15.94 (n=149),\allowbreak{} MCU 10.56 and Baseline MCW 10.38 pp. This pooled advantage does not dominate each model:\allowbreak{} Nemotron \texttt{posterior\_\allowbreak{}focus} achieves 10.15 pp framing and 16.73 pp Bayesian MAE; Qwen3-\allowbreak{}8B \texttt{single} achieves 12.98 pp framing MAE. A first-\allowbreak{}slot capture or enumeration inertia explanation remains a hypothesis,\allowbreak{} because these internal processes were not measured.\par

% END M:334

% BEGIN M:336 text

\textbf{Stage 1 Dev readouts.} On the same generated candidates,\allowbreak{} MCW has lower Bayesian error than MCU in 6/\allowbreak{}6 configurations (Sonnet 5.5:\allowbreak{} 0.88 versus 5.89 pp; codex:\allowbreak{} 0.0001 versus 1.49 pp); informed-\allowbreak{}advice absorption favors MCW in 6/\allowbreak{}6; hindsight e\_\allowbreak{}PO favors MCW in 5/\allowbreak{}6. These readout comparisons do not isolate prompt-\allowbreak{}level weight elicitation. P2-\allowbreak{}A supplies that separate intervention.\par

% END M:336

% BEGIN M:338 text

\clearpage\begin{center}\includegraphics[width=\linewidth,height=.68\textheight,keepaspectratio]{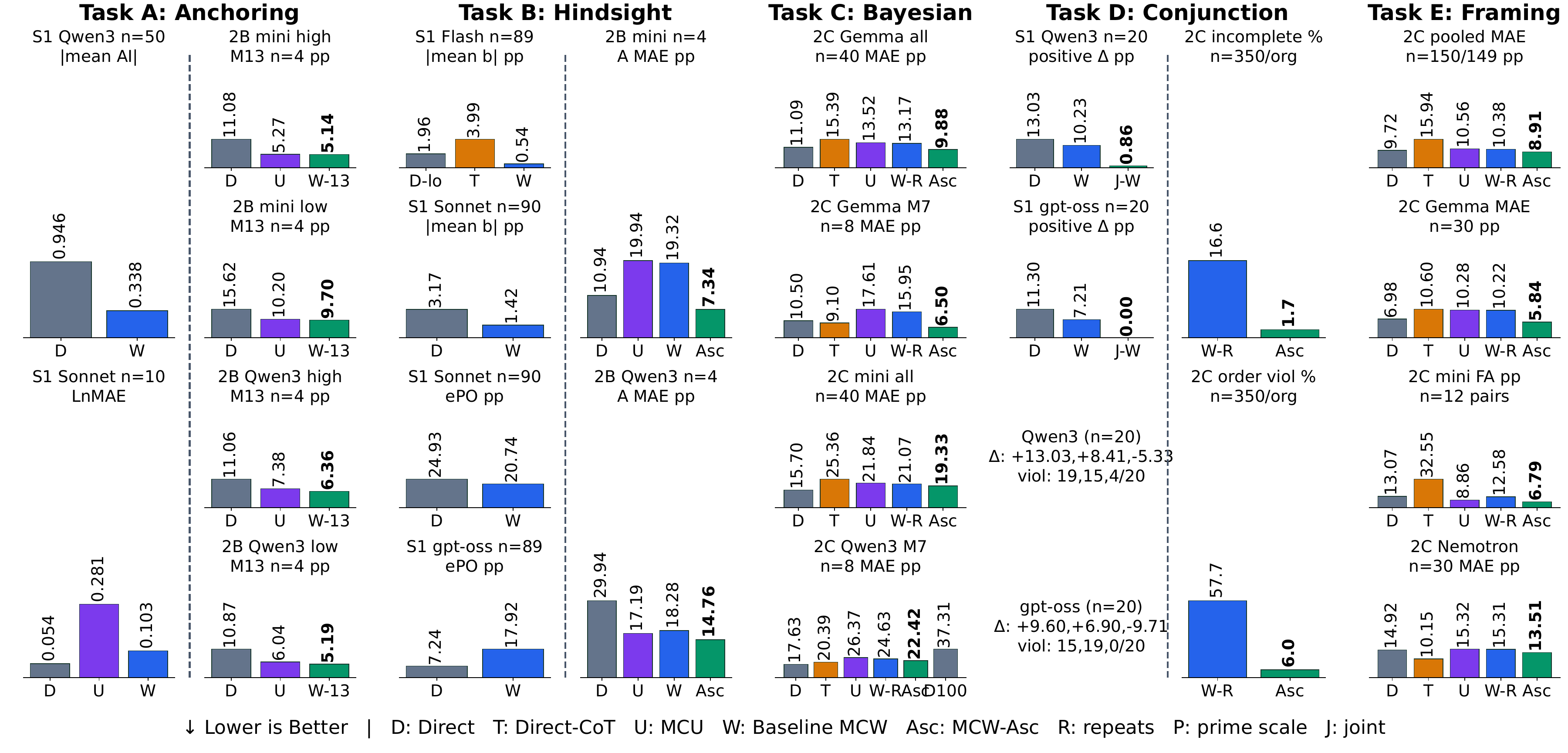}\end{center}
Figure 5. Task-\allowbreak{}specific descriptive comparisons across Stage 1 Formal,\allowbreak{} Stage 1 Dev,\allowbreak{} Stage 2B,\allowbreak{} and Stage 2C slices (all panels:\allowbreak{} lower is better). Subpanels retain distinct vertical scales,\allowbreak{} units (\texttt{|mean AI|},\allowbreak{} \texttt{LnMAE},\allowbreak{} \texttt{pp},\allowbreak{} \texttt{\%}),\allowbreak{} and denominators separated by dashed boundaries; missing bars indicate combinations not run in that slice rather than zero or imputed values. Protocol labels map to concrete stage-\allowbreak{}specific arms in \texttt{story\_\allowbreak{}figures\_\allowbreak{}ledger.json}:\allowbreak{} (1) in the Stage 1 Formal Flash hindsight slice (\texttt{D-\allowbreak{}lo /\allowbreak{} T /\allowbreak{} W},\allowbreak{} paired $n=89$,\allowbreak{} $|\text{mean } b|$ in \texttt{pp}),\allowbreak{} the bar plotted under legend slot \texttt{T} (\texttt{Direct-\allowbreak{}CoT}) represents the frozen Flash-\allowbreak{}high direct reasoning-\allowbreak{}budget configuration ($3.99\text{ pp}$,\allowbreak{} contrasted against Flash-\allowbreak{}low direct \texttt{D-\allowbreak{}lo} at $1.96\text{ pp}$ and Flash-\allowbreak{}high multi \texttt{W} at $0.54\text{ pp}$),\allowbreak{} which is a reasoning-\allowbreak{}budget configuration contrast rather than an explicit chain-\allowbreak{}of-\allowbreak{}thought prompt intervention; (2) the bottom two panels in row B (\texttt{Stage 2B A},\allowbreak{} $n=4$/\allowbreak{}arm) are narrative Bayesian prior-\allowbreak{}anchor proxies rather than ProbCOPA hindsight reconstruction trials,\allowbreak{} and the Stage 2C panels in row D (\texttt{incomplete \%} and \texttt{order violation \%},\allowbreak{} $n=350$/\allowbreak{}org) are multi-\allowbreak{}candidate structural format diagnostics rather than conjunction probability trials; (3) every \texttt{MCU} (\texttt{V2}) bar shown here (Stage 1 Dev \texttt{A\_\allowbreak{}ln} $n=10$,\allowbreak{} Stage 2B \texttt{multi\_\allowbreak{}free}/\allowbreak{}\texttt{multi},\allowbreak{} and Stage 2C \texttt{multi\_\allowbreak{}repeat}) is a same-\allowbreak{}generation equal-\allowbreak{}weight readout computed from the corresponding weighted multi-\allowbreak{}candidate output,\allowbreak{} not an independent unweighted prompt arm (prompt-\allowbreak{}level weight elicitation is isolated separately in post-\allowbreak{}cohort Suite P2-\allowbreak{}A,\allowbreak{} Table 5); and (4) \texttt{Baseline MCW} denotes Stage 1 \texttt{ASK\_\allowbreak{}MULTI},\allowbreak{} Stage 2B distinct-\allowbreak{}candidate \texttt{multi\_\allowbreak{}free}/\allowbreak{}\texttt{multi},\allowbreak{} or Stage 2C repeat-\allowbreak{}allowed \texttt{multi\_\allowbreak{}repeat},\allowbreak{} whereas upgraded green bars denote slice-\allowbreak{}specific variants---Stage 1 D joint presentation (\texttt{Joint + MCW}),\allowbreak{} Stage 2B point-\allowbreak{}first ascending without repeat (\texttt{1p5\_\allowbreak{}asc}),\allowbreak{} Stage 2B prime-\allowbreak{}scale distinct multi (\texttt{M=13 multi}),\allowbreak{} or Stage 2C combined point-\allowbreak{}first ascending with repeats (\texttt{1p5\_\allowbreak{}asc\_\allowbreak{}repeat})---not a single uniform protocol across stages.\par\clearpage

% END M:338

% BEGIN M:340 text

\clearpage\begin{center}\includegraphics[width=\linewidth,height=.68\textheight,keepaspectratio]{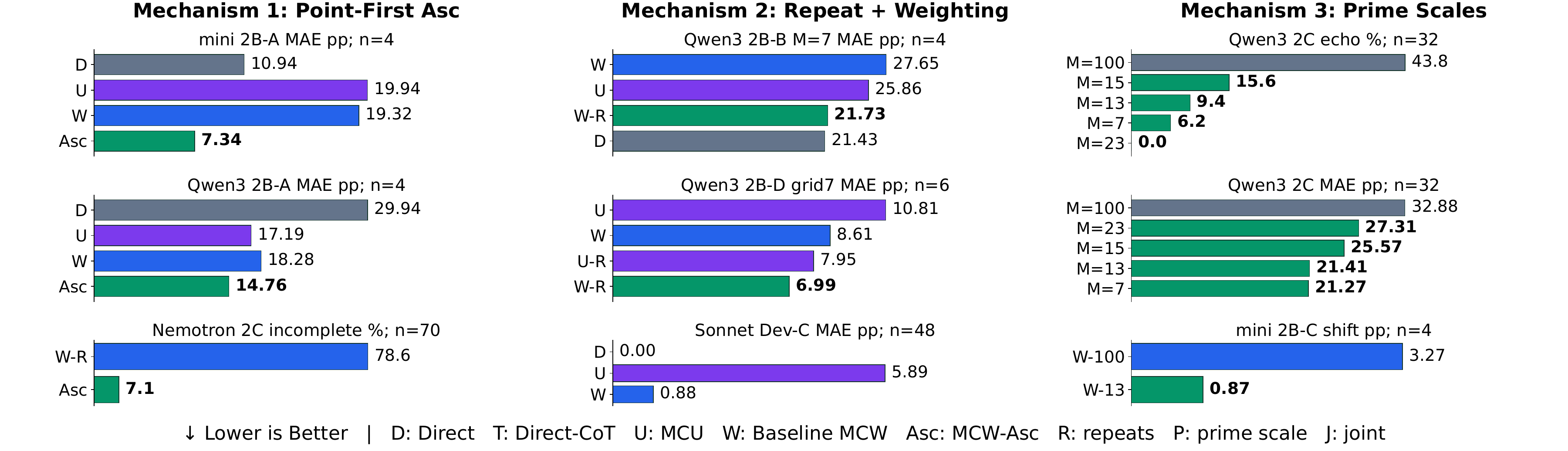}\end{center}
Figure 6. Exploratory protocol-\allowbreak{}manipulation comparisons across ordering (left),\allowbreak{} repeat allowance and weighting readouts (middle),\allowbreak{} and non-\allowbreak{}decimal response scales (right; all panels:\allowbreak{} lower is better). The embedded panel banners ``Mechanism 1--3'' are descriptive,\allowbreak{} hypothesis-\allowbreak{}generating groupings of observable prompt manipulations and do not identify internal cognitive,\allowbreak{} attentional,\allowbreak{} or neural causal mechanisms,\allowbreak{} which remain unmeasured hypotheses (Section 6). Each subpanel retains its own stage,\allowbreak{} model,\allowbreak{} sample size ($n=4$ or $6$ in Stage 2B; $n=32$ or $70$ in Stage 2C; $n=48$ in Stage 1 Dev C),\allowbreak{} and native metric (\texttt{MAE pp},\allowbreak{} \texttt{incomplete \%},\allowbreak{} \texttt{echo \%},\allowbreak{} or anchor \texttt{shift pp}),\allowbreak{} without forming a common cross-\allowbreak{}task score. Across all panels,\allowbreak{} \texttt{MCU} (\texttt{V2}) bars are same-\allowbreak{}generation equal-\allowbreak{}weight readouts from weighted multi-\allowbreak{}candidate responses (\texttt{multi\_\allowbreak{}free},\allowbreak{} \texttt{multi},\allowbreak{} \texttt{multi\_\allowbreak{}repeat},\allowbreak{} or Stage 1 Dev \texttt{ASK\_\allowbreak{}MULTI}) rather than separate unweighted prompts; \texttt{MCW-\allowbreak{}Asc} in Stage 2B Block A denotes \texttt{1p5\_\allowbreak{}asc} (distinct candidates),\allowbreak{} whereas only Stage 2C uses the combined \texttt{1p5\_\allowbreak{}asc\_\allowbreak{}repeat} protocol.\par\clearpage

% END M:340

% BEGIN M:342 text

\subsection{Preregistered Post-\allowbreak{}Cohort Formal Verification Suites (P1 and P2)}

% END M:342

% BEGIN M:344 text

\subsubsection{Suite P1:\allowbreak{} Orthogonal PF \ensuremath{\times} ASC and Scale \ensuremath{\times} Cue-\allowbreak{}Format Factorials}\label{sec:5.3.1}

% END M:344

% BEGIN M:346 text

P1 completes 1,\allowbreak{}920/\allowbreak{}1,\allowbreak{}920 terminal units across \texttt{sf:\allowbreak{}qwen3-\allowbreak{}8b},\allowbreak{} \texttt{aistudio:\allowbreak{}gemma-\allowbreak{}4-\allowbreak{}26b-\allowbreak{}a4b-\allowbreak{}it},\allowbreak{} \texttt{nv:\allowbreak{}nemotron-\allowbreak{}3.5-\allowbreak{}super-\allowbreak{}vl-\allowbreak{}120b} and \texttt{xf:\allowbreak{}spark-\allowbreak{}x2.6},\allowbreak{} R=2,\allowbreak{} B=10,\allowbreak{}000,\allowbreak{} seed=20261110,\allowbreak{} with 0\% final missing/\allowbreak{}truncated outcomes. Terminal completion is distinct from first-\allowbreak{}attempt parsing. Block A crosses point-\allowbreak{}first (PF) and ascending order (ASC) in A00/\allowbreak{}A01/\allowbreak{}A10/\allowbreak{}A11:\allowbreak{} 4 models \ensuremath{\times} 4 arms \ensuremath{\times} 72 units = 1,\allowbreak{}152 units. Each arm contains 12 Bayesian clusters (24 units) and 12 paired-\allowbreak{}framing clusters (48 units). A\_\allowbreak{}Bayesian measures MAE (SESOI 1.5 pp); A\_\allowbreak{}Framing measures FrameGap,\allowbreak{} positive/\allowbreak{}negative polarity asymmetry (SESOI 2.0 pp). All six pooled tests have p\_\allowbreak{}Holm=1.0000. Model-\allowbreak{}specific secondary estimates include Gemma Bayesian Delta\_\allowbreak{}A -\allowbreak{}0.76 pp (p=0.0012),\allowbreak{} Nemotron FrameGap Delta\_\allowbreak{}A -\allowbreak{}0.21 pp (p=0.0132),\allowbreak{} both below their SESOIs; Nemotron Bayesian interaction +2.90 pp (p=0.0270); and Qwen FrameGap Delta\_\allowbreak{}A +4.31 pp (p=0.0012).\par

% END M:346

% BEGIN M:348 text

Block B crosses M100/\allowbreak{}M13/\allowbreak{}M15 with pct/\allowbreak{}freq:\allowbreak{} 4 models \ensuremath{\times} 6 cells \ensuremath{\times} 32 units = 768 units,\allowbreak{} using 16 Bayesian clusters. All P1 intervals in Table 4 are \textbf{95\% simultaneous CIs},\allowbreak{} JSON \texttt{ci95\_\allowbreak{}simultaneous},\allowbreak{} constructed by single-\allowbreak{}step Bonferroni percentile bootstrap at alpha/\allowbreak{}(2k),\allowbreak{} with k=3 or 6 within each family. Models are equally weighted; inference resamples paired scenario clusters,\allowbreak{} retaining replicates within their clusters.\par

% END M:348

% BEGIN M:350 text

\textbf{Table 4. P1 pooled primary and selected model-\allowbreak{}specific secondary effects.} Each label binds family,\allowbreak{} endpoint,\allowbreak{} population and contrast to the summary JSON. Delta\_\allowbreak{}P and Delta\_\allowbreak{}A are factorial main effects; interaction is the difference of differences. Delta\_\allowbreak{}13\_\allowbreak{}100\_\allowbreak{}pct/\allowbreak{}freq and Delta\_\allowbreak{}15\_\allowbreak{}100\_\allowbreak{}pct/\allowbreak{}freq compare scales within cue format; DiD compares the scale effect across cue formats. Negative errors favor the first condition.\par

% END M:350

% BEGIN M:352 table

\begingroup\scriptsize\setlength{\tabcolsep}{1.3mm}\setlength{\LTpre}{4pt}\setlength{\LTpost}{6pt}
\begin{longtable}{>{\raggedright\arraybackslash}p{45.541mm}>{\raggedright\arraybackslash}p{20.373mm}>{\raggedright\arraybackslash}p{20.373mm}>{\raggedright\arraybackslash}p{19.175mm}>{\raggedright\arraybackslash}p{36.433mm}>{\raggedright\arraybackslash}p{11.505mm}}
\toprule
Population /\allowbreak{} family & Endpoint & Contrast & Delta (pp) & 95\% simultaneous CI & p\_\allowbreak{}Holm \\
\midrule\endfirsthead
\toprule
Population /\allowbreak{} family & Endpoint & Contrast & Delta (pp) & 95\% simultaneous CI & p\_\allowbreak{}Holm \\
\midrule\endhead
\bottomrule\endfoot
A\_\allowbreak{}Bayesian; four\_\allowbreak{}models\_\allowbreak{}equal & MAE & Delta\_\allowbreak{}P & -\allowbreak{}0.17 & [-\allowbreak{}1.629,\allowbreak{} +1.357] & 1.0000 \\
A\_\allowbreak{}Bayesian; four\_\allowbreak{}models\_\allowbreak{}equal & MAE & Delta\_\allowbreak{}A & +0.51 & [-\allowbreak{}0.751,\allowbreak{} +1.949] & 1.0000 \\
A\_\allowbreak{}Bayesian; four\_\allowbreak{}models\_\allowbreak{}equal & MAE & interaction & -\allowbreak{}0.04 & [-\allowbreak{}2.872,\allowbreak{} +2.173] & 1.0000 \\
A\_\allowbreak{}Bayesian; aistudio:\allowbreak{}gemma-\allowbreak{}4-\allowbreak{}26b-\allowbreak{}a4b-\allowbreak{}it & MAE & Delta\_\allowbreak{}A & -\allowbreak{}0.76 & [-\allowbreak{}1.640,\allowbreak{} -\allowbreak{}0.129] & 0.0012 \\
A\_\allowbreak{}Bayesian; nv:\allowbreak{}nemotron-\allowbreak{}3.5-\allowbreak{}super-\allowbreak{}vl-\allowbreak{}120b & MAE & interaction & +2.90 & [+0.233,\allowbreak{} +6.093] & 0.0270 \\
A\_\allowbreak{}Framing; four\_\allowbreak{}models\_\allowbreak{}equal & FrameGap & Delta\_\allowbreak{}P & +0.93 & [-\allowbreak{}0.801,\allowbreak{} +3.987] & 1.0000 \\
A\_\allowbreak{}Framing; four\_\allowbreak{}models\_\allowbreak{}equal & FrameGap & Delta\_\allowbreak{}A & +0.68 & [-\allowbreak{}1.389,\allowbreak{} +2.213] & 1.0000 \\
A\_\allowbreak{}Framing; four\_\allowbreak{}models\_\allowbreak{}equal & FrameGap & interaction & -\allowbreak{}0.51 & [-\allowbreak{}1.811,\allowbreak{} +0.823] & 1.0000 \\
A\_\allowbreak{}Framing; sf:\allowbreak{}qwen3-\allowbreak{}8b & FrameGap & Delta\_\allowbreak{}A & +4.31 & [+1.216,\allowbreak{} +8.245] & 0.0012 \\
A\_\allowbreak{}Framing; nv:\allowbreak{}nemotron-\allowbreak{}3.5-\allowbreak{}super-\allowbreak{}vl-\allowbreak{}120b & FrameGap & Delta\_\allowbreak{}A & -\allowbreak{}0.21 & [-\allowbreak{}0.452,\allowbreak{} -\allowbreak{}0.029] & 0.0132 \\
B\_\allowbreak{}MAE; four\_\allowbreak{}models\_\allowbreak{}equal & MAE & Delta\_\allowbreak{}13\_\allowbreak{}100\_\allowbreak{}pct & +6.83 & [+4.289,\allowbreak{} +9.232] & 0.0012 \\
B\_\allowbreak{}MAE; four\_\allowbreak{}models\_\allowbreak{}equal & MAE & Delta\_\allowbreak{}15\_\allowbreak{}100\_\allowbreak{}pct & +6.80 & [+4.883,\allowbreak{} +8.826] & 0.0012 \\
B\_\allowbreak{}MAE; four\_\allowbreak{}models\_\allowbreak{}equal & MAE & Delta\_\allowbreak{}13\_\allowbreak{}100\_\allowbreak{}freq & +8.18 & [+5.651,\allowbreak{} +10.528] & 0.0012 \\
B\_\allowbreak{}MAE; four\_\allowbreak{}models\_\allowbreak{}equal & MAE & Delta\_\allowbreak{}15\_\allowbreak{}100\_\allowbreak{}freq & +11.50 & [+7.584,\allowbreak{} +16.066] & 0.0012 \\
B\_\allowbreak{}MAE; four\_\allowbreak{}models\_\allowbreak{}equal & MAE & DiD\_\allowbreak{}13 & +1.34 & [-\allowbreak{}1.527,\allowbreak{} +3.938] & 0.2058 \\
B\_\allowbreak{}MAE; four\_\allowbreak{}models\_\allowbreak{}equal & MAE & DiD\_\allowbreak{}15 & +4.70 & [-\allowbreak{}0.034,\allowbreak{} +10.189] & 0.0180 \\
B\_\allowbreak{}Echo\_\allowbreak{}D2; four\_\allowbreak{}models\_\allowbreak{}equal & Echo\_\allowbreak{}D2 & Delta\_\allowbreak{}13\_\allowbreak{}100\_\allowbreak{}pct & +2.34 & [-\allowbreak{}3.125,\allowbreak{} +9.375] & 1.0000 \\
B\_\allowbreak{}Echo\_\allowbreak{}D2; four\_\allowbreak{}models\_\allowbreak{}equal & Echo\_\allowbreak{}D2 & Delta\_\allowbreak{}15\_\allowbreak{}100\_\allowbreak{}pct & +1.56 & [-\allowbreak{}2.344,\allowbreak{} +5.469] & 1.0000 \\
B\_\allowbreak{}Echo\_\allowbreak{}D2; four\_\allowbreak{}models\_\allowbreak{}equal & Echo\_\allowbreak{}D2 & Delta\_\allowbreak{}13\_\allowbreak{}100\_\allowbreak{}freq & -\allowbreak{}1.56 & [-\allowbreak{}7.812,\allowbreak{} +3.906] & 1.0000 \\
B\_\allowbreak{}Echo\_\allowbreak{}D2; four\_\allowbreak{}models\_\allowbreak{}equal & Echo\_\allowbreak{}D2 & Delta\_\allowbreak{}15\_\allowbreak{}100\_\allowbreak{}freq & +0.78 & [-\allowbreak{}2.344,\allowbreak{} +4.688] & 1.0000 \\
B\_\allowbreak{}Echo\_\allowbreak{}D2; four\_\allowbreak{}models\_\allowbreak{}equal & Echo\_\allowbreak{}D2 & DiD\_\allowbreak{}13 & -\allowbreak{}3.91 & [-\allowbreak{}15.625,\allowbreak{} +5.469] & 1.0000 \\
B\_\allowbreak{}Echo\_\allowbreak{}D2; four\_\allowbreak{}models\_\allowbreak{}equal & Echo\_\allowbreak{}D2 & DiD\_\allowbreak{}15 & -\allowbreak{}0.78 & [-\allowbreak{}6.250,\allowbreak{} +4.688] & 1.0000 \\
B\_\allowbreak{}cand\_\allowbreak{}str\_\allowbreak{}copy\_\allowbreak{}any; four\_\allowbreak{}models\_\allowbreak{}equal & cand\_\allowbreak{}str\_\allowbreak{}copy\_\allowbreak{}any & Delta\_\allowbreak{}13\_\allowbreak{}100\_\allowbreak{}pct & -\allowbreak{}30.47 & [-\allowbreak{}36.201,\allowbreak{} -\allowbreak{}25.000] & 0.0012 \\
B\_\allowbreak{}cand\_\allowbreak{}str\_\allowbreak{}copy\_\allowbreak{}any; four\_\allowbreak{}models\_\allowbreak{}equal & cand\_\allowbreak{}str\_\allowbreak{}copy\_\allowbreak{}any & Delta\_\allowbreak{}15\_\allowbreak{}100\_\allowbreak{}pct & -\allowbreak{}29.69 & [-\allowbreak{}35.938,\allowbreak{} -\allowbreak{}23.438] & 0.0012 \\
B\_\allowbreak{}cand\_\allowbreak{}str\_\allowbreak{}copy\_\allowbreak{}any; four\_\allowbreak{}models\_\allowbreak{}equal & cand\_\allowbreak{}str\_\allowbreak{}copy\_\allowbreak{}any & Delta\_\allowbreak{}13\_\allowbreak{}100\_\allowbreak{}freq & -\allowbreak{}5.47 & [-\allowbreak{}13.281,\allowbreak{} -\allowbreak{}0.781] & 0.0116 \\
B\_\allowbreak{}cand\_\allowbreak{}str\_\allowbreak{}copy\_\allowbreak{}any; four\_\allowbreak{}models\_\allowbreak{}equal & cand\_\allowbreak{}str\_\allowbreak{}copy\_\allowbreak{}any & Delta\_\allowbreak{}15\_\allowbreak{}100\_\allowbreak{}freq & -\allowbreak{}5.47 & [-\allowbreak{}13.281,\allowbreak{} -\allowbreak{}0.781] & 0.0116 \\
B\_\allowbreak{}cand\_\allowbreak{}str\_\allowbreak{}copy\_\allowbreak{}any; four\_\allowbreak{}models\_\allowbreak{}equal & cand\_\allowbreak{}str\_\allowbreak{}copy\_\allowbreak{}any & DiD\_\allowbreak{}13 & +25.00 & [+17.188,\allowbreak{} +32.031] & 0.0012 \\
B\_\allowbreak{}cand\_\allowbreak{}str\_\allowbreak{}copy\_\allowbreak{}any; four\_\allowbreak{}models\_\allowbreak{}equal & cand\_\allowbreak{}str\_\allowbreak{}copy\_\allowbreak{}any & DiD\_\allowbreak{}15 & +24.22 & [+15.625,\allowbreak{} +32.031] & 0.0012 \\
\end{longtable}
\endgroup

% END M:352

% BEGIN M:383 text

The pct operational string-\allowbreak{}match rate is 40/\allowbreak{}128 = 31.25\% at M100,\allowbreak{} 1/\allowbreak{}128 = 0.78\% at M13 and 2/\allowbreak{}128 = 1.56\% at M15:\allowbreak{} reductions -\allowbreak{}30.47 and -\allowbreak{}29.69 pp,\allowbreak{} with nonzero residual matching. Under freq it is 7/\allowbreak{}128 = 5.47\% at M100 and 0/\allowbreak{}128 = 0.00\% at both smaller scales (difference -\allowbreak{}5.47 pp,\allowbreak{} p\_\allowbreak{}Holm=0.0116). The minimum formal freq cue integer is 22 > 15:\allowbreak{} legal raw outputs [0,\allowbreak{}M] structurally preclude matching at M13/\allowbreak{}M15. Thus the result is a large reduction in operational string matching,\allowbreak{} not direct evidence of a copying mechanism. The DiDs are +25.00 and +24.22 pp. In contrast,\allowbreak{} unified percentage-\allowbreak{}tolerance B\_\allowbreak{}Echo\_\allowbreak{}D2 effects range -\allowbreak{}3.91 to +2.34 pp,\allowbreak{} all p\_\allowbreak{}Holm=1.0000. Bayesian B\_\allowbreak{}MAE rises +6.83 /\allowbreak{} +6.80 pp in pct and +8.18 /\allowbreak{} +11.50 pp in freq,\allowbreak{} all p\_\allowbreak{}Holm=0.0012. The MAE DiD\_\allowbreak{}13 is +1.34 pp; DiD\_\allowbreak{}15 is +4.70 pp with simultaneous CI [-\allowbreak{}0.03,\allowbreak{} +10.19] and p\_\allowbreak{}Holm=0.0180. The interval uses fixed Bonferroni tail alpha/\allowbreak{}(2k),\allowbreak{} k=6 (\texttt{analyze\_\allowbreak{}p1\_\allowbreak{}factorial.py},\allowbreak{} line 31); Holm uses step-\allowbreak{}down rank correction of p-\allowbreak{}values. They are different constructions and need not agree about zero inclusion.\par

% END M:383

P1 Block B operational string-\allowbreak{}matching counts across response scales and cue formats (identical to Appendix I.10).\par

% BEGIN M:464 table

\begingroup\scriptsize\setlength{\tabcolsep}{1.3mm}\setlength{\LTpre}{4pt}\setlength{\LTpost}{6pt}
\begin{longtable}{>{\raggedright\arraybackslash}p{57.902mm}>{\raggedright\arraybackslash}p{20.140mm}>{\raggedright\arraybackslash}p{60.419mm}>{\raggedright\arraybackslash}p{20.140mm}}
\toprule
Measure & Cell & Matched /\allowbreak{} valid & Rate \\
\midrule\endfirsthead
\toprule
Measure & Cell & Matched /\allowbreak{} valid & Rate \\
\midrule\endhead
\bottomrule\endfoot
P1 operational matching & M100\_\allowbreak{}pct & 40/\allowbreak{}128 & 31.25\% \\
P1 operational matching & M13\_\allowbreak{}pct & 1/\allowbreak{}128 & 0.78\% \\
P1 operational matching & M15\_\allowbreak{}pct & 2/\allowbreak{}128 & 1.56\% \\
P1 operational matching & M100\_\allowbreak{}freq & 7/\allowbreak{}128 & 5.47\% \\
P1 operational matching & M13\_\allowbreak{}freq & 0/\allowbreak{}128 & 0.00\% \\
P1 operational matching & M15\_\allowbreak{}freq & 0/\allowbreak{}128 & 0.00\% \\
\end{longtable}
\endgroup

% END M:464

% BEGIN M:385 text

\subsubsection{Suite P2:\allowbreak{} Prompt-\allowbreak{}Level Weight Ablation and Three-\allowbreak{}Option Normalization versus Call Decoupling}\label{sec:5.3.2}

% END M:385

% BEGIN M:387 text

P2 has 2,\allowbreak{}560/\allowbreak{}2,\allowbreak{}560 planned/\allowbreak{}admitted terminal calls and 2,\allowbreak{}048/\allowbreak{}2,\allowbreak{}048 complete evaluation units. First-\allowbreak{}attempt parsing is 2,\allowbreak{}556/\allowbreak{}2,\allowbreak{}560 = 99.84\%; final missing/\allowbreak{}truncated is 0\%. The same four API models are equally weighted. R=2,\allowbreak{} B=10,\allowbreak{}000,\allowbreak{} seed=20261110. P2-\allowbreak{}A has 1,\allowbreak{}280 units,\allowbreak{} with 12 Bayesian and 8 paired-\allowbreak{}framing clusters per model. E0\_\allowbreak{}direct returns one estimate; E1\_\allowbreak{}generic\_\allowbreak{}cot requests generic arithmetic decomposition; E2\_\allowbreak{}unweighted\_\allowbreak{}mcv requests candidates without weights; E3\_\allowbreak{}weighted\_\allowbreak{}mcv requests candidates with weights. Delta\_\allowbreak{}elicit = E3(M\_\allowbreak{}W) \ensuremath{-} E2(M\_\allowbreak{}U) is the prompt-\allowbreak{}level intervention; delta\_\allowbreak{}readout = E3(M\_\allowbreak{}W) \ensuremath{-} E3(M\_\allowbreak{}U) holds generation fixed. Direct-\allowbreak{}CoT here is E1\_\allowbreak{}generic\_\allowbreak{}cot,\allowbreak{} which differs from the Stage 2C posterior-\allowbreak{}focus scaffold.\par

% END M:387

% BEGIN M:389 text

\textbf{Table 5. P2 complete pooled arm means,\allowbreak{} primary effects,\allowbreak{} same-\allowbreak{}generation readouts and P2-\allowbreak{}B arm metrics.} The four panels are identically ordered in both languages and Appendix I. \texttt{95\% marginal CI} is JSON \texttt{ci95\_\allowbreak{}raw}; \texttt{95\% simultaneous CI} is JSON \texttt{ci95\_\allowbreak{}simultaneous},\allowbreak{} a single-\allowbreak{}step Bonferroni percentile bootstrap at alpha/\allowbreak{}(2k) (\texttt{analyze\_\allowbreak{}p2\_\allowbreak{}supplemental.py},\allowbreak{} line 47). P2-\allowbreak{}A k=3 per endpoint and P2-\allowbreak{}B k=6 over the two primary endpoints. Values are pp unless percentages or ratios are indicated; C2-\allowbreak{}C1's simultaneous upper endpoint remains -\allowbreak{}0.006 rather than rounding to zero.\par

% END M:389

% BEGIN M:391 table

\begingroup\scriptsize\setlength{\tabcolsep}{1.3mm}\setlength{\LTpre}{4pt}\setlength{\LTpost}{6pt}
\begin{longtable}{>{\raggedright\arraybackslash}p{19.810mm}>{\raggedright\arraybackslash}p{22.286mm}>{\raggedright\arraybackslash}p{34.667mm}>{\raggedright\arraybackslash}p{42.095mm}>{\raggedright\arraybackslash}p{37.143mm}}
\toprule
Endpoint & E0\_\allowbreak{}direct & E1\_\allowbreak{}generic\_\allowbreak{}cot & E2\_\allowbreak{}unweighted\_\allowbreak{}mcv & E3\_\allowbreak{}weighted\_\allowbreak{}mcv \\
\midrule\endfirsthead
\toprule
Endpoint & E0\_\allowbreak{}direct & E1\_\allowbreak{}generic\_\allowbreak{}cot & E2\_\allowbreak{}unweighted\_\allowbreak{}mcv & E3\_\allowbreak{}weighted\_\allowbreak{}mcv \\
\midrule\endhead
\bottomrule\endfoot
bayes\_\allowbreak{}MAE & 5.05 & 0.76 & 7.75 & 10.51 \\
frame\_\allowbreak{}MAE & 7.89 & 0.00 & 0.08 & 2.13 \\
frame\_\allowbreak{}FrameGap & 9.22 & 0.00 & 0.16 & 4.04 \\
\end{longtable}
\endgroup

% END M:391

% BEGIN M:397 table

\begingroup\scriptsize\setlength{\tabcolsep}{1.3mm}\setlength{\LTpre}{4pt}\setlength{\LTpost}{6pt}
\begin{longtable}{>{\raggedright\arraybackslash}p{33.708mm}>{\raggedright\arraybackslash}p{14.193mm}>{\raggedright\arraybackslash}p{14.193mm}>{\raggedright\arraybackslash}p{17.741mm}>{\raggedright\arraybackslash}p{26.612mm}>{\raggedright\arraybackslash}p{33.708mm}>{\raggedright\arraybackslash}p{10.645mm}}
\toprule
Population /\allowbreak{} family & Endpoint & Contrast & Delta (pp) & 95\% marginal CI & 95\% simultaneous CI & p\_\allowbreak{}Holm \\
\midrule\endfirsthead
\toprule
Population /\allowbreak{} family & Endpoint & Contrast & Delta (pp) & 95\% marginal CI & 95\% simultaneous CI & p\_\allowbreak{}Holm \\
\midrule\endhead
\bottomrule\endfoot
P2-\allowbreak{}A & bayes\_\allowbreak{}MAE & delta\_\allowbreak{}elicit & +2.75 & [+1.75,\allowbreak{} +3.79] & [+1.539,\allowbreak{} +4.031] & 0.0006 \\
P2-\allowbreak{}A & bayes\_\allowbreak{}MAE & delta\_\allowbreak{}vs\_\allowbreak{}cot & +9.74 & [+8.21,\allowbreak{} +11.43] & [+7.943,\allowbreak{} +11.846] & 0.0006 \\
P2-\allowbreak{}A & bayes\_\allowbreak{}MAE & delta\_\allowbreak{}vs\_\allowbreak{}direct & +5.46 & [+3.91,\allowbreak{} +7.11] & [+3.599,\allowbreak{} +7.512] & 0.0006 \\
P2-\allowbreak{}A & frame\_\allowbreak{}MAE & delta\_\allowbreak{}elicit & +2.05 & [+0.48,\allowbreak{} +4.17] & [+0.291,\allowbreak{} +4.785] & 0.0006 \\
P2-\allowbreak{}A & frame\_\allowbreak{}MAE & delta\_\allowbreak{}vs\_\allowbreak{}cot & +2.13 & [+0.56,\allowbreak{} +4.23] & [+0.389,\allowbreak{} +4.863] & 0.0006 \\
P2-\allowbreak{}A & frame\_\allowbreak{}MAE & delta\_\allowbreak{}vs\_\allowbreak{}direct & -\allowbreak{}5.76 & [-\allowbreak{}8.24,\allowbreak{} -\allowbreak{}3.76] & [-\allowbreak{}8.920,\allowbreak{} -\allowbreak{}3.500] & 0.0006 \\
P2-\allowbreak{}A & frame\_\allowbreak{}FrameGap & delta\_\allowbreak{}elicit & +3.88 & [+0.67,\allowbreak{} +8.20] & [+0.318,\allowbreak{} +9.508] & 0.0036 \\
P2-\allowbreak{}A & frame\_\allowbreak{}FrameGap & delta\_\allowbreak{}vs\_\allowbreak{}cot & +4.04 & [+0.85,\allowbreak{} +8.32] & [+0.497,\allowbreak{} +9.617] & 0.0006 \\
P2-\allowbreak{}A & frame\_\allowbreak{}FrameGap & delta\_\allowbreak{}vs\_\allowbreak{}direct & -\allowbreak{}5.18 & [-\allowbreak{}9.55,\allowbreak{} -\allowbreak{}0.25] & [-\allowbreak{}10.426,\allowbreak{} +0.906] & 0.0390 \\
P2-\allowbreak{}B; category 2 & MAE\_\allowbreak{}norm & C2-\allowbreak{}C1 & -\allowbreak{}1.27 & [-\allowbreak{}2.35,\allowbreak{} -\allowbreak{}0.33] & [-\allowbreak{}2.804,\allowbreak{} -\allowbreak{}0.006] & 0.0410 \\
P2-\allowbreak{}B; category 2 & Acc\_\allowbreak{}argmax & C2-\allowbreak{}C1 & +3.91 & [-\allowbreak{}1.95,\allowbreak{} +9.38] & [-\allowbreak{}3.906,\allowbreak{} +10.938] & 0.8455 \\
P2-\allowbreak{}B; category 2 & MAE\_\allowbreak{}norm & C3-\allowbreak{}C1 & -\allowbreak{}1.71 & [-\allowbreak{}2.76,\allowbreak{} -\allowbreak{}0.75] & [-\allowbreak{}3.158,\allowbreak{} -\allowbreak{}0.451] & 0.0012 \\
P2-\allowbreak{}B; category 2 & Acc\_\allowbreak{}argmax & C3-\allowbreak{}C1 & +2.34 & [-\allowbreak{}4.30,\allowbreak{} +8.98] & [-\allowbreak{}6.250,\allowbreak{} +11.719] & 1.0000 \\
P2-\allowbreak{}B; category 3 & MAE\_\allowbreak{}norm & C3-\allowbreak{}C2 & -\allowbreak{}0.44 & [-\allowbreak{}1.37,\allowbreak{} +0.46] & [-\allowbreak{}1.722,\allowbreak{} +0.724] & 1.0000 \\
P2-\allowbreak{}B; category 3 & Acc\_\allowbreak{}argmax & C3-\allowbreak{}C2 & -\allowbreak{}1.56 & [-\allowbreak{}7.03,\allowbreak{} +3.91] & [-\allowbreak{}8.594,\allowbreak{} +5.469] & 1.0000 \\
\end{longtable}
\endgroup

% END M:397

% BEGIN M:415 table

\begingroup\scriptsize\setlength{\tabcolsep}{1.3mm}\setlength{\LTpre}{4pt}\setlength{\LTpost}{6pt}
\begin{longtable}{>{\raggedright\arraybackslash}p{16.147mm}>{\raggedright\arraybackslash}p{12.111mm}>{\raggedright\arraybackslash}p{12.111mm}>{\raggedright\arraybackslash}p{16.147mm}>{\raggedright\arraybackslash}p{36.332mm}>{\raggedright\arraybackslash}p{60.553mm}}
\toprule
Endpoint & E3 M\_\allowbreak{}W & E3 M\_\allowbreak{}U & E3 Top-\allowbreak{}1 & delta\_\allowbreak{}readout (pp) & elicit/\allowbreak{}readout magnitude ratio \\
\midrule\endfirsthead
\toprule
Endpoint & E3 M\_\allowbreak{}W & E3 M\_\allowbreak{}U & E3 Top-\allowbreak{}1 & delta\_\allowbreak{}readout (pp) & elicit/\allowbreak{}readout magnitude ratio \\
\midrule\endhead
\bottomrule\endfoot
bayes\_\allowbreak{}MAE & 10.51 & 10.56 & 10.63 & -\allowbreak{}0.052 & 52.98 \\
frame\_\allowbreak{}MAE & 2.13 & 2.18 & 2.39 & -\allowbreak{}0.049 & 41.63 \\
frame\_\allowbreak{}FrameGap & 4.04 & 4.11 & 4.28 & -\allowbreak{}0.067 & 57.74 \\
\end{longtable}
\endgroup

% END M:415

% BEGIN M:421 table

\begingroup\scriptsize\setlength{\tabcolsep}{1.3mm}\setlength{\LTpre}{4pt}\setlength{\LTpost}{6pt}
\begin{longtable}{>{\raggedright\arraybackslash}p{35.301mm}>{\raggedright\arraybackslash}p{20.539mm}>{\raggedright\arraybackslash}p{17.972mm}>{\raggedright\arraybackslash}p{12.837mm}>{\raggedright\arraybackslash}p{35.943mm}>{\raggedright\arraybackslash}p{30.808mm}}
\toprule
Arm (m\_\allowbreak{}w) & MAE\_\allowbreak{}norm & MAE\_\allowbreak{}raw & S (\%) & Acc\_\allowbreak{}argmax (\%) & Tie rate (\%) \\
\midrule\endfirsthead
\toprule
Arm (m\_\allowbreak{}w) & MAE\_\allowbreak{}norm & MAE\_\allowbreak{}raw & S (\%) & Acc\_\allowbreak{}argmax (\%) & Tie rate (\%) \\
\midrule\endhead
\bottomrule\endfoot
C1\_\allowbreak{}joint\_\allowbreak{}constrained & 8.67 & 9.05 & 102.58 & 77.73 & 1.95 \\
C2\_\allowbreak{}joint\_\allowbreak{}unconstrained & 7.40 & 7.63 & 101.11 & 81.64 & 1.95 \\
C3\_\allowbreak{}separate\_\allowbreak{}unconstrained & 6.96 & 8.44 & 100.30 & 80.08 & 2.73 \\
\end{longtable}
\endgroup

% END M:421

% BEGIN M:427 text

The weight-\allowbreak{}elicitation effects are +2.75 pp [marginal 95\% CI +1.75,\allowbreak{} +3.79],\allowbreak{} +2.05 [+0.48,\allowbreak{} +4.17],\allowbreak{} and +3.88 [+0.67,\allowbreak{} +8.20],\allowbreak{} p\_\allowbreak{}Holm=0.0006 /\allowbreak{} 0.0006 /\allowbreak{} 0.0036. Their same-\allowbreak{}generation counterparts are -\allowbreak{}0.052 /\allowbreak{} -\allowbreak{}0.049 /\allowbreak{} -\allowbreak{}0.067 pp,\allowbreak{} approximately 42\ensuremath{\times}--58\ensuremath{\times} smaller in magnitude. Generic arithmetic CoT errors are 0.00--0.76 pp. Among nine improvement tests,\allowbreak{} only frame\_\allowbreak{}MAE E3 \ensuremath{-} Direct passes the improvement gate:\allowbreak{} -\allowbreak{}5.76 pp,\allowbreak{} marginal CI [-\allowbreak{}8.24,\allowbreak{} -\allowbreak{}3.76],\allowbreak{} p\_\allowbreak{}Holm=0.0006. This is driven largely by Gemma-\allowbreak{}4-\allowbreak{}26B (-\allowbreak{}31.50 pp); leaving one model out yields [-\allowbreak{}10.23,\allowbreak{} +2.82] pp. Bayesian E3 \ensuremath{-} Direct increases error +5.46 pp; FrameGap decreases -\allowbreak{}5.18 pp but fails the full improvement gate. Prompt effects and same-\allowbreak{}generation aggregation should therefore be reported separately.\par

% END M:427

% BEGIN M:429 text

P2-\allowbreak{}B's 1,\allowbreak{}280 planned/\allowbreak{}admitted calls yield 768 evaluation units (16 clusters per model). The three options are mutually exclusive and exhaustive. C1\_\allowbreak{}joint\_\allowbreak{}constrained asks for a joint response summing to 100\%; C2\_\allowbreak{}joint\_\allowbreak{}unconstrained removes that generation constraint and normalizes afterward; C3\_\allowbreak{}separate\_\allowbreak{}unconstrained elicits each option in separate calls and then normalizes. The confirmatory readout is M\_\allowbreak{}W. Normalization rescales by the raw sum S; this does not imply the raw outputs themselves sum to 100\%.\par

% END M:429

% BEGIN M:431 text

C2 \ensuremath{-} C1 improves MAE\_\allowbreak{}norm by -\allowbreak{}1.27 pp (marginal CI [-\allowbreak{}2.35,\allowbreak{} -\allowbreak{}0.33]; simultaneous [-\allowbreak{}2.804,\allowbreak{} -\allowbreak{}0.006]; p\_\allowbreak{}Holm=0.0410),\allowbreak{} with leave-\allowbreak{}one-\allowbreak{}model-\allowbreak{}out range [-\allowbreak{}2.11,\allowbreak{} -\allowbreak{}0.57] pp. Acc\_\allowbreak{}argmax increases +3.91 pp,\allowbreak{} p\_\allowbreak{}Holm=0.8455:\allowbreak{} category 2,\allowbreak{} a local error improvement. C3 \ensuremath{-} C1 improves MAE\_\allowbreak{}norm -\allowbreak{}1.71 pp (marginal [-\allowbreak{}2.76,\allowbreak{} -\allowbreak{}0.75]; simultaneous [-\allowbreak{}3.16,\allowbreak{} -\allowbreak{}0.45]; p\_\allowbreak{}Holm=0.0012),\allowbreak{} but accuracy +2.34 pp has p\_\allowbreak{}Holm=1.0000:\allowbreak{} category 2. C3 \ensuremath{-} C2 is -\allowbreak{}0.44 pp (marginal [-\allowbreak{}1.37,\allowbreak{} +0.46]; simultaneous [-\allowbreak{}1.72,\allowbreak{} +0.72]; p\_\allowbreak{}Holm=1.0000),\allowbreak{} accuracy -\allowbreak{}1.56 pp,\allowbreak{} p\_\allowbreak{}Holm=1.0000:\allowbreak{} category 3,\allowbreak{} no detected additional benefit. This is neither equivalence nor noninferiority. The category rule requires joint evidence for normalized error and argmax accuracy for category 1; no contrast meets it.\par

% END M:431

% BEGIN M:433 text

\section{Discussion and Synthesis}

% END M:433

% BEGIN M:435 text

The Three-\allowbreak{}Class Structural Behavioral Spectrum is a descriptive organization across nine model families and ten reasoning-\allowbreak{}budget configurations (Flash low/\allowbreak{}high count separately). It is not a causal law of architecture,\allowbreak{} training or scale. Stage 1 CLI settings and later API endpoints have distinct provenance and cannot be assumed exchangeable.\par

% END M:435

% BEGIN M:437 text

\textbf{Class I:\allowbreak{} Non-\allowbreak{}Thinking Compact \& Open-\allowbreak{}Weight Models.} Qwen3-\allowbreak{}8B,\allowbreak{} GPT-\allowbreak{}4o-\allowbreak{}mini,\allowbreak{} Gemma-\allowbreak{}4-\allowbreak{}26B-\allowbreak{}A4B-\allowbreak{}IT,\allowbreak{} Nemotron-\allowbreak{}3.5-\allowbreak{}Super-\allowbreak{}VL-\allowbreak{}120B and spark-\allowbreak{}x2.6 show heterogeneous sensitivity to protocol organization. The class includes a large MoE and should not be read as a parameter-\allowbreak{}size threshold. The Nemotron 78.6\% \ensuremath{\rightarrow} 7.1\% collapse comparison is \texttt{multi\_\allowbreak{}repeat} \ensuremath{\rightarrow} \texttt{1p5\_\allowbreak{}asc\_\allowbreak{}repeat}. Its posterior-\allowbreak{}focus errors (10.15 pp framing /\allowbreak{} 16.73 pp Bayesian) and Qwen's \texttt{single} framing error (12.98 pp) are counterexamples to universal Upgraded MCW-\allowbreak{}Asc superiority. Stage 2C pooled framing benefits do not imply P1 isolated PF or ASC effects,\allowbreak{} which largely fail their pooled gates.\par

% END M:437

% BEGIN M:439 text

\textbf{Class II:\allowbreak{} Explicit Reasoning Configurations.} Gemini-\allowbreak{}3.8-\allowbreak{}Flash high versus low and gpt-\allowbreak{}oss-\allowbreak{}120b show that arithmetic performance and contextual susceptibility differ. gpt-\allowbreak{}oss-\allowbreak{}120b's B4 post-\allowbreak{}outcome confirmation change is +10.68 pp; C4 is +0.19 pp and does not license the same practical claim. Flash B1 uses all n=90 items:\allowbreak{} high -\allowbreak{}3.97 versus low -\allowbreak{}1.94 pp. B3 uses the paired n=89 subset:\allowbreak{} high -\allowbreak{}3.99 \ensuremath{\rightarrow} +0.54 pp,\allowbreak{} low -\allowbreak{}1.96 \ensuremath{\rightarrow} -\allowbreak{}0.32 pp. These denominators and estimands should not be substituted. Attention/\allowbreak{}working-\allowbreak{}memory competition or motivated self-\allowbreak{}rationalization could explain some pattern differences,\allowbreak{} but each remains a hypothesis without internal measurement or a discriminating intervention.\par

% END M:439

% BEGIN M:441 text

\textbf{Class III:\allowbreak{} Frontier Commercial Flagships.} Claude Sonnet 5.5 and codex use frozen medium settings; server-\allowbreak{}side internal reasoning is unobservable. Their Stage 1 Dev C MCW errors are 0.88 /\allowbreak{} 0.0001 pp versus MCU 5.89 /\allowbreak{} 1.49 pp. Arithmetic ceilings coexist with Sonnet B3 +1.75 pp,\allowbreak{} B4 -\allowbreak{}4.19 pp and D2 = +2.30 pp,\allowbreak{} alongside codex D2 = +2.46 pp contextual effects. Medium settings must not be interpreted as evidence about absence of internal reasoning.\par

% END M:441

% BEGIN M:443 table

Table 6. Conditional protocol recommendations and evidence boundaries across task structures.\par
\begingroup\scriptsize\setlength{\tabcolsep}{1.3mm}\setlength{\LTpre}{4pt}\setlength{\LTpost}{6pt}
\begin{longtable}{>{\raggedright\arraybackslash}p{45.805mm}>{\raggedright\arraybackslash}p{51.091mm}>{\raggedright\arraybackslash}p{64.304mm}}
\toprule
Task structure & Conditional protocol recommendation & Evidence boundary \\
\midrule\endfirsthead
\toprule
Task structure & Conditional protocol recommendation & Evidence boundary \\
\midrule\endhead
\bottomrule\endfoot
Structured closed arithmetic near ceiling & Compare Direct and short Direct-\allowbreak{}CoT first & Extra candidates/\allowbreak{}weights can increase error; validate the actual endpoint \\
Compact/\allowbreak{}open-\allowbreak{}weight narrative estimates with format failures & Test Upgraded MCW-\allowbreak{}Asc alongside Direct and posterior-\allowbreak{}focus & Organization can repair format; model-\allowbreak{}specific counterexamples remain \\
Positive/\allowbreak{}negative attribute framing & Select using objective MAE and FrameGap separately & Lower asymmetry need not imply lower normative error \\
Hindsight,\allowbreak{} irrelevant anchors,\allowbreak{} status/\allowbreak{}default/\allowbreak{}advice & Retain signed shifts and source-\allowbreak{}specific outcomes & No protocol has uniform debiasing support \\
Three exhaustive mutually exclusive options & C2 single unconstrained generation plus post-\allowbreak{}hoc normalization is a local MAE\_\allowbreak{}norm option & Category 2; accuracy improvement unconfirmed; separate calls add no detected benefit over C2 \\
\end{longtable}
\endgroup

% END M:443

% BEGIN M:451 text

A fixed three-\allowbreak{}candidate K\_\allowbreak{}opt=3 MCV protocol is an independently testable practical hypothesis,\allowbreak{} not an established optimum. The option count in P2-\allowbreak{}B is not a manipulation of candidate count. The idea that multiple estimates split competing perspectives is a hypothesis; the experiment observes numeric outputs and cannot establish that internal account.\par

% END M:451

% BEGIN M:453 text

First-\allowbreak{}slot capture,\allowbreak{} enumeration inertia,\allowbreak{} mental arithmetic error versus scale quantization loss,\allowbreak{} and an RLHF single-\allowbreak{}point preference explanation all remain hypotheses. Current interventions do not separate these explanations,\allowbreak{} and a structural output-\allowbreak{}range exclusion is sufficient to account for some zero string-\allowbreak{}match cells.\par

% END M:453

% BEGIN M:455 text

Limitations include small authored item banks,\allowbreak{} public-\allowbreak{}item exposure,\allowbreak{} provider changes,\allowbreak{} nonrandom execution windows,\allowbreak{} and composite GPT-\allowbreak{}OSS routing with unlogged per-\allowbreak{}unit backend identity. Five candidates from one call are dependent,\allowbreak{} and output length cannot establish a fivefold cost saving. Non-\allowbreak{}rejection or a below-\allowbreak{}SESOI estimate is not equivalence. Our precise probability targets support objective error claims,\allowbreak{} not broad out-\allowbreak{}of-\allowbreak{}sample calibration claims. No new participant or model data were collected for this manuscript package.\par

% END M:455

% BEGIN M:457 text

The Stage 1 frozen 45-\allowbreak{}test six-\allowbreak{}configuration cohort remains distinct from post-\allowbreak{}cohort exploratory Stage 2B/\allowbreak{}2C and independently preregistered formal P1/\allowbreak{}P2. Candidate-\allowbreak{}spread prediction,\allowbreak{} internal-\allowbreak{}mechanism explanations and broad generalization remain future tests. None becomes newly confirmed by completing this cohort.\par

% END M:457

% BEGIN M:459 text

\subsection*{AI Tool Disclosure and Bibliographic Note}

% END M:459

% BEGIN M:461 text

AI tools assisted coding,\allowbreak{} drafting and editorial work; the author retains responsibility. Stage 1 Sections 3--5.1 and Appendices A--H retain their original bytes; references.bib retains all 70 entries from v3 and the named self-\allowbreak{}citation jhang2026hearsay,\allowbreak{} with only internal unverified annotations in zhang2025verbalized,\allowbreak{} roemmele2011choice and zhu2024incoherent and the latter's unverified URL removed. Hash checks establish provenance. The package performs no submission or upload.\par

% END M:461

\clearpage\phantomsection\label{sec:references}\nocite{*}\bibliographystyle{plainnat}\bibliography{references}

% BEGIN A:1 text

\clearpage\section*{English paper appendices A-\allowbreak{}H}

% END A:1

% BEGIN A:3 text

Original preregistration materials and Addendum 1 are public in the study repository. The v2 CSVs and analysis products are local package deliverables,\allowbreak{} without a claim that they are publicly uploaded. Original third-\allowbreak{}party SynAnchors and COPA/\allowbreak{}ProbCOPA texts are not reproduced.\par

% END A:3

% BEGIN A:5 text

\clearpage\section*{Appendix A. Materials,\allowbreak{} scale,\allowbreak{} and seeds}\label{sec:appendices}

% END A:5

% BEGIN A:7 text

\textbf{A:\allowbreak{} anchoring.} SynAnchors 0.1 semantic subset,\allowbreak{} Hugging Face \texttt{TimTargaryen/\allowbreak{}SynAnchors} revision \texttt{e6df428},\allowbreak{} whose data card specifies Apache-\allowbreak{}2.0. We use questions and truths,\allowbreak{} not its anchors or prompts. Eligibility was frozen before development:\allowbreak{} the question must stand alone; truth must be positive and finite,\allowbreak{} with clear units and reference time (time-\allowbreak{}invariant quantities need no year); anchors must be expressible at the item's precision. Of 60 items,\allowbreak{} 36 were eligible and 24 excluded:\allowbreak{} 19 lacked factual truth (subjective preferences,\allowbreak{} bargaining,\allowbreak{} compensation,\allowbreak{} etc.),\allowbreak{} 3 had ambiguous units or quantities,\allowbreak{} and 2 had clearly questionable truths. With fewer than 40 eligible items,\allowbreak{} we followed the plan and wrote 24 additional items with verifiable sources and review. Development has 10 items (6 SynAnchors,\allowbreak{} 4 new); formal testing has 50 (30,\allowbreak{} 20). Analyses also describe source strata. Anchors were rounded to customary item-\allowbreak{}specific precision and frozen individually,\allowbreak{} requiring 0 < low anchor < high anchor.\par

% END A:7

% BEGIN A:9 text

\textbf{B:\allowbreak{} hindsight.} ProbCOPA,\allowbreak{} GitHub \texttt{McGill-\allowbreak{}NLP/\allowbreak{}probabilistic-\allowbreak{}reasoning} revision \texttt{b983b12}. Its repository LICENSE is MIT,\allowbreak{} with no separate data-\allowbreak{}license clarification in README/\allowbreak{}LICENSE. There are 105 COPA effect items,\allowbreak{} each with two outcomes rated from 0--100 for likelihood by 25--30 native speakers. All 105 were matched using source IDs and premise text. Synthetic canary entries inserted by the authors were removed by deleting rows with \texttt{type == canary} and retaining the bytes of all other rows. Each source file was reduced from 5,\allowbreak{}908 to 5,\allowbreak{}858 rows; downloaded cleaning code was not executed. Human ratings are descriptive only:\allowbreak{} mean ratings (\texttt{mean\_\allowbreak{}response}) are primary and medians secondary. Spearman correlations with F aggregate by UID and use average ranks for ties. Acquisition,\allowbreak{} reconstruction,\allowbreak{} and file SHA-\allowbreak{}256 values appear in \texttt{third-\allowbreak{}party-\allowbreak{}data.md}.\par

% END A:9

% BEGIN A:11 text

\textbf{C:\allowbreak{} base rates.} There are 24 newly written scenarios,\allowbreak{} without classic item wording. Bayesian answers are 8.33,\allowbreak{} 50.00,\allowbreak{} and 79.41\% for accuracy set A (90\%/\allowbreak{}10\%),\allowbreak{} and 13.91,\allowbreak{} 64.00,\allowbreak{} and 87.27\% for set B (80\%/\allowbreak{}5\%). Prompts define the event meant by ``positive'' and state that the person is randomly selected from positive cases,\allowbreak{} without other information.\par

% END A:11

% BEGIN A:13 text

\textbf{D:\allowbreak{} conjunction.} Item bank v0.2. Authors stored complete conjunction and disjunction sentences for person items. Components must be single,\allowbreak{} independently truth-\allowbreak{}evaluable facts,\allowbreak{} without protected attributes as the basis for A or B. Person items use ``now'' as a shared assessment time,\allowbreak{} and event pairs are restricted to ``the same day.'' Development includes P 4 and W 3 (1 per dependence stratum); formal testing includes P 20 and W 15 (5 per stratum).\par

% END A:13

% BEGIN A:15 text

\textbf{E:\allowbreak{} status quo and defaults.} The 40 items span 19 domains,\allowbreak{} including 4 each on software settings,\allowbreak{} insurance add-\allowbreak{}ons,\allowbreak{} and subscriptions. Scenarios establish registration,\allowbreak{} activation,\allowbreak{} or initial setup; cost-\allowbreak{}related items specify that the choice precedes payment. T is P in 20 items and Q in 20,\allowbreak{} with P listed first; the formal split is 17/\allowbreak{}15. Stratification by whether T is first or second is therefore a cross-\allowbreak{}item description,\allowbreak{} without identifying a position effect.\par

% END A:15

% BEGIN A:17 text

\textbf{Scale} (per configuration; \texttt{manifest\_\allowbreak{}ids.json})\par

% END A:17

% BEGIN A:19 table

\begingroup\scriptsize\setlength{\tabcolsep}{1.3mm}\setlength{\LTpre}{4pt}\setlength{\LTpost}{6pt}
\begin{longtable}{>{\raggedright\arraybackslash}p{18.482mm}>{\raggedright\arraybackslash}p{53.598mm}>{\raggedright\arraybackslash}p{25.875mm}>{\raggedright\arraybackslash}p{20.330mm}>{\raggedright\arraybackslash}p{11.089mm}>{\raggedright\arraybackslash}p{24.027mm}}
\toprule
Experiment & Clusters (development/\allowbreak{}formal) & Units per item & Development & Formal & Repeat subset \\
\midrule\endfirsthead
\toprule
Experiment & Clusters (development/\allowbreak{}formal) & Units per item & Development & Formal & Repeat subset \\
\midrule\endhead
\bottomrule\endfoot
A & 10/\allowbreak{}50 items & 10 & 100 & 500 & 80 (8 items) \\
B & 15/\allowbreak{}90 items & 16 & 240 & 1,\allowbreak{}440 & 192 (12 items) \\
C & 4/\allowbreak{}20 scenarios (8/\allowbreak{}40 items) & 12 & 96 & 480 & 96 (4 scenarios) \\
D & 7/\allowbreak{}35 items (P 4/\allowbreak{}20,\allowbreak{} W 3/\allowbreak{}15) & 10 & 70 & 350 & 60 (P 4,\allowbreak{} W 2) \\
E & 8/\allowbreak{}32 items & 14 & 112 & 448 & 56 (4 items) \\
Total &  &  & 618 & 3,\allowbreak{}218 & 484 \\
\end{longtable}
\endgroup

% END A:19

% BEGIN A:28 text

\textbf{Seeds}\par

% END A:28

% BEGIN A:30 table

\begingroup\scriptsize\setlength{\tabcolsep}{1.3mm}\setlength{\LTpre}{4pt}\setlength{\LTpost}{6pt}
\begin{longtable}{>{\raggedright\arraybackslash}p{76.358mm}>{\raggedright\arraybackslash}p{48.077mm}>{\raggedright\arraybackslash}p{36.765mm}}
\toprule
Purpose & Seed & Source \\
\midrule\endfirsthead
\toprule
Purpose & Seed & Source \\
\midrule\endhead
\bottomrule\endfoot
A split/\allowbreak{}repeats (SynAnchors) & 20261440/\allowbreak{}20261441 & \texttt{gen\_\allowbreak{}a.py} \\
A split/\allowbreak{}repeats (new items) & 20261442/\allowbreak{}20261443 & \texttt{gen\_\allowbreak{}a.py} \\
B split/\allowbreak{}repeats & 20261450/\allowbreak{}20261451 & \texttt{gen\_\allowbreak{}b.py} \\
C split/\allowbreak{}repeats & 20261410/\allowbreak{}20261412 & \texttt{gen\_\allowbreak{}c.py},\allowbreak{} \texttt{manifest.py} \\
D joint order/\allowbreak{}split/\allowbreak{}repeats & 20261430/\allowbreak{}20261431/\allowbreak{}20261432 & \texttt{gen\_\allowbreak{}d.py} \\
E split/\allowbreak{}repeats & 20261460/\allowbreak{}20261461 & \texttt{manifest.py} \\
Execution/\allowbreak{}call base seeds (development/\allowbreak{}formal) & 20261080/\allowbreak{}20261081 (same as \ensuremath{\delta}-\allowbreak{}NLI) & \texttt{run\_\allowbreak{}psych3.py} \\
Analysis/\allowbreak{}power bootstrap & 2026101903 & \texttt{analysis\_\allowbreak{}psych3.py} \\
\end{longtable}
\endgroup

% END A:30

% BEGIN A:41 text

Execution order initializes from ``base seed \ensuremath{\times} 10 + configuration offset,\allowbreak{}'' shuffling clusters and then units within clusters. Configuration offsets follow \ensuremath{\delta}-\allowbreak{}NLI (qwen3-\allowbreak{}off 0,\allowbreak{} codex 3); agy uses the first 8 hexadecimal digits of the configuration-\allowbreak{}name string's SHA-\allowbreak{}256 (e.g.,\allowbreak{} of the string \texttt{agy:\allowbreak{}claude-\allowbreak{}sonnet-\allowbreak{}5-\allowbreak{}5-\allowbreak{}medium}). A call seed is the first 8 SHA-\allowbreak{}256 digits of ``set|configuration|unit ID|a<attempt number>|base seed,\allowbreak{}'' affecting Qwen3 only.\par

% END A:41

% BEGIN A:43 text

\textbf{Table A1. Admitted units and historical exclusions}\par

% END A:43

% BEGIN A:45 text

Excluded formal lines are reported once per configuration (on its r0 row),\allowbreak{} not allocated to a round. F0/\allowbreak{}F1 rows partition Flash high; they are not additional configurations.\par

% END A:45

% BEGIN A:47 table

\begingroup\scriptsize\setlength{\tabcolsep}{1.3mm}\setlength{\LTpre}{4pt}\setlength{\LTpost}{6pt}
\begin{longtable}{>{\raggedright\arraybackslash}p{33.237mm}>{\raggedright\arraybackslash}p{33.237mm}>{\raggedright\arraybackslash}p{20.453mm}>{\raggedright\arraybackslash}p{12.783mm}>{\raggedright\arraybackslash}p{17.897mm}>{\raggedright\arraybackslash}p{35.793mm}}
\toprule
Configuration & Stage /\allowbreak{} round & Admitted & OK & Missing & Excluded lines \\
\midrule\endfirsthead
\toprule
Configuration & Stage /\allowbreak{} round & Admitted & OK & Missing & Excluded lines \\
\midrule\endhead
\bottomrule\endfoot
codex & dev r0 & 618 & 618 & 0 & 6 \\
codex & formal r0 & 3218 & 3215 & 3 & 22 \\
codex & formal r1 & 484 & 484 & 0 & see r0 \\
Sonnet 5.5 & dev r0 & 618 & 618 & 0 & 3 \\
Sonnet 5.5 & formal r0 & 3218 & 3218 & 0 & 9 \\
Sonnet 5.5 & formal r1 & 484 & 484 & 0 & see r0 \\
Qwen3 & dev r0 & 618 & 617 & 1 & 0 \\
Qwen3 & formal r0 & 3218 & 3211 & 7 & 1 \\
Qwen3 & formal r1 & 484 & 483 & 1 & see r0 \\
gpt-\allowbreak{}oss-\allowbreak{}120b & dev r0 & 618 & 618 & 0 & 4 \\
gpt-\allowbreak{}oss-\allowbreak{}120b & formal r0 & 3218 & 3210 & 8 & 24 \\
gpt-\allowbreak{}oss-\allowbreak{}120b & formal r1 & 484 & 484 & 0 & see r0 \\
Flash low & dev r0 & 618 & 618 & 0 & 1 \\
Flash low & formal r0 & 3218 & 3217 & 1 & 10 \\
Flash low & formal r1 & 484 & 484 & 0 & see r0 \\
Flash high & dev r0 & 618 & 618 & 0 & 1 \\
Flash high & formal r0 & 3218 & 3217 & 1 & 16 \\
Flash high & formal r1 & 484 & 484 & 0 & see r0 \\
Flash high F0 & r0 & 2687 & 2686 & 1 & 13 \\
Flash high F0 & r1 & 362 & 362 & 0 & see r0 \\
Flash high F1 & r0 & 531 & 531 & 0 & 3 \\
Flash high F1 & r1 & 122 & 122 & 0 & see r0 \\
\end{longtable}
\endgroup

% END A:47

% BEGIN A:72 text

\textbf{Table A2. Formal missingness retained under Addendum 2 and the frozen rules}\par

% END A:72

% BEGIN A:74 text

Missing records remain missing; historical excluded lines are a different category. No missing result is silently treated as an observed number.\par

% END A:74

% BEGIN A:76 table

\begingroup\scriptsize\setlength{\tabcolsep}{1.3mm}\setlength{\LTpre}{4pt}\setlength{\LTpost}{6pt}
\begin{longtable}{>{\raggedright\arraybackslash}p{46.354mm}>{\raggedright\arraybackslash}p{35.657mm}>{\raggedright\arraybackslash}p{17.829mm}>{\raggedright\arraybackslash}p{24.960mm}>{\raggedright\arraybackslash}p{31.200mm}}
\toprule
Configuration & Experiment & Round & Missing & Batch \\
\midrule\endfirsthead
\toprule
Configuration & Experiment & Round & Missing & Batch \\
\midrule\endhead
\bottomrule\endfoot
codex & A & r0 & 3 & original batch \\
Qwen3 & E & r0 & 6 & original batch \\
Qwen3 & C & r0 & 1 & original batch \\
Qwen3 & E & r1 & 1 & original batch \\
gpt-\allowbreak{}oss-\allowbreak{}120b & B & r0 & 6 & original batch \\
gpt-\allowbreak{}oss-\allowbreak{}120b & A & r0 & 2 & original batch \\
Flash low & B & r0 & 1 & original batch \\
Flash high & B & r0 & 1 & F0 \\
\end{longtable}
\endgroup

% END A:76

% BEGIN A:87 text

\clearpage\section*{Appendix B. Subject settings,\allowbreak{} parsing,\allowbreak{} and retries}

% END A:87

% BEGIN A:89 text

\textbf{Qwen3.} Qwen3-\allowbreak{}8B,\allowbreak{} HF commit b968826,\allowbreak{} converted to GGUF Q8\_\allowbreak{}0 using llama.cpp build b11560. Parameters are temperature 0.6,\allowbreak{} top-\allowbreak{}k 20,\allowbreak{} top-\allowbreak{}p 0.95,\allowbreak{} min-\allowbreak{}p 0,\allowbreak{} with other samplers disabled; context 16,\allowbreak{}384 and generation limit 8,\allowbreak{}192 tokens. The HF tokenizer's chat template is used with \texttt{enable\_\allowbreak{}thinking=False},\allowbreak{} one user message,\allowbreak{} and no system prompt. Exhausting the length limit before an end token counts as truncation.\par

% END A:89

% BEGIN A:91 text

\textbf{codex.} codex CLI 0.163.0-\allowbreak{}alpha.10.2,\allowbreak{} separately installed under Addendum 1,\allowbreak{} with executable SHA-\allowbreak{}256 in the fingerprint; gpt-\allowbreak{}6.1-\allowbreak{}sol,\allowbreak{} reasoning effort medium,\allowbreak{} ChatGPT login. Isolation options,\allowbreak{} the global \texttt{AGENTS.md} hash,\allowbreak{} event-\allowbreak{}stream checks,\allowbreak{} and internal-\allowbreak{}log checks follow \ensuremath{\delta}-\allowbreak{}NLI. Tool contamination invalidates an attempt and permits one retry separate from parsing retries; persistent contamination is \texttt{tool\_\allowbreak{}contaminated}. Disconnections invalidate attempts and permit retries,\allowbreak{} stopping only after three consecutive instances for the same unit. Calling errors or quota indications stop continuation; the unit is redone later. Changes to global instructions or log-\allowbreak{}integrity failures invalidate the entire session. psych3 executes codex sequentially in one worker.\par

% END A:91

% BEGIN A:93 text

\textbf{agy.} agy CLI 1.4.0 with automatic updates disabled. Each call starts outside a git repository in a fresh temporary directory containing only a custom agent definition:\allowbreak{} \texttt{tools:\allowbreak{} []},\allowbreak{} default components excluded,\allowbreak{} and body ``Answer the user's question directly.'' Execution uses \texttt{-\allowbreak{}-\allowbreak{}sandbox},\allowbreak{} without conversation continuation,\allowbreak{} and a 600-\allowbreak{}second call timeout. Any event-\allowbreak{}stream step other than user input and final answer---including tool events,\allowbreak{} rejected actions,\allowbreak{} or unknown step types---counts as contamination and triggers a retry. Timeout,\allowbreak{} disconnection,\allowbreak{} and truncation invalidate attempts and permit retries; three consecutive instances of the same class for one unit stop execution for review. Rate limits trigger exponential backoff from 5 to 60 minutes; after 6 continuous hours,\allowbreak{} execution sleeps until the next day; 3 cumulative days for the same unit stop execution. Completed units remain valid. Before and after each unit,\allowbreak{} environment snapshots compare agy version,\allowbreak{} permission-\allowbreak{}setting hash,\allowbreak{} setting-\allowbreak{}key names,\allowbreak{} agent-\allowbreak{}file hash,\allowbreak{} and server-\allowbreak{}reported model name; any change stops execution before the next unit. Finalization is per unit:\allowbreak{} persist raw attempt \ensuremath{\rightarrow} append result \ensuremath{\rightarrow} read back and verify \ensuremath{\rightarrow} atomically update registration.\par

% END A:93

% BEGIN A:95 text

\textbf{Parsing.} Probability cells use \ensuremath{\delta}-\allowbreak{}NLI parser v2. Direct responses use the final answer field; an invalid number on that line fails without fallback. Multi-\allowbreak{}estimate responses use the last group beginning with a line resembling ``Estimate 1.'' Exactly 5 lines numbered 1--5 are required,\allowbreak{} with values from 0--100 carrying percent signs and weights in [0,\allowbreak{} 1]; a positive weight sum is normalized. A accepts finite positive quantities,\allowbreak{} including decimals and scientific notation such as \texttt{1.2e6},\allowbreak{} but not commas or ranges; item-\allowbreak{}consistent unit suffixes may be ignored. Nonpositive or nonfinite values,\allowbreak{} an empty final field,\allowbreak{} an incomplete final multi-\allowbreak{}estimate group,\allowbreak{} or total weight \ensuremath{\leq} 0 all fail. Joint D reads the final four lines for direct responses,\allowbreak{} or the final group of 5 lines each containing four percentages and a weight. Any missing number fails the entire cell; partial use is prohibited. Before parsing agy answers,\allowbreak{} Unicode space-\allowbreak{}separator characters (Zs) are replaced with ordinary spaces. For A's comparison answer,\allowbreak{} the first whole-\allowbreak{}word occurrence of higher/\allowbreak{}lower/\allowbreak{}equal in the full response is used,\allowbreak{} case-\allowbreak{}insensitively.\par

% END A:95

% BEGIN A:97 text

\textbf{Strict reader.} Analysis reads only records from normally finalized,\allowbreak{} noninvalidated sessions with matching fingerprints (registered units for agy),\allowbreak{} and matching result/\allowbreak{}raw-\allowbreak{}file hashes (\texttt{analyse\_\allowbreak{}psych3.load\_\allowbreak{}valid(...,\allowbreak{} strict=True)}). Residual records from unfinalized sessions and superseded retries are excluded.\par

% END A:97

% BEGIN A:99 text

\clearpage\section*{Appendix C. Missing-\allowbreak{}value substitutions and preregistered descriptive measures}

% END A:99

% BEGIN A:101 text

\textbf{General missingness-\allowbreak{}protection rule} (PLAN \S{}\S{}1,\allowbreak{} 5b.3,\allowbreak{} 7.1 rules 3--4). Observed values remain unchanged; only missing cells are replaced. Linear statistics have lower-\allowbreak{} and upper-\allowbreak{}bound versions. The lower version assigns 0 to missing cells with positive coefficients and 100 to those with negative coefficients; the upper version reverses this. The positive-\allowbreak{}direction tail probability uses the lower version (proportion of resampled statistics \ensuremath{\leq} 0); the negative direction uses the upper version (\ensuremath{\geq} 0). Conservative two-\allowbreak{}sided p is twice the smaller tail proportion,\allowbreak{} capped at 1. Protection uses the full formal item set for that hypothesis,\allowbreak{} equal item weights,\allowbreak{} and whole-\allowbreak{}cluster resampling; complete-\allowbreak{}item and protected analyses report separate denominators. Clipping bounds C's item-\allowbreak{}level slopes at approximately \ensuremath{\pm}3.01,\allowbreak{} which supplies finite worst-\allowbreak{}case values. When a hypothesis requires both accuracy sets to be complete within a scenario,\allowbreak{} a scenario with one complete set is excluded from the complete-\allowbreak{}item analysis but filled by substitution in the protected analysis. A has no upper bound and no protection (Section 3.4).\par

% END A:101

% BEGIN A:103 text

\textbf{Statistic-\allowbreak{}specific substitutions in D} (any missing number in a joint-\allowbreak{}response cell makes all four missing)\par

% END A:103

% BEGIN A:105 table

\begingroup\scriptsize\setlength{\tabcolsep}{1.3mm}\setlength{\LTpre}{4pt}\setlength{\LTpost}{6pt}
\begin{longtable}{>{\raggedright\arraybackslash}p{31.896mm}>{\raggedright\arraybackslash}p{37.294mm}>{\raggedright\arraybackslash}p{92.009mm}}
\toprule
Statistic & Supported direction & Worst-\allowbreak{}case substitution \\
\midrule\endfirsthead
\toprule
Statistic & Supported direction & Worst-\allowbreak{}case substitution \\
\midrule\endhead
\bottomrule\endfoot
D1:\allowbreak{} \ensuremath{\Delta}\_\allowbreak{}S & Positive & Missing P\_\allowbreak{}S(A and B) = 0; P\_\allowbreak{}S(B) = 100 \\
D2:\allowbreak{} \ensuremath{\Delta}\_\allowbreak{}S \ensuremath{-} \ensuremath{\Delta}\_\allowbreak{}J & Positive & Minimize \ensuremath{\Delta}\_\allowbreak{}S as above; if joint values are missing,\allowbreak{} P\_\allowbreak{}J(A and B) = 100 and P\_\allowbreak{}J(B) = 0 (whole-\allowbreak{}cell missingness gives \ensuremath{\Delta}\_\allowbreak{}J = +100) \\
D3:\allowbreak{} \ensuremath{\Delta}\_\allowbreak{}S,\allowbreak{}multi \ensuremath{-} \ensuremath{\Delta}\_\allowbreak{}S,\allowbreak{}direct & Two-\allowbreak{}sided & Obtain feasible lower and upper difference bounds; use the lower for the positive direction and upper for the negative direction; crossing zero does not confirm direction \\
D4:\allowbreak{} \ensuremath{\Delta}\_\allowbreak{}J,\allowbreak{}multi \ensuremath{-} \ensuremath{\Delta}\_\allowbreak{}J,\allowbreak{}direct & Two-\allowbreak{}sided & As D3,\allowbreak{} using joint elicitation \\
\end{longtable}
\endgroup

% END A:105

% BEGIN A:112 text

\textbf{E.} Item-\allowbreak{}level E1/\allowbreak{}E2,\allowbreak{} y(direct,\allowbreak{}T) \ensuremath{-} y(direct,\allowbreak{}O),\allowbreak{} ranges over [\ensuremath{-}100,\allowbreak{} 100]. E3/\allowbreak{}E4,\allowbreak{} y(multi,\allowbreak{}T) \ensuremath{-} y(multi,\allowbreak{}O) \ensuremath{-} y(direct,\allowbreak{}T) + y(direct,\allowbreak{}O),\allowbreak{} ranges over [\ensuremath{-}200,\allowbreak{} 200]. Both use the general rule. The advice control uses the same cell-\allowbreak{}level substitutions and descriptive intervals,\allowbreak{} outside Holm.\par

% END A:112

% BEGIN A:114 text

\textbf{Combined reporting.} Complete-\allowbreak{}item and protected results are combined into one report only when no cells required by the entire confirmatory family are missing. Because a test without missing cells can still change Holm position through other members' conservative p-\allowbreak{}values,\allowbreak{} the two versions are otherwise reported separately.\par

% END A:114

% BEGIN A:116 text

\textbf{Preregistered descriptive measures} (outside Holm; no significance conclusions)\par

% END A:116

% BEGIN A:118 text

-\allowbreak{} A:\allowbreak{} log errors relative to truth in each condition,\allowbreak{} including unanchored; weighted proportions of multi-\allowbreak{}estimate candidates beyond the anchor (above high anchors,\allowbreak{} below low anchors); agreement between comparison answers and final estimates; strata by item source.
-\allowbreak{} B:\allowbreak{} b stratified by whether the disclosed outcome is higher/\allowbreak{}lower in F; weighted proportions of multi-\allowbreak{}estimate HI candidates below the same-\allowbreak{}format F aggregate; Spearman correlations between F and human ratings; input-\allowbreak{}condition difference c = \ensuremath{\frac{1}{2}} \ensuremath{\Sigma}\ensuremath{_k} [PO(O\_\allowbreak{}k) \ensuremath{-} F(O\_\allowbreak{}k)].
-\allowbreak{} C:\allowbreak{} three-\allowbreak{}point scatterplots per item and nonmonotonicity rates; clipping-\allowbreak{}bound proportions and sensitivity to [0.1\%,\allowbreak{} 99.9\%] clipping; proportions within 1 percentage point of Bayesian answers; weight assigned to the multi-\allowbreak{}estimate candidate nearest the Bayesian answer; joint descriptions of |slope \ensuremath{-} 1|,\allowbreak{} intercept,\allowbreak{} and Bayesian error.
-\allowbreak{} D:\allowbreak{} violation rates for P and W separately by presentation and response format; means and SDs of \ensuremath{\Delta},\allowbreak{} \ensuremath{\Delta}\_\allowbreak{}min,\allowbreak{} and additive-\allowbreak{}identity deviation X for W by dependence stratum; disjunction excess; weighted candidate-\allowbreak{}level violation proportions under joint multi-\allowbreak{}estimate elicitation; whether P(B) < P(A) in person items; repeat-\allowbreak{}subset distributions of differences between two reports for the same unit.
-\allowbreak{} E:\allowbreak{} ``label assigned to T \ensuremath{-} neutral'' and ``neutral \ensuremath{-} label assigned to the other option''; neutral-\allowbreak{}response distributions; E\_\allowbreak{}advice,\allowbreak{} S\_\allowbreak{}E = E\_\allowbreak{}advice \ensuremath{-} E\_\allowbreak{}status\_\allowbreak{}quo (also relative to defaults),\allowbreak{} and their format changes; E\_\allowbreak{}status\_\allowbreak{}quo \ensuremath{-} E\_\allowbreak{}default; weighted candidate proportions on either side of 50; strata with T first/\allowbreak{}second.
-\allowbreak{} Cross-\allowbreak{}experiment (PLAN \S{}10):\allowbreak{} within each configuration,\allowbreak{} paired differences and 95\% intervals on original scales for A's AI\_\allowbreak{}irrelevant,\allowbreak{} AI\_\allowbreak{}informative,\allowbreak{} and S; B's b and e\_\allowbreak{}PO; C's slope and Bayesian error; D's \ensuremath{\Delta} under separate and joint elicitation; and E's E\_\allowbreak{}status\_\allowbreak{}quo,\allowbreak{} E\_\allowbreak{}default,\allowbreak{} and E\_\allowbreak{}advice,\allowbreak{} showing values and differences across formats. \ensuremath{\delta}-\allowbreak{}NLI s is a comparison only; no ratios are used.\par

% END A:118

% BEGIN A:126 text

\clearpage\section*{Appendix D. Development diagnostics and power}

% END A:126

% BEGIN A:128 text

\textbf{Power simulations} (PLAN \S{}\S{}7,\allowbreak{} 7.1; \texttt{power\_\allowbreak{}dev\_\allowbreak{}summary.md}). Development clusters are resampled while retaining condition correlations within items. At the hypothesis-\allowbreak{}statistic level,\allowbreak{} mean contributions are shifted to the specified effect (AI scale in A),\allowbreak{} and planned numbers of clusters are resampled. Hypotheses in one family are resampled together; with fewer than 4 development clusters,\allowbreak{} they are instead resampled separately and this is noted. Missingness is generated from development cell-\allowbreak{}failure rates with a minimum of 2\%; D joint elicitation fails as a whole cell. Scenarios are zero effect,\allowbreak{} the SESOI,\allowbreak{} and twice the SESOI (plus a ceiling scenario in C),\allowbreak{} under 1,\allowbreak{} 1.5,\allowbreak{} and 2 times the development SD. Each scenario has 1,\allowbreak{}000 simulations,\allowbreak{} each running the complete family Holm procedure and decisions with inner bootstrap B = 499. Power targets use complete-\allowbreak{}item-\allowbreak{}subset decisions. Protected-\allowbreak{}version attainment rates are reported alongside them but do not determine screening. For example,\allowbreak{} Sonnet B1's direction-\allowbreak{}confirmation rate at the SESOI is 1.000 for complete items and 0.003 with protection; the frozen rule retains it.\par

% END A:128

% BEGIN A:131 text

Table H1 gives the unchanged lists and all six development acceptance sets. Exact downgrade reasons remain in the audited supplementary JSON,\allowbreak{} separate from formal recommendation-\allowbreak{}control interpretation caveats. The E no-\allowbreak{}expansion exception preserves development fingerprints. Qwen3 is retained by decision as the descriptive benchmark.\par

% END A:131

% BEGIN A:133 text

\clearpage\section*{Appendix E. Execution details,\allowbreak{} addenda,\allowbreak{} and accounting}

% END A:133

% BEGIN A:135 text

\textbf{Fingerprints.} The execution fingerprint binds the plan,\allowbreak{} material files and item manifest (\texttt{manifest\_\allowbreak{}ids.json}),\allowbreak{} prompt and parsing code,\allowbreak{} runner code (including \ensuremath{\delta}-\allowbreak{}NLI \texttt{run.py} and \texttt{parse.py}),\allowbreak{} subject-\allowbreak{}wrapper code,\allowbreak{} subject settings (version,\allowbreak{} model,\allowbreak{} reasoning effort,\allowbreak{} isolation options,\allowbreak{} instruction-\allowbreak{}source hashes,\allowbreak{} etc.),\allowbreak{} and retry rules. For the original agy F0 batches it also binds night-\allowbreak{}scheduler code and settings:\allowbreak{} 4 workers,\allowbreak{} Asia/\allowbreak{}Shanghai,\allowbreak{} new units 00:\allowbreak{}00--07:\allowbreak{}45,\allowbreak{} hard cutoff 08:\allowbreak{}00,\allowbreak{} and automatic-\allowbreak{}release limit 8. The analysis freeze separately binds analysis-\allowbreak{}code hashes,\allowbreak{} B,\allowbreak{} seed,\allowbreak{} \ensuremath{\alpha},\allowbreak{} hypothesis specifications,\allowbreak{} configuration lists,\allowbreak{} and development acceptance. Startup recomputes the execution fingerprint and rejects mismatches against the freeze.\par

% END A:135

% BEGIN A:137 text

\textbf{Time windows (original batches).} Night schedules for codex and Qwen3 start new attempts,\allowbreak{} including retries,\allowbreak{} only between 00:\allowbreak{}00 and 07:\allowbreak{}45. Calls in progress are not forcibly stopped at 08:\allowbreak{}00; units blocked by the window after 07:\allowbreak{}45 are redone the next night. Since Addendum 1,\allowbreak{} codex also permits daytime batches bounded by end time or unit count; no new units start after the boundary. Historical periods for the initial three configurations were codex 10-\allowbreak{}25 00:\allowbreak{}01 to 10-\allowbreak{}27 20:\allowbreak{}53; Sonnet 10-\allowbreak{}23 00:\allowbreak{}01 to 10-\allowbreak{}28 01:\allowbreak{}00; and Qwen3 10-\allowbreak{}23 00:\allowbreak{}01 to 10-\allowbreak{}24 06:\allowbreak{}00.\par

% END A:137

% BEGIN A:139 text

The original agy F0 batches used four workers,\allowbreak{} the 00:\allowbreak{}00-\allowbreak{}07:\allowbreak{}45 admission window and an 08:\allowbreak{}00 hard cutoff. F1 is the approved protocol exception for agy:\allowbreak{}gemini-\allowbreak{}3.8-\allowbreak{}flash-\allowbreak{}high:\allowbreak{} one worker,\allowbreak{} continuous admission from the approved start until formal coverage completed,\allowbreak{} and no 08:\allowbreak{}00 hard cutoff,\allowbreak{} with completion and safety stopping. The subject configuration,\allowbreak{} materials,\allowbreak{} prompts,\allowbreak{} parsing and ten-\allowbreak{}hypothesis list remain unchanged; the distinct F1 batch fingerprint is preserved.\par

% END A:139

% BEGIN A:141 text

\textbf{Addendum 1 (2026-\allowbreak{}10-\allowbreak{}24,\allowbreak{} commit \texttt{940aee5}).}\par

% END A:141

% BEGIN A:143 text

-\allowbreak{} At formal startup on 10-\allowbreak{}23 00:\allowbreak{}00,\allowbreak{} the desktop-\allowbreak{}installed codex CLI had automatically updated. Version checks refused startup,\allowbreak{} and no formal codex units ran. The same version,\allowbreak{} 0.163.0-\allowbreak{}alpha.10.2,\allowbreak{} was separately installed through npm that day in a location not updated by the desktop app. The new entry point verifies executable SHA-\allowbreak{}256 on startup,\allowbreak{} retaining per-\allowbreak{}call version checks. Model,\allowbreak{} reasoning effort,\allowbreak{} login,\allowbreak{} isolation options,\allowbreak{} global \texttt{AGENTS.md},\allowbreak{} prompts,\allowbreak{} parsing,\allowbreak{} inclusion,\allowbreak{} and retry rules are unchanged. Only version-\allowbreak{}string equality is claimed,\allowbreak{} not byte identity with the original executable.
-\allowbreak{} The preregistered codex fingerprint included the model entry's SHA-\allowbreak{}256 from \texttt{codex debug models}. Snapshots on 10-\allowbreak{}24 showed the same model entry alternating between two versions within minutes,\allowbreak{} neither matching the preregistered hash. There were six differences:\allowbreak{} two plugin/\allowbreak{}application sections in base instructions; browser/\allowbreak{}computer confirmation policy; one default-\allowbreak{}collaboration-\allowbreak{}mode sentence; context window (500,\allowbreak{}000 versus 272,\allowbreak{}000 tokens); and \texttt{available\_\allowbreak{}access\_\allowbreak{}programs}. Most concern features disabled through isolation and unused by the task,\allowbreak{} but the context window differs as well. We did not test effects on answers. The fingerprint was changed to bind executable SHA-\allowbreak{}256. The complete entry and hash are saved at session start,\allowbreak{} which identifies only the version observed then,\allowbreak{} not which version served an individual call. Additional snapshots were scheduled every two hours for 14 days from 10-\allowbreak{}24.
-\allowbreak{} Development was fully rerun under the new fingerprint (10-\allowbreak{}24 14:\allowbreak{}35--17:\allowbreak{}14,\allowbreak{} seven sessions) and accepted using original criteria:\allowbreak{} 618/\allowbreak{}618 included,\allowbreak{} 618 first-\allowbreak{}attempt parses,\allowbreak{} 0 missing. The preregistered development set and an interrupted partial rerun remain archived unchanged. Only codex's entry in the analysis freeze changed (executable path/\allowbreak{}hash,\allowbreak{} removal of directory-\allowbreak{}entry hash,\allowbreak{} fingerprint ID); the other five configurations were unchanged.
-\allowbreak{} Execution changes without fingerprint effects:\allowbreak{} sessions stopped solely by inability to connect to OpenAI automatically restart after waits of 5,\allowbreak{} 10,\allowbreak{} and 20 minutes,\allowbreak{} with at most 8 restarts per runner launch and failed units redone. Night jobs no longer open a console window; closing a terminal had interrupted the first formal night on 10-\allowbreak{}23 at 00:\allowbreak{}14.\par

% END A:143

% BEGIN A:148 text

\textbf{Addendum 2 (2026-\allowbreak{}10-\allowbreak{}29; public commit to be added after release),\allowbreak{} \S{}4 rules and decisions.} These do not change prompts,\allowbreak{} parsing,\allowbreak{} fingerprints,\allowbreak{} or scoring of completed answers. agy rules run in a separate script before the frozen night queue; fingerprint-\allowbreak{}bound files are unchanged and every action has an audit record.\par

% END A:148

% BEGIN A:150 text

1. Another recognized connection-\allowbreak{}failure message was added to codex automatic continuation (10-\allowbreak{}25); failed units remain subject to rerun.
2. Initial gpt-\allowbreak{}oss B710 disposition (decided 10-\allowbreak{}25,\allowbreak{} executed 10-\allowbreak{}26):\allowbreak{} a unit rejected by provider content-\allowbreak{}safety filtering three times on each of two nights was administratively marked missing (\texttt{manual\_\allowbreak{}disposition}),\allowbreak{} referencing all six actual attempts without further calls. The same transaction registered two earlier interrupted-\allowbreak{}session residues as excluded; neither had been included.
3. General content-\allowbreak{}safety rule (10-\allowbreak{}26):\allowbreak{} an agy formal unit stopped after three attempts strictly matching provider content-\allowbreak{}safety criteria is retried the next night. A second such stop on another night leads to an administrative missing record.
4. gpt-\allowbreak{}oss B710-\allowbreak{}specific rule (10-\allowbreak{}27):\allowbreak{} all six units disclosing the second outcome were blocked,\allowbreak{} whereas the other ten were not. Only for this item and configuration,\allowbreak{} one strictly verified filtering stop suffices for an administrative missing record. On 10-\allowbreak{}28,\allowbreak{} a B710 unit with one capacity error (HTTP 503) and two content-\allowbreak{}safety refusals across three attempts was also assigned missing under this rule,\allowbreak{} with per-\allowbreak{}attempt reasons recorded. Units completed before filtering remain included.
5. Setting conflicts (night of 10-\allowbreak{}25 to 26):\allowbreak{} agy settings were restored to preregistered baseline before each night,\allowbreak{} but another project repeatedly added permission rules during execution. Environment checks stopped agy configurations; manual release followed the next day and affected units were redone. Other projects subsequently avoided agy during the night window.
6. Insufficient capacity (from 10-\allowbreak{}29):\allowbreak{} stops with three attempts strictly matching HTTP 503,\allowbreak{} at least 30 minutes since the last attempt,\allowbreak{} can be automatically released and rerun,\allowbreak{} at most 2 times per unit and 6 per model per night; beyond these limits,\allowbreak{} review is required. Stopped sessions remain invalidated.
7. Manual review:\allowbreak{} other stops (response-\allowbreak{}check failures,\allowbreak{} disconnections,\allowbreak{} environment checks,\allowbreak{} refused startup) were reviewed and released on 10-\allowbreak{}23,\allowbreak{} 24,\allowbreak{} 25,\allowbreak{} 26,\allowbreak{} and 29. Units were redone under unchanged rules,\allowbreak{} with an audit file for each review.\par

% END A:150

% BEGIN A:159 text

The completed-\allowbreak{}cohort dispositions and stage/\allowbreak{}round counts appear in Tables A1-\allowbreak{}A2. Missing values remain missing and enter the original complete-\allowbreak{}subset and worst-\allowbreak{}case procedures. The post-\allowbreak{}hoc closure of 359 codex units retains its disclosed inclusion status; it is not reclassified as a preregistered closure.\par

% END A:159

% BEGIN A:161 text

\clearpage\section*{Appendix F. Execution-\allowbreak{}window amendment and full endpoint inventory}

% END A:161

% BEGIN A:163 text

The Flash-\allowbreak{}high identities are retained verbatim:\allowbreak{}\par

% END A:163

% BEGIN A:165 text

F0:\allowbreak{} \texttt{\seqsplit{a532f2d198d661032c96b98e9eb0c88e15f391400c4e789dc0c8e9f652fe0a11}}\par

% END A:165

% BEGIN A:167 text

F1:\allowbreak{} \texttt{\seqsplit{8c35dff8fab0d60b39ee4fba09786f0ddca63c43df1c51a515be07438e434885}}\par

% END A:167

% BEGIN A:169 text

F1 binds a new start boundary,\allowbreak{} remaining order,\allowbreak{} inherited development baseline and independent commit receipts. The original agy F0 batches used four workers,\allowbreak{} the 00:\allowbreak{}00-\allowbreak{}07:\allowbreak{}45 admission window and an 08:\allowbreak{}00 hard cutoff. F1 is the approved protocol exception for agy:\allowbreak{}gemini-\allowbreak{}3.8-\allowbreak{}flash-\allowbreak{}high:\allowbreak{} one worker,\allowbreak{} continuous admission from the approved start until formal coverage completed,\allowbreak{} and no 08:\allowbreak{}00 hard cutoff,\allowbreak{} with completion and safety stopping. The subject configuration,\allowbreak{} materials,\allowbreak{} prompts,\allowbreak{} parsing and ten-\allowbreak{}hypothesis list remain unchanged; the distinct F1 batch fingerprint is preserved. Cluster D /\allowbreak{} w01 spans F0 (7 r0 calls) and F1 (3 r0 calls) across both response formats. There are zero cross-\allowbreak{}batch r0/\allowbreak{}r1 repeat pairs:\allowbreak{} every repeated unit has its r0 and r1 calls in the same batch. Its union with F0 uses the original ten-\allowbreak{}hypothesis list; no development set or response prompt is changed. The host analysis passed v1 byte reproduction and seven-\allowbreak{}batch strict admission; this builder does not claim to rerun protected raw admission in the sandbox.\par

% END A:169

% BEGIN A:171 text

\textbf{Table F1. Flash-\allowbreak{}high primary-\allowbreak{}call assignment by material condition}\par

% END A:171

% BEGIN A:173 text

F0 and F1 are non-\allowbreak{}randomized execution windows confounded with item schedule order and collection time. Differences are exploratory and do not identify a causal execution-\allowbreak{}window or model effect.\par

% END A:173

% BEGIN A:175 table

\begingroup\scriptsize\setlength{\tabcolsep}{1.3mm}\setlength{\LTpre}{4pt}\setlength{\LTpost}{6pt}
\begin{longtable}{>{\raggedright\arraybackslash}p{16.872mm}>{\raggedright\arraybackslash}p{33.745mm}>{\raggedright\arraybackslash}p{91.111mm}>{\raggedright\arraybackslash}p{16.872mm}}
\toprule
Batch & Experiment & Condition (frozen cell key) & Units \\
\midrule\endfirsthead
\toprule
Batch & Experiment & Condition (frozen cell key) & Units \\
\midrule\endhead
\bottomrule\endfoot
F0 & A & ('informative',\allowbreak{} 'high') & 86 \\
F0 & A & ('informative',\allowbreak{} 'low') & 86 \\
F0 & A & ('irrelevant',\allowbreak{} 'high') & 86 \\
F0 & A & ('irrelevant',\allowbreak{} 'low') & 86 \\
F0 & A & ('none',\allowbreak{} 'none') & 86 \\
F0 & B & ('F',\allowbreak{} 1) & 152 \\
F0 & B & ('F',\allowbreak{} 2) & 152 \\
F0 & B & ('HI',\allowbreak{} 1) & 152 \\
F0 & B & ('HI',\allowbreak{} 2) & 152 \\
F0 & B & ('HP',\allowbreak{} 1) & 152 \\
F0 & B & ('HP',\allowbreak{} 2) & 152 \\
F0 & B & ('PO',\allowbreak{} 1) & 152 \\
F0 & B & ('PO',\allowbreak{} 2) & 152 \\
F0 & C & ('frequency',\allowbreak{} 1) & 68 \\
F0 & C & ('frequency',\allowbreak{} 10) & 68 \\
F0 & C & ('frequency',\allowbreak{} 30) & 68 \\
F0 & C & ('probability',\allowbreak{} 1) & 68 \\
F0 & C & ('probability',\allowbreak{} 10) & 68 \\
F0 & C & ('probability',\allowbreak{} 30) & 68 \\
F0 & D & ('J',\allowbreak{} 'a') & 59 \\
F0 & D & ('S',\allowbreak{} 'a') & 60 \\
F0 & D & ('S',\allowbreak{} 'b') & 59 \\
F0 & D & ('S',\allowbreak{} 'conj') & 60 \\
F0 & D & ('S',\allowbreak{} 'disj') & 59 \\
F0 & E & ('default\_\allowbreak{}O',\allowbreak{}) & 48 \\
F0 & E & ('default\_\allowbreak{}T',\allowbreak{}) & 48 \\
F0 & E & ('neutral',\allowbreak{}) & 48 \\
F0 & E & ('recommend\_\allowbreak{}O',\allowbreak{}) & 48 \\
F0 & E & ('recommend\_\allowbreak{}T',\allowbreak{}) & 48 \\
F0 & E & ('status\_\allowbreak{}O',\allowbreak{}) & 48 \\
F0 & E & ('status\_\allowbreak{}T',\allowbreak{}) & 48 \\
F1 & A & ('informative',\allowbreak{} 'high') & 14 \\
F1 & A & ('informative',\allowbreak{} 'low') & 14 \\
F1 & A & ('irrelevant',\allowbreak{} 'high') & 14 \\
F1 & A & ('irrelevant',\allowbreak{} 'low') & 14 \\
F1 & A & ('none',\allowbreak{} 'none') & 14 \\
F1 & B & ('F',\allowbreak{} 1) & 28 \\
F1 & B & ('F',\allowbreak{} 2) & 28 \\
F1 & B & ('HI',\allowbreak{} 1) & 28 \\
F1 & B & ('HI',\allowbreak{} 2) & 28 \\
F1 & B & ('HP',\allowbreak{} 1) & 28 \\
F1 & B & ('HP',\allowbreak{} 2) & 28 \\
F1 & B & ('PO',\allowbreak{} 1) & 28 \\
F1 & B & ('PO',\allowbreak{} 2) & 28 \\
F1 & C & ('frequency',\allowbreak{} 1) & 12 \\
F1 & C & ('frequency',\allowbreak{} 10) & 12 \\
F1 & C & ('frequency',\allowbreak{} 30) & 12 \\
F1 & C & ('probability',\allowbreak{} 1) & 12 \\
F1 & C & ('probability',\allowbreak{} 10) & 12 \\
F1 & C & ('probability',\allowbreak{} 30) & 12 \\
F1 & D & ('J',\allowbreak{} 'a') & 11 \\
F1 & D & ('S',\allowbreak{} 'a') & 10 \\
F1 & D & ('S',\allowbreak{} 'b') & 11 \\
F1 & D & ('S',\allowbreak{} 'conj') & 10 \\
F1 & D & ('S',\allowbreak{} 'disj') & 11 \\
F1 & E & ('default\_\allowbreak{}O',\allowbreak{}) & 16 \\
F1 & E & ('default\_\allowbreak{}T',\allowbreak{}) & 16 \\
F1 & E & ('neutral',\allowbreak{}) & 16 \\
F1 & E & ('recommend\_\allowbreak{}O',\allowbreak{}) & 16 \\
F1 & E & ('recommend\_\allowbreak{}T',\allowbreak{}) & 16 \\
F1 & E & ('status\_\allowbreak{}O',\allowbreak{}) & 16 \\
F1 & E & ('status\_\allowbreak{}T',\allowbreak{}) & 16 \\
\end{longtable}
\endgroup

% END A:175

% BEGIN A:240 text

\textbf{Table F2. All 120 planned endpoints and development eligibility}\par

% END A:240

% BEGIN A:242 text

Each endpoint is reported once per configuration. Confirmatory decisions are in H2 only. Development eligibility is separate from formal recommendation-\allowbreak{}control interpretation caveats. Full exact eligibility reasons and feasible missingness bounds are retained in the audited supplementary JSON.\par

% END A:242

% BEGIN A:244 table

\begingroup\scriptsize\setlength{\tabcolsep}{1.3mm}\setlength{\LTpre}{4pt}\setlength{\LTpost}{6pt}
\begin{longtable}{>{\raggedright\arraybackslash}p{20.375mm}>{\raggedright\arraybackslash}p{7.837mm}>{\raggedright\arraybackslash}p{42.317mm}>{\raggedright\arraybackslash}p{43.885mm}>{\raggedright\arraybackslash}p{20.375mm}>{\raggedright\arraybackslash}p{18.612mm}}
\toprule
Configuration & H & Complete /\allowbreak{} planned clusters & Estimate [unadjusted 95\% CI] & Missing cells & Eligibility \\
\midrule\endfirsthead
\toprule
Configuration & H & Complete /\allowbreak{} planned clusters & Estimate [unadjusted 95\% CI] & Missing cells & Eligibility \\
\midrule\endhead
\bottomrule\endfoot
codex & A1 & 50/\allowbreak{}50 & -\allowbreak{}0.030 [-\allowbreak{}0.156,\allowbreak{} 0.087] & 0 & dev\_\allowbreak{}diagnostic \\
codex & A2 & 50/\allowbreak{}50 & 0.459 [0.285,\allowbreak{} 0.657] & 0 & dev\_\allowbreak{}diagnostic \\
codex & A3 & 49/\allowbreak{}50 & 0.005 [-\allowbreak{}0.151,\allowbreak{} 0.158] & 2 & dev\_\allowbreak{}diagnostic \\
codex & A4 & 49/\allowbreak{}50 & 0.051 [-\allowbreak{}0.108,\allowbreak{} 0.206] & 2 & dev\_\allowbreak{}diagnostic \\
codex & B1 & 90/\allowbreak{}90 & -\allowbreak{}2.289 [-\allowbreak{}3.589,\allowbreak{} -\allowbreak{}1.033] & 0 & frozen\_\allowbreak{}confirmatory \\
codex & B2 & 90/\allowbreak{}90 & 48.256 [44.144,\allowbreak{} 52.028] & 0 & dev\_\allowbreak{}power \\
codex & B3 & 90/\allowbreak{}90 & -\allowbreak{}0.363 [-\allowbreak{}1.570,\allowbreak{} 0.797] & 0 & frozen\_\allowbreak{}confirmatory \\
codex & B4 & 90/\allowbreak{}90 & 0.905 [0.337,\allowbreak{} 2.013] & 0 & frozen\_\allowbreak{}confirmatory \\
codex & C1 & 20/\allowbreak{}20 & -\allowbreak{}0.000 [-\allowbreak{}0.000,\allowbreak{} -\allowbreak{}0.000] & 0 & dev\_\allowbreak{}diagnostic \\
codex & C2 & 20/\allowbreak{}20 & 0.000 [0.000,\allowbreak{} 0.000] & 0 & dev\_\allowbreak{}diagnostic \\
codex & C3 & 20/\allowbreak{}20 & -\allowbreak{}0.000 [-\allowbreak{}0.000,\allowbreak{} -\allowbreak{}0.000] & 0 & dev\_\allowbreak{}diagnostic \\
codex & C4 & 20/\allowbreak{}20 & -\allowbreak{}0.002 [-\allowbreak{}0.002,\allowbreak{} -\allowbreak{}0.001] & 0 & dev\_\allowbreak{}diagnostic \\
codex & D1 & 20/\allowbreak{}20 & -\allowbreak{}1.425 [-\allowbreak{}3.850,\allowbreak{} 0.375] & 0 & frozen\_\allowbreak{}confirmatory \\
codex & D2 & 20/\allowbreak{}20 & 2.458 [1.175,\allowbreak{} 3.875] & 0 & frozen\_\allowbreak{}confirmatory \\
codex & D3 & 20/\allowbreak{}20 & -\allowbreak{}0.741 [-\allowbreak{}1.708,\allowbreak{} 0.231] & 0 & frozen\_\allowbreak{}confirmatory \\
codex & D4 & 20/\allowbreak{}20 & 0.143 [-\allowbreak{}0.134,\allowbreak{} 0.446] & 0 & frozen\_\allowbreak{}confirmatory \\
codex & E1 & 32/\allowbreak{}32 & -\allowbreak{}0.312 [-\allowbreak{}0.938,\allowbreak{} 0.000] & 0 & frozen\_\allowbreak{}confirmatory \\
codex & E2 & 32/\allowbreak{}32 & 0.000 [0.000,\allowbreak{} 0.000] & 0 & frozen\_\allowbreak{}confirmatory \\
codex & E3 & 32/\allowbreak{}32 & -\allowbreak{}0.383 [-\allowbreak{}2.086,\allowbreak{} 1.234] & 0 & frozen\_\allowbreak{}confirmatory \\
codex & E4 & 32/\allowbreak{}32 & -\allowbreak{}0.023 [-\allowbreak{}1.555,\allowbreak{} 1.523] & 0 & frozen\_\allowbreak{}confirmatory \\
Qwen3 & A1 & 50/\allowbreak{}50 & 0.946 [-\allowbreak{}0.231,\allowbreak{} 2.231] & 0 & descriptive\_\allowbreak{}only \\
Qwen3 & A2 & 50/\allowbreak{}50 & -\allowbreak{}1.316 [-\allowbreak{}2.827,\allowbreak{} 0.060] & 0 & descriptive\_\allowbreak{}only \\
Qwen3 & A3 & 50/\allowbreak{}50 & -\allowbreak{}0.608 [-\allowbreak{}1.955,\allowbreak{} 0.723] & 0 & descriptive\_\allowbreak{}only \\
Qwen3 & A4 & 50/\allowbreak{}50 & 1.576 [-\allowbreak{}0.114,\allowbreak{} 3.309] & 0 & descriptive\_\allowbreak{}only \\
Qwen3 & B1 & 90/\allowbreak{}90 & -\allowbreak{}7.751 [-\allowbreak{}12.228,\allowbreak{} -\allowbreak{}3.201] & 0 & descriptive\_\allowbreak{}only \\
Qwen3 & B2 & 90/\allowbreak{}90 & 54.128 [49.861,\allowbreak{} 58.583] & 0 & descriptive\_\allowbreak{}only \\
Qwen3 & B3 & 90/\allowbreak{}90 & 5.721 [0.677,\allowbreak{} 10.757] & 0 & descriptive\_\allowbreak{}only \\
Qwen3 & B4 & 90/\allowbreak{}90 & 19.053 [15.600,\allowbreak{} 22.624] & 0 & descriptive\_\allowbreak{}only \\
Qwen3 & C1 & 20/\allowbreak{}20 & -\allowbreak{}0.002 [-\allowbreak{}0.003,\allowbreak{} -\allowbreak{}0.000] & 0 & descriptive\_\allowbreak{}only \\
Qwen3 & C2 & 19/\allowbreak{}20 & -\allowbreak{}0.001 [-\allowbreak{}0.003,\allowbreak{} -\allowbreak{}0.000] & 1 & descriptive\_\allowbreak{}only \\
Qwen3 & C3 & 20/\allowbreak{}20 & -\allowbreak{}0.012 [-\allowbreak{}0.017,\allowbreak{} -\allowbreak{}0.008] & 0 & descriptive\_\allowbreak{}only \\
Qwen3 & C4 & 20/\allowbreak{}20 & 0.647 [0.537,\allowbreak{} 0.761] & 0 & descriptive\_\allowbreak{}only \\
Qwen3 & D1 & 20/\allowbreak{}20 & 13.025 [8.000,\allowbreak{} 19.075] & 0 & descriptive\_\allowbreak{}only \\
Qwen3 & D2 & 20/\allowbreak{}20 & 15.775 [9.400,\allowbreak{} 22.900] & 0 & descriptive\_\allowbreak{}only \\
Qwen3 & D3 & 20/\allowbreak{}20 & -\allowbreak{}4.613 [-\allowbreak{}12.936,\allowbreak{} 3.610] & 0 & descriptive\_\allowbreak{}only \\
Qwen3 & D4 & 20/\allowbreak{}20 & -\allowbreak{}2.583 [-\allowbreak{}5.984,\allowbreak{} 0.726] & 0 & descriptive\_\allowbreak{}only \\
Qwen3 & E1 & 32/\allowbreak{}32 & -\allowbreak{}1.648 [-\allowbreak{}9.719,\allowbreak{} 6.562] & 0 & descriptive\_\allowbreak{}only \\
Qwen3 & E2 & 32/\allowbreak{}32 & -\allowbreak{}0.532 [-\allowbreak{}8.913,\allowbreak{} 6.589] & 0 & descriptive\_\allowbreak{}only \\
Qwen3 & E3 & 31/\allowbreak{}32 & 4.042 [-\allowbreak{}5.568,\allowbreak{} 14.325] & 2 & descriptive\_\allowbreak{}only \\
Qwen3 & E4 & 29/\allowbreak{}32 & 5.659 [-\allowbreak{}4.104,\allowbreak{} 15.685] & 3 & descriptive\_\allowbreak{}only \\
Sonnet 5.5 & A1 & 50/\allowbreak{}50 & 0.019 [-\allowbreak{}0.069,\allowbreak{} 0.098] & 0 & frozen\_\allowbreak{}confirmatory \\
Sonnet 5.5 & A2 & 50/\allowbreak{}50 & 0.471 [0.340,\allowbreak{} 0.606] & 0 & dev\_\allowbreak{}power \\
Sonnet 5.5 & A3 & 50/\allowbreak{}50 & -\allowbreak{}0.057 [-\allowbreak{}0.187,\allowbreak{} 0.049] & 0 & frozen\_\allowbreak{}confirmatory \\
Sonnet 5.5 & A4 & 50/\allowbreak{}50 & 0.131 [-\allowbreak{}0.044,\allowbreak{} 0.365] & 0 & dev\_\allowbreak{}power \\
Sonnet 5.5 & B1 & 90/\allowbreak{}90 & -\allowbreak{}3.169 [-\allowbreak{}4.772,\allowbreak{} -\allowbreak{}1.633] & 0 & frozen\_\allowbreak{}confirmatory \\
Sonnet 5.5 & B2 & 90/\allowbreak{}90 & 3.931 [1.678,\allowbreak{} 6.344] & 0 & frozen\_\allowbreak{}confirmatory \\
Sonnet 5.5 & B3 & 90/\allowbreak{}90 & 1.750 [0.613,\allowbreak{} 2.836] & 0 & frozen\_\allowbreak{}confirmatory \\
Sonnet 5.5 & B4 & 90/\allowbreak{}90 & -\allowbreak{}4.193 [-\allowbreak{}6.646,\allowbreak{} -\allowbreak{}1.792] & 0 & frozen\_\allowbreak{}confirmatory \\
Sonnet 5.5 & C1 & 20/\allowbreak{}20 & -\allowbreak{}0.000 [-\allowbreak{}0.000,\allowbreak{} -\allowbreak{}0.000] & 0 & dev\_\allowbreak{}diagnostic \\
Sonnet 5.5 & C2 & 20/\allowbreak{}20 & 0.000 [0.000,\allowbreak{} 0.000] & 0 & dev\_\allowbreak{}diagnostic \\
Sonnet 5.5 & C3 & 20/\allowbreak{}20 & -\allowbreak{}0.061 [-\allowbreak{}0.069,\allowbreak{} -\allowbreak{}0.053] & 0 & dev\_\allowbreak{}diagnostic \\
Sonnet 5.5 & C4 & 20/\allowbreak{}20 & 1.006 [0.888,\allowbreak{} 1.121] & 0 & dev\_\allowbreak{}diagnostic \\
Sonnet 5.5 & D1 & 20/\allowbreak{}20 & -\allowbreak{}1.480 [-\allowbreak{}3.730,\allowbreak{} 0.080] & 0 & frozen\_\allowbreak{}confirmatory \\
Sonnet 5.5 & D2 & 20/\allowbreak{}20 & 2.300 [1.305,\allowbreak{} 3.490] & 0 & frozen\_\allowbreak{}confirmatory \\
Sonnet 5.5 & D3 & 20/\allowbreak{}20 & -\allowbreak{}0.417 [-\allowbreak{}1.560,\allowbreak{} 0.508] & 0 & frozen\_\allowbreak{}confirmatory \\
Sonnet 5.5 & D4 & 20/\allowbreak{}20 & -\allowbreak{}0.095 [-\allowbreak{}0.439,\allowbreak{} 0.211] & 0 & frozen\_\allowbreak{}confirmatory \\
Sonnet 5.5 & E1 & 32/\allowbreak{}32 & 0.344 [-\allowbreak{}1.250,\allowbreak{} 1.844] & 0 & dev\_\allowbreak{}power \\
Sonnet 5.5 & E2 & 32/\allowbreak{}32 & -\allowbreak{}1.406 [-\allowbreak{}3.125,\allowbreak{} 0.219] & 0 & frozen\_\allowbreak{}confirmatory \\
Sonnet 5.5 & E3 & 32/\allowbreak{}32 & -\allowbreak{}0.298 [-\allowbreak{}3.066,\allowbreak{} 2.423] & 0 & dev\_\allowbreak{}power \\
Sonnet 5.5 & E4 & 32/\allowbreak{}32 & 0.389 [-\allowbreak{}2.530,\allowbreak{} 3.095] & 0 & frozen\_\allowbreak{}confirmatory \\
gpt-\allowbreak{}oss-\allowbreak{}120b & A1 & 50/\allowbreak{}50 & 0.039 [-\allowbreak{}0.944,\allowbreak{} 0.978] & 0 & frozen\_\allowbreak{}confirmatory \\
gpt-\allowbreak{}oss-\allowbreak{}120b & A2 & 50/\allowbreak{}50 & -\allowbreak{}0.040 [-\allowbreak{}1.288,\allowbreak{} 1.099] & 0 & dev\_\allowbreak{}power \\
gpt-\allowbreak{}oss-\allowbreak{}120b & A3 & 50/\allowbreak{}50 & 0.158 [-\allowbreak{}0.777,\allowbreak{} 1.133] & 0 & dev\_\allowbreak{}power \\
gpt-\allowbreak{}oss-\allowbreak{}120b & A4 & 49/\allowbreak{}50 & 0.147 [-\allowbreak{}1.068,\allowbreak{} 1.473] & 1 & dev\_\allowbreak{}power \\
gpt-\allowbreak{}oss-\allowbreak{}120b & B1 & 89/\allowbreak{}90 & 0.028 [-\allowbreak{}2.764,\allowbreak{} 2.837] & 1 & dev\_\allowbreak{}power \\
gpt-\allowbreak{}oss-\allowbreak{}120b & B2 & 89/\allowbreak{}90 & 49.612 [46.197,\allowbreak{} 52.988] & 2 & dev\_\allowbreak{}power \\
gpt-\allowbreak{}oss-\allowbreak{}120b & B3 & 89/\allowbreak{}90 & 6.143 [3.198,\allowbreak{} 9.236] & 2 & dev\_\allowbreak{}power \\
gpt-\allowbreak{}oss-\allowbreak{}120b & B4 & 89/\allowbreak{}90 & 10.679 [7.607,\allowbreak{} 13.805] & 2 & frozen\_\allowbreak{}confirmatory \\
gpt-\allowbreak{}oss-\allowbreak{}120b & C1 & 20/\allowbreak{}20 & -\allowbreak{}0.000 [-\allowbreak{}0.000,\allowbreak{} -\allowbreak{}0.000] & 0 & dev\_\allowbreak{}diagnostic \\
gpt-\allowbreak{}oss-\allowbreak{}120b & C2 & 20/\allowbreak{}20 & -\allowbreak{}0.000 [-\allowbreak{}0.000,\allowbreak{} -\allowbreak{}0.000] & 0 & dev\_\allowbreak{}diagnostic \\
gpt-\allowbreak{}oss-\allowbreak{}120b & C3 & 20/\allowbreak{}20 & -\allowbreak{}0.003 [-\allowbreak{}0.005,\allowbreak{} -\allowbreak{}0.000] & 0 & dev\_\allowbreak{}diagnostic \\
gpt-\allowbreak{}oss-\allowbreak{}120b & C4 & 20/\allowbreak{}20 & 0.191 [0.155,\allowbreak{} 0.231] & 0 & dev\_\allowbreak{}diagnostic \\
gpt-\allowbreak{}oss-\allowbreak{}120b & D1 & 20/\allowbreak{}20 & 9.600 [3.200,\allowbreak{} 17.100] & 0 & dev\_\allowbreak{}power \\
gpt-\allowbreak{}oss-\allowbreak{}120b & D2 & 20/\allowbreak{}20 & 17.350 [9.750,\allowbreak{} 25.550] & 0 & dev\_\allowbreak{}power \\
gpt-\allowbreak{}oss-\allowbreak{}120b & D3 & 20/\allowbreak{}20 & -\allowbreak{}2.695 [-\allowbreak{}10.673,\allowbreak{} 4.297] & 0 & dev\_\allowbreak{}power \\
gpt-\allowbreak{}oss-\allowbreak{}120b & D4 & 20/\allowbreak{}20 & -\allowbreak{}1.963 [-\allowbreak{}4.289,\allowbreak{} 0.402] & 0 & dev\_\allowbreak{}power \\
gpt-\allowbreak{}oss-\allowbreak{}120b & E1 & 32/\allowbreak{}32 & -\allowbreak{}1.133 [-\allowbreak{}7.561,\allowbreak{} 5.484] & 0 & dev\_\allowbreak{}power \\
gpt-\allowbreak{}oss-\allowbreak{}120b & E2 & 32/\allowbreak{}32 & 1.478 [-\allowbreak{}3.219,\allowbreak{} 6.572] & 0 & frozen\_\allowbreak{}confirmatory \\
gpt-\allowbreak{}oss-\allowbreak{}120b & E3 & 32/\allowbreak{}32 & 5.121 [-\allowbreak{}2.083,\allowbreak{} 12.396] & 0 & dev\_\allowbreak{}power \\
gpt-\allowbreak{}oss-\allowbreak{}120b & E4 & 32/\allowbreak{}32 & 3.144 [-\allowbreak{}2.042,\allowbreak{} 8.020] & 0 & dev\_\allowbreak{}power \\
Flash low & A1 & 50/\allowbreak{}50 & 0.042 [-\allowbreak{}0.047,\allowbreak{} 0.183] & 0 & dev\_\allowbreak{}diagnostic \\
Flash low & A2 & 50/\allowbreak{}50 & 0.179 [0.042,\allowbreak{} 0.310] & 0 & dev\_\allowbreak{}diagnostic \\
Flash low & A3 & 50/\allowbreak{}50 & 0.038 [-\allowbreak{}0.036,\allowbreak{} 0.145] & 0 & dev\_\allowbreak{}diagnostic \\
Flash low & A4 & 50/\allowbreak{}50 & -\allowbreak{}0.018 [-\allowbreak{}0.208,\allowbreak{} 0.178] & 0 & dev\_\allowbreak{}diagnostic \\
Flash low & B1 & 90/\allowbreak{}90 & -\allowbreak{}1.939 [-\allowbreak{}3.973,\allowbreak{} 0.022] & 0 & frozen\_\allowbreak{}confirmatory \\
Flash low & B2 & 90/\allowbreak{}90 & 10.967 [7.222,\allowbreak{} 14.844] & 0 & dev\_\allowbreak{}power \\
Flash low & B3 & 89/\allowbreak{}90 & 1.643 [0.410,\allowbreak{} 2.943] & 1 & frozen\_\allowbreak{}confirmatory \\
Flash low & B4 & 90/\allowbreak{}90 & 2.520 [-\allowbreak{}0.513,\allowbreak{} 5.570] & 0 & frozen\_\allowbreak{}confirmatory \\
Flash low & C1 & 20/\allowbreak{}20 & -\allowbreak{}0.000 [-\allowbreak{}0.000,\allowbreak{} -\allowbreak{}0.000] & 0 & dev\_\allowbreak{}diagnostic \\
Flash low & C2 & 20/\allowbreak{}20 & 0.000 [0.000,\allowbreak{} 0.000] & 0 & dev\_\allowbreak{}diagnostic \\
Flash low & C3 & 20/\allowbreak{}20 & -\allowbreak{}0.000 [-\allowbreak{}0.000,\allowbreak{} -\allowbreak{}0.000] & 0 & dev\_\allowbreak{}diagnostic \\
Flash low & C4 & 20/\allowbreak{}20 & 0.005 [0.003,\allowbreak{} 0.007] & 0 & dev\_\allowbreak{}diagnostic \\
Flash low & D1 & 20/\allowbreak{}20 & 0.085 [-\allowbreak{}1.015,\allowbreak{} 1.495] & 0 & frozen\_\allowbreak{}confirmatory \\
Flash low & D2 & 20/\allowbreak{}20 & 3.373 [1.137,\allowbreak{} 6.130] & 0 & frozen\_\allowbreak{}confirmatory \\
Flash low & D3 & 20/\allowbreak{}20 & 0.487 [-\allowbreak{}1.654,\allowbreak{} 2.775] & 0 & frozen\_\allowbreak{}confirmatory \\
Flash low & D4 & 20/\allowbreak{}20 & -\allowbreak{}0.167 [-\allowbreak{}0.772,\allowbreak{} 0.355] & 0 & frozen\_\allowbreak{}confirmatory \\
Flash low & E1 & 32/\allowbreak{}32 & -\allowbreak{}0.531 [-\allowbreak{}1.625,\allowbreak{} 0.188] & 0 & frozen\_\allowbreak{}confirmatory \\
Flash low & E2 & 32/\allowbreak{}32 & 0.000 [0.000,\allowbreak{} 0.000] & 0 & frozen\_\allowbreak{}confirmatory \\
Flash low & E3 & 32/\allowbreak{}32 & -\allowbreak{}0.731 [-\allowbreak{}2.833,\allowbreak{} 1.519] & 0 & dev\_\allowbreak{}power \\
Flash low & E4 & 32/\allowbreak{}32 & 1.645 [-\allowbreak{}0.183,\allowbreak{} 3.558] & 0 & fingerprint\_\allowbreak{}preservation\_\allowbreak{}exception \\
Flash high & A1 & 50/\allowbreak{}50 & 0.034 [0.006,\allowbreak{} 0.072] & 0 & dev\_\allowbreak{}diagnostic \\
Flash high & A2 & 50/\allowbreak{}50 & -\allowbreak{}0.026 [-\allowbreak{}0.241,\allowbreak{} 0.123] & 0 & dev\_\allowbreak{}diagnostic \\
Flash high & A3 & 50/\allowbreak{}50 & -\allowbreak{}0.048 [-\allowbreak{}0.103,\allowbreak{} -\allowbreak{}0.003] & 0 & dev\_\allowbreak{}diagnostic \\
Flash high & A4 & 50/\allowbreak{}50 & 0.197 [0.034,\allowbreak{} 0.458] & 0 & dev\_\allowbreak{}diagnostic \\
Flash high & B1 & 90/\allowbreak{}90 & -\allowbreak{}3.972 [-\allowbreak{}5.983,\allowbreak{} -\allowbreak{}2.122] & 0 & frozen\_\allowbreak{}confirmatory \\
Flash high & B2 & 90/\allowbreak{}90 & 0.850 [0.044,\allowbreak{} 1.761] & 0 & frozen\_\allowbreak{}confirmatory \\
Flash high & B3 & 89/\allowbreak{}90 & 4.526 [3.036,\allowbreak{} 6.125] & 1 & frozen\_\allowbreak{}confirmatory \\
Flash high & B4 & 90/\allowbreak{}90 & 1.403 [-\allowbreak{}1.909,\allowbreak{} 4.820] & 0 & dev\_\allowbreak{}power \\
Flash high & C1 & 20/\allowbreak{}20 & -\allowbreak{}0.000 [-\allowbreak{}0.000,\allowbreak{} -\allowbreak{}0.000] & 0 & dev\_\allowbreak{}diagnostic \\
Flash high & C2 & 20/\allowbreak{}20 & 0.000 [0.000,\allowbreak{} 0.000] & 0 & dev\_\allowbreak{}diagnostic \\
Flash high & C3 & 20/\allowbreak{}20 & 0.000 [-\allowbreak{}0.000,\allowbreak{} 0.000] & 0 & dev\_\allowbreak{}diagnostic \\
Flash high & C4 & 20/\allowbreak{}20 & 0.005 [0.003,\allowbreak{} 0.010] & 0 & dev\_\allowbreak{}diagnostic \\
Flash high & D1 & 20/\allowbreak{}20 & -\allowbreak{}1.115 [-\allowbreak{}2.115,\allowbreak{} -\allowbreak{}0.215] & 0 & frozen\_\allowbreak{}confirmatory \\
Flash high & D2 & 20/\allowbreak{}20 & 3.335 [1.245,\allowbreak{} 5.865] & 0 & frozen\_\allowbreak{}confirmatory \\
Flash high & D3 & 20/\allowbreak{}20 & 0.979 [-\allowbreak{}0.727,\allowbreak{} 2.689] & 0 & frozen\_\allowbreak{}confirmatory \\
Flash high & D4 & 20/\allowbreak{}20 & 0.758 [-\allowbreak{}0.169,\allowbreak{} 2.001] & 0 & frozen\_\allowbreak{}confirmatory \\
Flash high & E1 & 32/\allowbreak{}32 & 0.625 [0.000,\allowbreak{} 1.875] & 0 & frozen\_\allowbreak{}confirmatory \\
Flash high & E2 & 32/\allowbreak{}32 & 0.000 [0.000,\allowbreak{} 0.000] & 0 & frozen\_\allowbreak{}confirmatory \\
Flash high & E3 & 32/\allowbreak{}32 & -\allowbreak{}2.188 [-\allowbreak{}4.438,\allowbreak{} 0.075] & 0 & fingerprint\_\allowbreak{}preservation\_\allowbreak{}exception \\
Flash high & E4 & 32/\allowbreak{}32 & 0.142 [-\allowbreak{}1.577,\allowbreak{} 1.897] & 0 & frozen\_\allowbreak{}confirmatory \\
\end{longtable}
\endgroup

% END A:244

% BEGIN A:367 text

\clearpage\section*{Appendix G. Exploratory candidate-\allowbreak{}spread analyses}

% END A:367

% BEGIN A:369 text

\textbf{Table G1. Exploratory weighted candidate-\allowbreak{}spread distributions}\par

% END A:369

% BEGIN A:371 text

A spread is measured on the log scale; B-\allowbreak{}E in probability points. D joint spread averages conjunction and B component spreads,\allowbreak{} not all four statements. D valid counts refer to calls. Full distributions by condition:\allowbreak{} supplement-\allowbreak{}v2/\allowbreak{}g2-\allowbreak{}spread-\allowbreak{}by-\allowbreak{}condition.csv (198 rows). Item pairing counts:\allowbreak{} supplement-\allowbreak{}v2/\allowbreak{}g3-\allowbreak{}paired-\allowbreak{}cell-\allowbreak{}counts.csv (1,\allowbreak{}482 rows). These v2 CSVs are delivered locally,\allowbreak{} not claimed to be live on GitHub.\par

% END A:371

% BEGIN A:373 table

\begingroup\scriptsize\setlength{\tabcolsep}{1.3mm}\setlength{\LTpre}{4pt}\setlength{\LTpost}{6pt}
\begin{longtable}{>{\raggedright\arraybackslash}p{21.616mm}>{\raggedright\arraybackslash}p{8.314mm}>{\raggedright\arraybackslash}p{31.593mm}>{\raggedright\arraybackslash}p{8.314mm}>{\raggedright\arraybackslash}p{11.640mm}>{\raggedright\arraybackslash}p{13.302mm}>{\raggedright\arraybackslash}p{8.314mm}>{\raggedright\arraybackslash}p{9.977mm}>{\raggedright\arraybackslash}p{8.314mm}>{\raggedright\arraybackslash}p{21.616mm}}
\toprule
Configuration & Exp & Planned multi calls & Valid & Skipped & Unpaired & Q1 & Median & Q3 & Zero fraction \\
\midrule\endfirsthead
\toprule
Configuration & Exp & Planned multi calls & Valid & Skipped & Unpaired & Q1 & Median & Q3 & Zero fraction \\
\midrule\endhead
\bottomrule\endfoot
codex & A & 250 & 247 & 3 & 0 & 0.0096 & 0.0522 & 0.1455 & 0.000 \\
codex & B & 720 & 720 & 0 & 0 & 0.9581 & 6.9417 & 10.9545 & 0.075 \\
codex & C & 240 & 240 & 0 & 0 & 0.0000 & 0.0000 & 0.0000 & 1.000 \\
codex & D & 175 & 175 & 0 & 0 & 1.5945 & 6.0696 & 10.9545 & 0.000 \\
codex & E & 224 & 224 & 0 & 0 & 10.9545 & 11.1523 & 15.0831 & 0.000 \\
Sonnet 5.5 & A & 250 & 250 & 0 & 0 & 0.0503 & 0.1418 & 0.2400 & 0.000 \\
Sonnet 5.5 & B & 720 & 720 & 0 & 0 & 3.5163 & 6.2639 & 10.0369 & 0.000 \\
Sonnet 5.5 & C & 240 & 240 & 0 & 0 & 1.6325 & 6.5819 & 11.2777 & 0.000 \\
Sonnet 5.5 & D & 175 & 175 & 0 & 0 & 1.3434 & 3.4825 & 7.4795 & 0.000 \\
Sonnet 5.5 & E & 224 & 224 & 0 & 0 & 8.5398 & 9.7559 & 11.5814 & 0.000 \\
Qwen3 & A & 250 & 250 & 0 & 0 & 0.0429 & 0.0922 & 0.2110 & 0.000 \\
Qwen3 & B & 720 & 720 & 0 & 0 & 6.6144 & 11.2823 & 20.5183 & 0.074 \\
Qwen3 & C & 240 & 240 & 0 & 1 & 1.6926 & 2.7436 & 4.4615 & 0.004 \\
Qwen3 & D & 175 & 175 & 0 & 0 & 4.9686 & 7.2647 & 12.7720 & 0.000 \\
Qwen3 & E & 224 & 218 & 6 & 0 & 11.5293 & 13.9272 & 17.6014 & 0.000 \\
gpt-\allowbreak{}oss-\allowbreak{}120b & A & 250 & 248 & 2 & 0 & 0.0213 & 0.0680 & 0.1551 & 0.000 \\
gpt-\allowbreak{}oss-\allowbreak{}120b & B & 720 & 717 & 3 & 0 & 6.5990 & 11.9478 & 20.2037 & 0.006 \\
gpt-\allowbreak{}oss-\allowbreak{}120b & C & 240 & 240 & 0 & 0 & 0.8700 & 1.8166 & 2.4150 & 0.000 \\
gpt-\allowbreak{}oss-\allowbreak{}120b & D & 175 & 175 & 0 & 0 & 5.2180 & 8.7006 & 13.7148 & 0.000 \\
gpt-\allowbreak{}oss-\allowbreak{}120b & E & 224 & 224 & 0 & 0 & 12.8774 & 16.8300 & 19.3200 & 0.000 \\
Flash low & A & 250 & 250 & 0 & 0 & 0.0067 & 0.0465 & 0.1140 & 0.000 \\
Flash low & B & 720 & 719 & 1 & 0 & 2.7383 & 6.4177 & 10.6946 & 0.000 \\
Flash low & C & 240 & 240 & 0 & 0 & 0.0616 & 0.1101 & 0.1581 & 0.000 \\
Flash low & D & 175 & 175 & 0 & 0 & 1.8731 & 4.0924 & 8.5474 & 0.000 \\
Flash low & E & 224 & 224 & 0 & 0 & 8.0612 & 10.2956 & 11.6619 & 0.000 \\
Flash high & A & 250 & 250 & 0 & 0 & 0.0061 & 0.0367 & 0.1173 & 0.000 \\
Flash high & B & 720 & 719 & 1 & 0 & 4.6503 & 7.2902 & 11.3908 & 0.000 \\
Flash high & C & 240 & 240 & 0 & 0 & 0.0454 & 0.1096 & 0.1615 & 0.000 \\
Flash high & D & 175 & 175 & 0 & 0 & 1.8654 & 4.4272 & 8.6493 & 0.000 \\
Flash high & E & 224 & 224 & 0 & 0 & 9.1864 & 11.1692 & 13.0118 & 0.000 \\
\end{longtable}
\endgroup

% END A:373

% BEGIN A:407 text

\clearpage\section*{Appendix H. Complete six-\allowbreak{}configuration overview tables}

% END A:407

% BEGIN A:409 text

\textbf{Table H1. Development acceptance and frozen configuration-\allowbreak{}specific gates}\par

% END A:409

% BEGIN A:411 text

All six configurations triggered the C ceiling (48/\allowbreak{}48 development direct cells within one point). A known truth uses 20 unanchored estimates. No D floor gate triggered. Low E4 (39 required items) and high E3 (36) were downgraded rather than expanding 32 formal E items and invalidating development fingerprints. Qwen3 was retained as a descriptive benchmark by decision,\allowbreak{} with no confirmatory tests.\par

% END A:411

% BEGIN A:413 table

\begingroup\scriptsize\setlength{\tabcolsep}{1.3mm}\setlength{\LTpre}{4pt}\setlength{\LTpost}{6pt}
\begin{longtable}{>{\raggedright\arraybackslash}p{15.842mm}>{\raggedright\arraybackslash}p{21.935mm}>{\raggedright\arraybackslash}p{8.530mm}>{\raggedright\arraybackslash}p{15.842mm}>{\raggedright\arraybackslash}p{10.967mm}>{\raggedright\arraybackslash}p{24.372mm}>{\raggedright\arraybackslash}p{53.313mm}}
\toprule
Configuration & First parses /\allowbreak{} 618 & Missing & A known truth & Forwarded & Hypotheses & Downgrade categories \\
\midrule\endfirsthead
\toprule
Configuration & First parses /\allowbreak{} 618 & Missing & A known truth & Forwarded & Hypotheses & Downgrade categories \\
\midrule\endhead
\bottomrule\endfoot
codex & 618 & 0 & 80\% & 11 & B1 B3 B4 D1 D2 D3 D4 E1 E2 E3 E4 & dev\_\allowbreak{}diagnostic:\allowbreak{} 8; dev\_\allowbreak{}power:\allowbreak{} 1 \\
Sonnet 5.5 & 617 & 0 & 70\% & 12 & A1 A3 B1 B2 B3 B4 D1 D2 D3 D4 E2 E4 & dev\_\allowbreak{}power:\allowbreak{} 4; dev\_\allowbreak{}diagnostic:\allowbreak{} 4 \\
Qwen3 & 610 & 1 & 15\% & 0 & descriptive only & descriptive\_\allowbreak{}only:\allowbreak{} 20 \\
gpt-\allowbreak{}oss-\allowbreak{}120b & 612 & 0 & 40\% & 3 & A1 B4 E2 & dev\_\allowbreak{}power:\allowbreak{} 13; dev\_\allowbreak{}diagnostic:\allowbreak{} 4 \\
Flash low & 617 & 0 & 85\% & 9 & B1 B3 B4 D1 D2 D3 D4 E1 E2 & dev\_\allowbreak{}diagnostic:\allowbreak{} 8; dev\_\allowbreak{}power:\allowbreak{} 2; fingerprint\_\allowbreak{}preservation\_\allowbreak{}exception:\allowbreak{} 1 \\
Flash high & 616 & 0 & 80\% & 10 & B1 B2 B3 D1 D2 D3 D4 E1 E2 E4 & dev\_\allowbreak{}diagnostic:\allowbreak{} 8; dev\_\allowbreak{}power:\allowbreak{} 1; fingerprint\_\allowbreak{}preservation\_\allowbreak{}exception:\allowbreak{} 1 \\
\end{longtable}
\endgroup

% END A:413

% BEGIN A:422 text

\textbf{Table H2. All 45 frozen confirmatory tests and missingness protection}\par

% END A:422

% BEGIN A:424 text

One-\allowbreak{}sided unbounded endpoints are printed as unbounded. These are per-\allowbreak{}family Holm-\allowbreak{}step intervals,\allowbreak{} not fixed 95\% simultaneous intervals. Worst-\allowbreak{}case rows are shown whenever a family required protection. Complete-\allowbreak{}case rejected flags are preserved separately from the final directional decision. A uses AI units (SESOI 0.10),\allowbreak{} others 5 pp.\par

% END A:424

% BEGIN A:426 table

\begingroup\scriptsize\setlength{\tabcolsep}{1.3mm}\setlength{\LTpre}{4pt}\setlength{\LTpost}{6pt}
\begin{longtable}{>{\raggedright\arraybackslash}p{19.559mm}>{\raggedright\arraybackslash}p{7.523mm}>{\raggedright\arraybackslash}p{12.037mm}>{\raggedright\arraybackslash}p{45.137mm}>{\raggedright\arraybackslash}p{12.037mm}>{\raggedright\arraybackslash}p{21.064mm}>{\raggedright\arraybackslash}p{12.037mm}>{\raggedright\arraybackslash}p{18.807mm}}
\toprule
Configuration & H & Analysis & Estimate [adjusted bound /\allowbreak{} CI] & Clusters & p /\allowbreak{} Holm alpha & Rejected & Final status \\
\midrule\endfirsthead
\toprule
Configuration & H & Analysis & Estimate [adjusted bound /\allowbreak{} CI] & Clusters & p /\allowbreak{} Holm alpha & Rejected & Final status \\
\midrule\endhead
\bottomrule\endfoot
codex & B1 & complete & -\allowbreak{}2.289 [-\allowbreak{}3.389,\allowbreak{} unbounded] & 90 & 0.9998 /\allowbreak{} 0.0500 & False & not retained \\
codex & B3 & complete & -\allowbreak{}0.363 [-\allowbreak{}1.749,\allowbreak{} 0.931] & 90 & 0.5512 /\allowbreak{} 0.0250 & False & not retained \\
codex & B4 & complete & 0.905 [0.328,\allowbreak{} 2.538] & 90 & <.001 /\allowbreak{} 0.0167 & True & retained directional \\
codex & D1 & complete & -\allowbreak{}1.425 [-\allowbreak{}3.400,\allowbreak{} unbounded] & 20 & 0.9248 /\allowbreak{} 0.0500 & False & not retained \\
codex & D2 & complete & 2.458 [1.015,\allowbreak{} unbounded] & 20 & <.001 /\allowbreak{} 0.0125 & True & retained directional \\
codex & D3 & complete & -\allowbreak{}0.741 [-\allowbreak{}1.894,\allowbreak{} 0.457] & 20 & 0.1344 /\allowbreak{} 0.0167 & False & not retained \\
codex & D4 & complete & 0.143 [-\allowbreak{}0.172,\allowbreak{} 0.497] & 20 & 0.3352 /\allowbreak{} 0.0250 & False & not retained \\
codex & E1 & complete & -\allowbreak{}0.312 [-\allowbreak{}1.250,\allowbreak{} 0.000] & 32 & 0.7008 /\allowbreak{} 0.0167 & False & not retained \\
codex & E2 & complete & 0.000 [0.000,\allowbreak{} 0.000] & 32 & 1.0000 /\allowbreak{} 0.0500 & False & not retained \\
codex & E3 & complete & -\allowbreak{}0.383 [-\allowbreak{}2.578,\allowbreak{} 1.617] & 32 & 0.6748 /\allowbreak{} 0.0125 & False & not retained \\
codex & E4 & complete & -\allowbreak{}0.023 [-\allowbreak{}1.758,\allowbreak{} 1.773] & 32 & 0.9564 /\allowbreak{} 0.0250 & False & not retained \\
Sonnet 5.5 & A1 & complete & 0.019 [-\allowbreak{}0.069,\allowbreak{} unbounded] & 50 & 0.3192 /\allowbreak{} 0.0250 & False & not retained \\
Sonnet 5.5 & A3 & complete & -\allowbreak{}0.057 [-\allowbreak{}0.187,\allowbreak{} 0.049] & 50 & 0.3260 /\allowbreak{} 0.0500 & False & not retained \\
Sonnet 5.5 & B1 & complete & -\allowbreak{}3.169 [-\allowbreak{}4.497,\allowbreak{} unbounded] & 90 & 1.0000 /\allowbreak{} 0.0500 & False & not retained \\
Sonnet 5.5 & B2 & complete & 3.931 [1.497,\allowbreak{} unbounded] & 90 & <.001 /\allowbreak{} 0.0167 & True & retained directional \\
Sonnet 5.5 & B3 & complete & 1.750 [0.479,\allowbreak{} 2.996] & 90 & <.001 /\allowbreak{} 0.0250 & True & retained directional \\
Sonnet 5.5 & B4 & complete & -\allowbreak{}4.193 [-\allowbreak{}7.350,\allowbreak{} -\allowbreak{}1.081] & 90 & <.001 /\allowbreak{} 0.0125 & True & retained directional \\
Sonnet 5.5 & D1 & complete & -\allowbreak{}1.480 [-\allowbreak{}3.365,\allowbreak{} unbounded] & 20 & 0.9628 /\allowbreak{} 0.0500 & False & not retained \\
Sonnet 5.5 & D2 & complete & 2.300 [1.235,\allowbreak{} unbounded] & 20 & <.001 /\allowbreak{} 0.0125 & True & retained directional \\
Sonnet 5.5 & D3 & complete & -\allowbreak{}0.417 [-\allowbreak{}1.810,\allowbreak{} 0.660] & 20 & 0.4768 /\allowbreak{} 0.0167 & False & not retained \\
Sonnet 5.5 & D4 & complete & -\allowbreak{}0.095 [-\allowbreak{}0.500,\allowbreak{} 0.242] & 20 & 0.5724 /\allowbreak{} 0.0250 & False & not retained \\
Sonnet 5.5 & E2 & complete & -\allowbreak{}1.406 [-\allowbreak{}3.375,\allowbreak{} 0.469] & 32 & 0.0988 /\allowbreak{} 0.0250 & False & not retained \\
Sonnet 5.5 & E4 & complete & 0.389 [-\allowbreak{}2.530,\allowbreak{} 3.095] & 32 & 0.7768 /\allowbreak{} 0.0500 & False & not retained \\
gpt-\allowbreak{}oss-\allowbreak{}120b & A1 & complete & 0.039 [-\allowbreak{}0.704,\allowbreak{} unbounded] & 50 & 0.4374 /\allowbreak{} 0.0500 & False & not retained \\
gpt-\allowbreak{}oss-\allowbreak{}120b & B4 & complete & 10.679 [7.607,\allowbreak{} 13.805] & 89 & <.001 /\allowbreak{} 0.0500 & True & bound exceeds SESOI \\
gpt-\allowbreak{}oss-\allowbreak{}120b & B4 & worst-\allowbreak{}case & 10.035 [6.785,\allowbreak{} 14.320] & 90 & <.001 /\allowbreak{} 0.0500 & True & protected=True \\
gpt-\allowbreak{}oss-\allowbreak{}120b & E2 & complete & 1.478 [-\allowbreak{}3.219,\allowbreak{} 6.572] & 32 & 0.5748 /\allowbreak{} 0.0500 & False & not retained \\
Flash low & B1 & complete & -\allowbreak{}1.939 [-\allowbreak{}3.668,\allowbreak{} unbounded] & 90 & 0.9738 /\allowbreak{} 0.0500 & False & not retained \\
Flash low & B1 & worst-\allowbreak{}case & -\allowbreak{}1.939 [-\allowbreak{}3.668,\allowbreak{} unbounded] & 90 & 0.9738 /\allowbreak{} 0.0500 & False & protected=False \\
Flash low & B3 & complete & 1.643 [0.094,\allowbreak{} 3.243] & 89 & 0.0108 /\allowbreak{} 0.0167 & True & not retained \\
Flash low & B3 & worst-\allowbreak{}case & 1.141 [-\allowbreak{}0.792,\allowbreak{} 3.182] & 90 & 0.1724 /\allowbreak{} 0.0250 & False & protected=False \\
Flash low & B4 & complete & 2.520 [-\allowbreak{}0.810,\allowbreak{} 6.058] & 90 & 0.0996 /\allowbreak{} 0.0250 & False & not retained \\
Flash low & B4 & worst-\allowbreak{}case & 2.520 [-\allowbreak{}1.080,\allowbreak{} 6.341] & 90 & 0.0996 /\allowbreak{} 0.0167 & False & protected=False \\
Flash low & D1 & complete & 0.085 [-\allowbreak{}1.100,\allowbreak{} unbounded] & 20 & 0.4782 /\allowbreak{} 0.0167 & False & not retained \\
Flash low & D2 & complete & 3.373 [0.847,\allowbreak{} unbounded] & 20 & <.001 /\allowbreak{} 0.0125 & True & retained directional \\
Flash low & D3 & complete & 0.487 [-\allowbreak{}1.654,\allowbreak{} 2.775] & 20 & 0.6724 /\allowbreak{} 0.0500 & False & not retained \\
Flash low & D4 & complete & -\allowbreak{}0.167 [-\allowbreak{}0.868,\allowbreak{} 0.420] & 20 & 0.5892 /\allowbreak{} 0.0250 & False & not retained \\
Flash low & E1 & complete & -\allowbreak{}0.531 [-\allowbreak{}1.875,\allowbreak{} 0.188] & 32 & 0.3132 /\allowbreak{} 0.0250 & False & not retained \\
Flash low & E2 & complete & 0.000 [0.000,\allowbreak{} 0.000] & 32 & 1.0000 /\allowbreak{} 0.0500 & False & not retained \\
Flash high & B1 & complete & -\allowbreak{}3.972 [-\allowbreak{}5.639,\allowbreak{} unbounded] & 90 & 1.0000 /\allowbreak{} 0.0500 & False & not retained \\
Flash high & B1 & worst-\allowbreak{}case & -\allowbreak{}3.972 [-\allowbreak{}5.639,\allowbreak{} unbounded] & 90 & 1.0000 /\allowbreak{} 0.0500 & False & protected=False \\
Flash high & B2 & complete & 0.850 [0.044,\allowbreak{} unbounded] & 90 & 0.0196 /\allowbreak{} 0.0250 & True & retained directional \\
Flash high & B2 & worst-\allowbreak{}case & 0.850 [0.044,\allowbreak{} unbounded] & 90 & 0.0196 /\allowbreak{} 0.0250 & True & protected=True \\
Flash high & B3 & complete & 4.526 [2.762,\allowbreak{} 6.453] & 89 & <.001 /\allowbreak{} 0.0167 & True & retained directional \\
Flash high & B3 & worst-\allowbreak{}case & 4.114 [1.982,\allowbreak{} 6.584] & 90 & <.001 /\allowbreak{} 0.0167 & True & protected=True \\
Flash high & D1 & complete & -\allowbreak{}1.115 [-\allowbreak{}1.950,\allowbreak{} unbounded] & 20 & 0.9942 /\allowbreak{} 0.0500 & False & not retained \\
Flash high & D2 & complete & 3.335 [1.075,\allowbreak{} unbounded] & 20 & <.001 /\allowbreak{} 0.0125 & True & retained directional \\
Flash high & D3 & complete & 0.979 [-\allowbreak{}0.996,\allowbreak{} 2.891] & 20 & 0.2628 /\allowbreak{} 0.0250 & False & not retained \\
Flash high & D4 & complete & 0.758 [-\allowbreak{}0.294,\allowbreak{} 2.302] & 20 & 0.1544 /\allowbreak{} 0.0167 & False & not retained \\
Flash high & E1 & complete & 0.625 [0.000,\allowbreak{} 2.500] & 32 & 0.7184 /\allowbreak{} 0.0167 & False & not retained \\
Flash high & E2 & complete & 0.000 [0.000,\allowbreak{} 0.000] & 32 & 1.0000 /\allowbreak{} 0.0500 & False & not retained \\
Flash high & E4 & complete & 0.142 [-\allowbreak{}1.844,\allowbreak{} 2.092] & 32 & 0.8348 /\allowbreak{} 0.0250 & False & not retained \\
\end{longtable}
\endgroup

% END A:426

% BEGIN A:481 text

\textbf{Table H3. A,\allowbreak{} B,\allowbreak{} C descriptive summaries across six configurations}\par

% END A:481

% BEGIN A:483 text

Unadjusted descriptive cluster-\allowbreak{}bootstrap intervals; complete paired subsets can differ from single-\allowbreak{}format endpoint denominators. Supplemental B2 is HP minus HI; C1 is 1 minus direct probability slope; C2 is direct frequency minus probability slope; neither is itself a direct probability slope. Supplemental endpoints remain descriptive when outside the frozen list.\par

% END A:483

% BEGIN A:485 table

\begingroup\scriptsize\setlength{\tabcolsep}{1.3mm}\setlength{\LTpre}{4pt}\setlength{\LTpost}{6pt}
\begin{longtable}{>{\raggedright\arraybackslash}p{16.758mm}>{\raggedright\arraybackslash}p{19.336mm}>{\raggedright\arraybackslash}p{33.516mm}>{\raggedright\arraybackslash}p{29.649mm}>{\raggedright\arraybackslash}p{34.805mm}>{\raggedright\arraybackslash}p{19.336mm}}
\toprule
Configuration & Quantity & Direct /\allowbreak{} endpoint [95\% CI] & Multi-\allowbreak{}estimate [95\% CI] & Multi minus direct [95\% CI] & Paired clusters \\
\midrule\endfirsthead
\toprule
Configuration & Quantity & Direct /\allowbreak{} endpoint [95\% CI] & Multi-\allowbreak{}estimate [95\% CI] & Multi minus direct [95\% CI] & Paired clusters \\
\midrule\endhead
\bottomrule\endfoot
codex & A AI\_\allowbreak{}irrelevant & -\allowbreak{}0.0302 [-\allowbreak{}0.1487,\allowbreak{} 0.0887] & -\allowbreak{}0.0252 [-\allowbreak{}0.1100,\allowbreak{} 0.0511] & 0.0051 [-\allowbreak{}0.1535,\allowbreak{} 0.1553] & 49 \\
codex & A AI\_\allowbreak{}informative & 0.4291 [0.3012,\allowbreak{} 0.5552] & 0.4845 [0.3591,\allowbreak{} 0.6066] & 0.0554 [0.0173,\allowbreak{} 0.1016] & 50 \\
codex & A S & 0.4680 [0.2886,\allowbreak{} 0.6508] & 0.5193 [0.3683,\allowbreak{} 0.6758] & 0.0512 [-\allowbreak{}0.1034,\allowbreak{} 0.2102] & 49 \\
codex & B b & -\allowbreak{}2.2889 [-\allowbreak{}3.5889,\allowbreak{} -\allowbreak{}1.0333] & -\allowbreak{}2.6514 [-\allowbreak{}3.9655,\allowbreak{} -\allowbreak{}1.4595] & -\allowbreak{}0.3626 [-\allowbreak{}1.5705,\allowbreak{} 0.7967] & 90 \\
codex & B e\_\allowbreak{}PO & 0.0000 [0.0000,\allowbreak{} 0.0000] & 0.9054 [0.3369,\allowbreak{} 2.0125] & 0.9054 [0.3369,\allowbreak{} 2.0125] & 90 \\
codex & C probability & 1.0000 [1.0000,\allowbreak{} 1.0001] & 1.0000 [1.0000,\allowbreak{} 1.0000] & -\allowbreak{}0.0001 [-\allowbreak{}0.0001,\allowbreak{} -\allowbreak{}0.0000] & 20 \\
codex & C frequency & 1.0001 [1.0001,\allowbreak{} 1.0001] & 1.0000 [1.0000,\allowbreak{} 1.0000] & -\allowbreak{}0.0001 [-\allowbreak{}0.0001,\allowbreak{} -\allowbreak{}0.0001] & 20 \\
codex & C error & 0.0017 [0.0016,\allowbreak{} 0.0018] & 0.0002 [0.0001,\allowbreak{} 0.0003] & -\allowbreak{}0.0016 [-\allowbreak{}0.0017,\allowbreak{} -\allowbreak{}0.0014] & 20 \\
codex & B2 supplemental endpoint & 48.2556 [44.1444,\allowbreak{} 52.0278] & not a format mean & see endpoint definition & 90 \\
codex & C1 supplemental endpoint & -\allowbreak{}0.0000 [-\allowbreak{}0.0001,\allowbreak{} -\allowbreak{}0.0000] & not a format mean & see endpoint definition & 20 \\
codex & C2 supplemental endpoint & 0.0000 [0.0000,\allowbreak{} 0.0000] & not a format mean & see endpoint definition & 20 \\
Sonnet 5.5 & A AI\_\allowbreak{}irrelevant & 0.0185 [-\allowbreak{}0.0663,\allowbreak{} 0.0970] & -\allowbreak{}0.0383 [-\allowbreak{}0.1460,\allowbreak{} 0.0464] & -\allowbreak{}0.0568 [-\allowbreak{}0.1857,\allowbreak{} 0.0501] & 50 \\
Sonnet 5.5 & A AI\_\allowbreak{}informative & 0.4899 [0.3788,\allowbreak{} 0.6015] & 0.5641 [0.3982,\allowbreak{} 0.8048] & 0.0742 [-\allowbreak{}0.0716,\allowbreak{} 0.3065] & 50 \\
Sonnet 5.5 & A S & 0.4714 [0.3408,\allowbreak{} 0.6094] & 0.6024 [0.4069,\allowbreak{} 0.8564] & 0.1309 [-\allowbreak{}0.0491,\allowbreak{} 0.3761] & 50 \\
Sonnet 5.5 & B b & -\allowbreak{}3.1694 [-\allowbreak{}4.7722,\allowbreak{} -\allowbreak{}1.6333] & -\allowbreak{}1.4199 [-\allowbreak{}2.7041,\allowbreak{} -\allowbreak{}0.2263] & 1.7496 [0.6135,\allowbreak{} 2.8356] & 90 \\
Sonnet 5.5 & B e\_\allowbreak{}PO & 24.9333 [20.8667,\allowbreak{} 29.1889] & 20.7408 [17.1964,\allowbreak{} 24.5289] & -\allowbreak{}4.1926 [-\allowbreak{}6.6464,\allowbreak{} -\allowbreak{}1.7923] & 90 \\
Sonnet 5.5 & C probability & 1.0001 [1.0000,\allowbreak{} 1.0001] & 0.9392 [0.9309,\allowbreak{} 0.9473] & -\allowbreak{}0.0609 [-\allowbreak{}0.0691,\allowbreak{} -\allowbreak{}0.0527] & 20 \\
Sonnet 5.5 & C frequency & 1.0001 [1.0001,\allowbreak{} 1.0001] & 0.9434 [0.9367,\allowbreak{} 0.9511] & -\allowbreak{}0.0567 [-\allowbreak{}0.0634,\allowbreak{} -\allowbreak{}0.0490] & 20 \\
Sonnet 5.5 & C error & 0.0024 [0.0021,\allowbreak{} 0.0027] & 1.0082 [0.8906,\allowbreak{} 1.1242] & 1.0058 [0.8882,\allowbreak{} 1.1214] & 20 \\
Sonnet 5.5 & B2 supplemental endpoint & 3.9306 [1.6778,\allowbreak{} 6.3444] & not a format mean & see endpoint definition & 90 \\
Sonnet 5.5 & C1 supplemental endpoint & -\allowbreak{}0.0001 [-\allowbreak{}0.0001,\allowbreak{} -\allowbreak{}0.0000] & not a format mean & see endpoint definition & 20 \\
Sonnet 5.5 & C2 supplemental endpoint & 0.0000 [0.0000,\allowbreak{} 0.0001] & not a format mean & see endpoint definition & 20 \\
Qwen3 & A AI\_\allowbreak{}irrelevant & 0.9456 [-\allowbreak{}0.1951,\allowbreak{} 2.2417] & 0.3377 [-\allowbreak{}0.1843,\allowbreak{} 1.0605] & -\allowbreak{}0.6079 [-\allowbreak{}1.9899,\allowbreak{} 0.7651] & 50 \\
Qwen3 & A AI\_\allowbreak{}informative & -\allowbreak{}0.3706 [-\allowbreak{}1.4018,\allowbreak{} 0.4421] & 0.5976 [0.3781,\allowbreak{} 0.8066] & 0.9682 [0.1846,\allowbreak{} 1.9290] & 50 \\
Qwen3 & A S & -\allowbreak{}1.3162 [-\allowbreak{}2.8477,\allowbreak{} 0.0430] & 0.2599 [-\allowbreak{}0.5114,\allowbreak{} 0.8378] & 1.5761 [-\allowbreak{}0.1316,\allowbreak{} 3.3604] & 50 \\
Qwen3 & B b & -\allowbreak{}7.7506 [-\allowbreak{}12.2283,\allowbreak{} -\allowbreak{}3.2011] & -\allowbreak{}2.0294 [-\allowbreak{}5.3361,\allowbreak{} 1.2062] & 5.7212 [0.6766,\allowbreak{} 10.7569] & 90 \\
Qwen3 & B e\_\allowbreak{}PO & 6.0389 [3.7056,\allowbreak{} 8.8111] & 25.0916 [21.8431,\allowbreak{} 28.4780] & 19.0527 [15.6001,\allowbreak{} 22.6239] & 90 \\
Qwen3 & C probability & 1.0017 [1.0005,\allowbreak{} 1.0030] & 0.9894 [0.9847,\allowbreak{} 0.9939] & -\allowbreak{}0.0123 [-\allowbreak{}0.0171,\allowbreak{} -\allowbreak{}0.0076] & 20 \\
Qwen3 & C frequency & 1.0004 [0.9993,\allowbreak{} 1.0017] & 0.9883 [0.9853,\allowbreak{} 0.9911] & -\allowbreak{}0.0121 [-\allowbreak{}0.0152,\allowbreak{} -\allowbreak{}0.0092] & 19 \\
Qwen3 & C error & 0.0220 [0.0110,\allowbreak{} 0.0337] & 0.6693 [0.5572,\allowbreak{} 0.7843] & 0.6473 [0.5373,\allowbreak{} 0.7611] & 20 \\
Qwen3 & B2 supplemental endpoint & 54.1278 [49.8611,\allowbreak{} 58.5833] & not a format mean & see endpoint definition & 90 \\
Qwen3 & C1 supplemental endpoint & -\allowbreak{}0.0017 [-\allowbreak{}0.0030,\allowbreak{} -\allowbreak{}0.0005] & not a format mean & see endpoint definition & 20 \\
Qwen3 & C2 supplemental endpoint & -\allowbreak{}0.0013 [-\allowbreak{}0.0027,\allowbreak{} -\allowbreak{}0.0000] & not a format mean & see endpoint definition & 19 \\
gpt-\allowbreak{}oss-\allowbreak{}120b & A AI\_\allowbreak{}irrelevant & 0.0392 [-\allowbreak{}0.9207,\allowbreak{} 0.9738] & 0.1968 [0.0334,\allowbreak{} 0.4177] & 0.1576 [-\allowbreak{}0.7893,\allowbreak{} 1.0983] & 50 \\
gpt-\allowbreak{}oss-\allowbreak{}120b & A AI\_\allowbreak{}informative & -\allowbreak{}0.0168 [-\allowbreak{}0.8497,\allowbreak{} 0.4916] & 0.2807 [0.1285,\allowbreak{} 0.4300] & 0.2975 [-\allowbreak{}0.2785,\allowbreak{} 1.1478] & 49 \\
gpt-\allowbreak{}oss-\allowbreak{}120b & A S & -\allowbreak{}0.0539 [-\allowbreak{}1.2572,\allowbreak{} 1.1360] & 0.0927 [-\allowbreak{}0.2274,\allowbreak{} 0.3226] & 0.1466 [-\allowbreak{}1.1240,\allowbreak{} 1.4502] & 49 \\
gpt-\allowbreak{}oss-\allowbreak{}120b & B b & 0.0281 [-\allowbreak{}2.7640,\allowbreak{} 2.8371] & 6.1715 [3.5615,\allowbreak{} 8.7524] & 6.1434 [3.1982,\allowbreak{} 9.2360] & 89 \\
gpt-\allowbreak{}oss-\allowbreak{}120b & B e\_\allowbreak{}PO & 7.2433 [4.5579,\allowbreak{} 10.3933] & 17.9221 [15.2253,\allowbreak{} 20.6850] & 10.6788 [7.6066,\allowbreak{} 13.8052] & 89 \\
gpt-\allowbreak{}oss-\allowbreak{}120b & C probability & 1.0001 [1.0001,\allowbreak{} 1.0002] & 0.9976 [0.9952,\allowbreak{} 0.9997] & -\allowbreak{}0.0026 [-\allowbreak{}0.0049,\allowbreak{} -\allowbreak{}0.0005] & 20 \\
gpt-\allowbreak{}oss-\allowbreak{}120b & C frequency & 1.0001 [1.0001,\allowbreak{} 1.0001] & 0.9985 [0.9956,\allowbreak{} 1.0009] & -\allowbreak{}0.0016 [-\allowbreak{}0.0044,\allowbreak{} 0.0008] & 20 \\
gpt-\allowbreak{}oss-\allowbreak{}120b & C error & 0.0030 [0.0026,\allowbreak{} 0.0033] & 0.1936 [0.1584,\allowbreak{} 0.2346] & 0.1906 [0.1554,\allowbreak{} 0.2315] & 20 \\
gpt-\allowbreak{}oss-\allowbreak{}120b & B2 supplemental endpoint & 49.6118 [46.1966,\allowbreak{} 52.9882] & not a format mean & see endpoint definition & 89 \\
gpt-\allowbreak{}oss-\allowbreak{}120b & C1 supplemental endpoint & -\allowbreak{}0.0001 [-\allowbreak{}0.0002,\allowbreak{} -\allowbreak{}0.0001] & not a format mean & see endpoint definition & 20 \\
gpt-\allowbreak{}oss-\allowbreak{}120b & C2 supplemental endpoint & -\allowbreak{}0.0001 [-\allowbreak{}0.0001,\allowbreak{} -\allowbreak{}0.0000] & not a format mean & see endpoint definition & 20 \\
Flash low & A AI\_\allowbreak{}irrelevant & 0.0416 [-\allowbreak{}0.0479,\allowbreak{} 0.1867] & 0.0796 [-\allowbreak{}0.0384,\allowbreak{} 0.2330] & 0.0380 [-\allowbreak{}0.0359,\allowbreak{} 0.1452] & 50 \\
Flash low & A AI\_\allowbreak{}informative & 0.2208 [0.1281,\allowbreak{} 0.3230] & 0.2411 [0.1047,\allowbreak{} 0.4118] & 0.0203 [-\allowbreak{}0.1265,\allowbreak{} 0.1891] & 50 \\
Flash low & A S & 0.1792 [0.0385,\allowbreak{} 0.3094] & 0.1614 [-\allowbreak{}0.0215,\allowbreak{} 0.3537] & -\allowbreak{}0.0177 [-\allowbreak{}0.2067,\allowbreak{} 0.1799] & 50 \\
Flash low & B b & -\allowbreak{}1.9612 [-\allowbreak{}4.0393,\allowbreak{} -\allowbreak{}0.0393] & -\allowbreak{}0.3180 [-\allowbreak{}2.2554,\allowbreak{} 1.5576] & 1.6432 [0.4099,\allowbreak{} 2.9430] & 89 \\
Flash low & B e\_\allowbreak{}PO & 6.1944 [3.1389,\allowbreak{} 9.6389] & 8.7143 [5.5196,\allowbreak{} 12.1089] & 2.5199 [-\allowbreak{}0.5135,\allowbreak{} 5.5701] & 90 \\
Flash low & C probability & 1.0001 [1.0001,\allowbreak{} 1.0001] & 0.9999 [0.9998,\allowbreak{} 1.0000] & -\allowbreak{}0.0001 [-\allowbreak{}0.0003,\allowbreak{} -\allowbreak{}0.0001] & 20 \\
Flash low & C frequency & 1.0001 [1.0001,\allowbreak{} 1.0001] & 1.0001 [1.0000,\allowbreak{} 1.0001] & 0.0000 [-\allowbreak{}0.0000,\allowbreak{} 0.0001] & 20 \\
Flash low & C error & 0.0018 [0.0018,\allowbreak{} 0.0018] & 0.0064 [0.0047,\allowbreak{} 0.0087] & 0.0046 [0.0029,\allowbreak{} 0.0069] & 20 \\
Flash low & B2 supplemental endpoint & 10.9667 [7.2222,\allowbreak{} 14.8444] & not a format mean & see endpoint definition & 90 \\
Flash low & C1 supplemental endpoint & -\allowbreak{}0.0001 [-\allowbreak{}0.0001,\allowbreak{} -\allowbreak{}0.0001] & not a format mean & see endpoint definition & 20 \\
Flash low & C2 supplemental endpoint & 0.0000 [0.0000,\allowbreak{} 0.0000] & not a format mean & see endpoint definition & 20 \\
Flash high & A AI\_\allowbreak{}irrelevant & 0.0344 [0.0058,\allowbreak{} 0.0724] & -\allowbreak{}0.0131 [-\allowbreak{}0.0545,\allowbreak{} 0.0247] & -\allowbreak{}0.0475 [-\allowbreak{}0.1066,\allowbreak{} -\allowbreak{}0.0024] & 50 \\
Flash high & A AI\_\allowbreak{}informative & 0.0085 [-\allowbreak{}0.2115,\allowbreak{} 0.1641] & 0.1579 [0.0751,\allowbreak{} 0.2505] & 0.1494 [0.0082,\allowbreak{} 0.3948] & 50 \\
Flash high & A S & -\allowbreak{}0.0259 [-\allowbreak{}0.2456,\allowbreak{} 0.1243] & 0.1710 [0.0743,\allowbreak{} 0.2822] & 0.1969 [0.0305,\allowbreak{} 0.4770] & 50 \\
Flash high & B b & -\allowbreak{}3.9888 [-\allowbreak{}5.9719,\allowbreak{} -\allowbreak{}2.1573] & 0.5376 [-\allowbreak{}0.9949,\allowbreak{} 2.0855] & 4.5263 [3.0365,\allowbreak{} 6.1252] & 89 \\
Flash high & B e\_\allowbreak{}PO & 24.3056 [19.3889,\allowbreak{} 29.2222] & 25.7084 [21.8358,\allowbreak{} 29.5567] & 1.4029 [-\allowbreak{}1.9089,\allowbreak{} 4.8196] & 90 \\
Flash high & C probability & 1.0001 [1.0001,\allowbreak{} 1.0001] & 1.0001 [1.0000,\allowbreak{} 1.0002] & 0.0001 [-\allowbreak{}0.0000,\allowbreak{} 0.0002] & 20 \\
Flash high & C frequency & 1.0001 [1.0001,\allowbreak{} 1.0001] & 1.0001 [1.0000,\allowbreak{} 1.0002] & 0.0000 [-\allowbreak{}0.0001,\allowbreak{} 0.0001] & 20 \\
Flash high & C error & 0.0018 [0.0018,\allowbreak{} 0.0018] & 0.0072 [0.0046,\allowbreak{} 0.0114] & 0.0054 [0.0028,\allowbreak{} 0.0096] & 20 \\
Flash high & B2 supplemental endpoint & 0.8500 [0.0444,\allowbreak{} 1.7611] & not a format mean & see endpoint definition & 90 \\
Flash high & C1 supplemental endpoint & -\allowbreak{}0.0001 [-\allowbreak{}0.0001,\allowbreak{} -\allowbreak{}0.0001] & not a format mean & see endpoint definition & 20 \\
Flash high & C2 supplemental endpoint & 0.0000 [0.0000,\allowbreak{} 0.0000] & not a format mean & see endpoint definition & 20 \\
\end{longtable}
\endgroup

% END A:485

% BEGIN A:554 text

\textbf{Table H4. D,\allowbreak{} E descriptive summaries across six configurations}\par

% END A:554

% BEGIN A:556 text

Unadjusted descriptive cluster-\allowbreak{}bootstrap intervals; complete paired subsets can differ from single-\allowbreak{}format endpoint denominators. Supplemental B2 is HP minus HI; C1 is 1 minus direct probability slope; C2 is direct frequency minus probability slope; neither is itself a direct probability slope. Supplemental endpoints remain descriptive when outside the frozen list.\par

% END A:556

% BEGIN A:558 table

\begingroup\scriptsize\setlength{\tabcolsep}{1.3mm}\setlength{\LTpre}{4pt}\setlength{\LTpost}{6pt}
\begin{longtable}{>{\raggedright\arraybackslash}p{16.846mm}>{\raggedright\arraybackslash}p{18.628mm}>{\raggedright\arraybackslash}p{33.693mm}>{\raggedright\arraybackslash}p{29.805mm}>{\raggedright\arraybackslash}p{34.989mm}>{\raggedright\arraybackslash}p{19.438mm}}
\toprule
Configuration & Quantity & Direct /\allowbreak{} endpoint [95\% CI] & Multi-\allowbreak{}estimate [95\% CI] & Multi minus direct [95\% CI] & Paired clusters \\
\midrule\endfirsthead
\toprule
Configuration & Quantity & Direct /\allowbreak{} endpoint [95\% CI] & Multi-\allowbreak{}estimate [95\% CI] & Multi minus direct [95\% CI] & Paired clusters \\
\midrule\endhead
\bottomrule\endfoot
codex & D delta\_\allowbreak{}S & -\allowbreak{}1.4250 [-\allowbreak{}3.8500,\allowbreak{} 0.3750] & -\allowbreak{}2.1658 [-\allowbreak{}4.3306,\allowbreak{} -\allowbreak{}0.3609] & -\allowbreak{}0.7409 [-\allowbreak{}1.7079,\allowbreak{} 0.2306] & 20 \\
codex & D delta\_\allowbreak{}J & -\allowbreak{}3.8825 [-\allowbreak{}6.3075,\allowbreak{} -\allowbreak{}2.0325] & -\allowbreak{}3.7396 [-\allowbreak{}5.9672,\allowbreak{} -\allowbreak{}1.9636] & 0.1429 [-\allowbreak{}0.1341,\allowbreak{} 0.4459] & 20 \\
codex & E E\_\allowbreak{}status & -\allowbreak{}0.3125 [-\allowbreak{}0.9375,\allowbreak{} 0.0000] & -\allowbreak{}0.6953 [-\allowbreak{}2.3984,\allowbreak{} 0.8906] & -\allowbreak{}0.3828 [-\allowbreak{}2.0859,\allowbreak{} 1.2344] & 32 \\
codex & E E\_\allowbreak{}default & 0.0000 [0.0000,\allowbreak{} 0.0000] & -\allowbreak{}0.0234 [-\allowbreak{}1.5547,\allowbreak{} 1.5234] & -\allowbreak{}0.0234 [-\allowbreak{}1.5547,\allowbreak{} 1.5234] & 32 \\
codex & E E\_\allowbreak{}recommend & 49.8438 [48.2812,\allowbreak{} 51.5625] & 50.8125 [48.5234,\allowbreak{} 53.0859] & 0.9688 [-\allowbreak{}1.0000,\allowbreak{} 2.9453] & 32 \\
codex & E neutral direct & 50.625 & n=32 & descriptive & 32 \\
codex & E advice control direct & 49.844 [48.281,\allowbreak{} 51.562] & missing bounds [49.844,\allowbreak{} 49.844] & downgrade=False & 32 \\
codex & E neutral multi & 49.922 & n=32 & descriptive & 32 \\
codex & E advice control multi & 50.812 [48.523,\allowbreak{} 53.086] & missing bounds [50.812,\allowbreak{} 50.812] & downgrade=False & 32 \\
codex & D2 supplemental S minus J & 2.458 [1.175,\allowbreak{} 3.875] & direct comparison & not a format contrast & 20 \\
Sonnet 5.5 & D delta\_\allowbreak{}S & -\allowbreak{}1.4800 [-\allowbreak{}3.7300,\allowbreak{} 0.0800] & -\allowbreak{}1.8965 [-\allowbreak{}4.5800,\allowbreak{} -\allowbreak{}0.0055] & -\allowbreak{}0.4165 [-\allowbreak{}1.5597,\allowbreak{} 0.5083] & 20 \\
Sonnet 5.5 & D delta\_\allowbreak{}J & -\allowbreak{}3.7800 [-\allowbreak{}6.4300,\allowbreak{} -\allowbreak{}1.8550] & -\allowbreak{}3.8750 [-\allowbreak{}6.7585,\allowbreak{} -\allowbreak{}1.8514] & -\allowbreak{}0.0950 [-\allowbreak{}0.4385,\allowbreak{} 0.2107] & 20 \\
Sonnet 5.5 & E E\_\allowbreak{}status & 0.3438 [-\allowbreak{}1.2500,\allowbreak{} 1.8438] & 0.0453 [-\allowbreak{}1.8969,\allowbreak{} 1.9297] & -\allowbreak{}0.2984 [-\allowbreak{}3.0656,\allowbreak{} 2.4234] & 32 \\
Sonnet 5.5 & E E\_\allowbreak{}default & -\allowbreak{}1.4062 [-\allowbreak{}3.1250,\allowbreak{} 0.2188] & -\allowbreak{}1.0175 [-\allowbreak{}3.4947,\allowbreak{} 1.1403] & 0.3888 [-\allowbreak{}2.5297,\allowbreak{} 3.0953] & 32 \\
Sonnet 5.5 & E E\_\allowbreak{}recommend & 52.9688 [50.0312,\allowbreak{} 55.8750] & 49.0409 [47.0156,\allowbreak{} 51.0678] & -\allowbreak{}3.9278 [-\allowbreak{}5.7788,\allowbreak{} -\allowbreak{}2.0466] & 32 \\
Sonnet 5.5 & E neutral direct & 43.656 & n=32 & descriptive & 32 \\
Sonnet 5.5 & E advice control direct & 52.969 [50.031,\allowbreak{} 55.875] & missing bounds [52.969,\allowbreak{} 52.969] & downgrade=False & 32 \\
Sonnet 5.5 & E neutral multi & 43.660 & n=32 & descriptive & 32 \\
Sonnet 5.5 & E advice control multi & 49.041 [47.016,\allowbreak{} 51.068] & missing bounds [49.041,\allowbreak{} 49.041] & downgrade=False & 32 \\
Sonnet 5.5 & D2 supplemental S minus J & 2.300 [1.305,\allowbreak{} 3.490] & direct comparison & not a format contrast & 20 \\
Qwen3 & D delta\_\allowbreak{}S & 13.0250 [8.0000,\allowbreak{} 19.0750] & 8.4120 [1.9751,\allowbreak{} 15.1723] & -\allowbreak{}4.6130 [-\allowbreak{}12.9365,\allowbreak{} 3.6095] & 20 \\
Qwen3 & D delta\_\allowbreak{}J & -\allowbreak{}2.7500 [-\allowbreak{}7.5000,\allowbreak{} 1.4000] & -\allowbreak{}5.3333 [-\allowbreak{}9.3099,\allowbreak{} -\allowbreak{}1.7365] & -\allowbreak{}2.5833 [-\allowbreak{}5.9837,\allowbreak{} 0.7263] & 20 \\
Qwen3 & E E\_\allowbreak{}status & -\allowbreak{}3.3142 [-\allowbreak{}11.3265,\allowbreak{} 4.3013] & 0.7275 [-\allowbreak{}3.6466,\allowbreak{} 4.8992] & 4.0417 [-\allowbreak{}5.5677,\allowbreak{} 14.3247] & 31 \\
Qwen3 & E E\_\allowbreak{}default & 0.1021 [-\allowbreak{}9.1462,\allowbreak{} 7.8159] & 5.7613 [-\allowbreak{}0.5824,\allowbreak{} 11.8820] & 5.6592 [-\allowbreak{}4.1036,\allowbreak{} 15.6847] & 29 \\
Qwen3 & E E\_\allowbreak{}recommend & 63.9616 [50.3052,\allowbreak{} 77.0968] & 36.0131 [31.0953,\allowbreak{} 40.4462] & -\allowbreak{}27.9485 [-\allowbreak{}39.7935,\allowbreak{} -\allowbreak{}15.7437] & 31 \\
Qwen3 & E neutral direct & 53.359 & n=32 & descriptive & 32 \\
Qwen3 & E advice control direct & 64.932 [51.046,\allowbreak{} 77.323] & missing bounds [64.932,\allowbreak{} 64.932] & downgrade=False & 32 \\
Qwen3 & E neutral multi & 45.695 & n=32 & descriptive & 32 \\
Qwen3 & E advice control multi & 36.013 [31.095,\allowbreak{} 40.446] & missing bounds [33.544,\allowbreak{} 36.669] & downgrade=False & 31 \\
Qwen3 & D2 supplemental S minus J & 15.775 [9.400,\allowbreak{} 22.900] & direct comparison & not a format contrast & 20 \\
gpt-\allowbreak{}oss-\allowbreak{}120b & D delta\_\allowbreak{}S & 9.6000 [3.2000,\allowbreak{} 17.1000] & 6.9050 [3.9825,\allowbreak{} 10.3775] & -\allowbreak{}2.6950 [-\allowbreak{}10.6725,\allowbreak{} 4.2975] & 20 \\
gpt-\allowbreak{}oss-\allowbreak{}120b & D delta\_\allowbreak{}J & -\allowbreak{}7.7500 [-\allowbreak{}12.4500,\allowbreak{} -\allowbreak{}4.1000] & -\allowbreak{}9.7134 [-\allowbreak{}12.9747,\allowbreak{} -\allowbreak{}7.1175] & -\allowbreak{}1.9634 [-\allowbreak{}4.2893,\allowbreak{} 0.4016] & 20 \\
gpt-\allowbreak{}oss-\allowbreak{}120b & E E\_\allowbreak{}status & -\allowbreak{}1.1334 [-\allowbreak{}7.5606,\allowbreak{} 5.4844] & 3.9875 [0.9813,\allowbreak{} 7.0375] & 5.1209 [-\allowbreak{}2.0831,\allowbreak{} 12.3959] & 32 \\
gpt-\allowbreak{}oss-\allowbreak{}120b & E E\_\allowbreak{}default & 1.4781 [-\allowbreak{}3.2188,\allowbreak{} 6.5719] & 4.6219 [0.3625,\allowbreak{} 8.7203] & 3.1438 [-\allowbreak{}2.0422,\allowbreak{} 8.0203] & 32 \\
gpt-\allowbreak{}oss-\allowbreak{}120b & E E\_\allowbreak{}recommend & 50.3438 [42.1250,\allowbreak{} 57.3750] & 27.1091 [22.1556,\allowbreak{} 31.8578] & -\allowbreak{}23.2347 [-\allowbreak{}31.3428,\allowbreak{} -\allowbreak{}14.1425] & 32 \\
gpt-\allowbreak{}oss-\allowbreak{}120b & E neutral direct & 42.672 & n=32 & descriptive & 32 \\
gpt-\allowbreak{}oss-\allowbreak{}120b & E advice control direct & 50.344 [42.125,\allowbreak{} 57.375] & missing bounds [50.344,\allowbreak{} 50.344] & downgrade=False & 32 \\
gpt-\allowbreak{}oss-\allowbreak{}120b & E neutral multi & 49.591 & n=32 & descriptive & 32 \\
gpt-\allowbreak{}oss-\allowbreak{}120b & E advice control multi & 27.109 [22.156,\allowbreak{} 31.858] & missing bounds [27.109,\allowbreak{} 27.109] & downgrade=False & 32 \\
gpt-\allowbreak{}oss-\allowbreak{}120b & D2 supplemental S minus J & 17.350 [9.750,\allowbreak{} 25.550] & direct comparison & not a format contrast & 20 \\
Flash low & D delta\_\allowbreak{}S & 0.0850 [-\allowbreak{}1.0150,\allowbreak{} 1.4950] & 0.5722 [-\allowbreak{}1.6420,\allowbreak{} 2.9412] & 0.4872 [-\allowbreak{}1.6540,\allowbreak{} 2.7748] & 20 \\
Flash low & D delta\_\allowbreak{}J & -\allowbreak{}3.2875 [-\allowbreak{}5.5700,\allowbreak{} -\allowbreak{}1.5275] & -\allowbreak{}3.4545 [-\allowbreak{}6.2085,\allowbreak{} -\allowbreak{}1.5231] & -\allowbreak{}0.1670 [-\allowbreak{}0.7720,\allowbreak{} 0.3549] & 20 \\
Flash low & E E\_\allowbreak{}status & -\allowbreak{}0.5312 [-\allowbreak{}1.6250,\allowbreak{} 0.1875] & -\allowbreak{}1.2622 [-\allowbreak{}3.1075,\allowbreak{} 0.6453] & -\allowbreak{}0.7309 [-\allowbreak{}2.8325,\allowbreak{} 1.5188] & 32 \\
Flash low & E E\_\allowbreak{}default & 0.0000 [0.0000,\allowbreak{} 0.0000] & 1.6453 [-\allowbreak{}0.1828,\allowbreak{} 3.5578] & 1.6453 [-\allowbreak{}0.1828,\allowbreak{} 3.5578] & 32 \\
Flash low & E E\_\allowbreak{}recommend & 65.6875 [59.6875,\allowbreak{} 71.0938] & 59.2622 [56.6569,\allowbreak{} 61.8575] & -\allowbreak{}6.4253 [-\allowbreak{}11.4000,\allowbreak{} -\allowbreak{}0.2187] & 32 \\
Flash low & E neutral direct & 46.875 & n=32 & descriptive & 32 \\
Flash low & E advice control direct & 65.688 [59.688,\allowbreak{} 71.094] & missing bounds [65.688,\allowbreak{} 65.688] & downgrade=False & 32 \\
Flash low & E neutral multi & 53.308 & n=32 & descriptive & 32 \\
Flash low & E advice control multi & 59.262 [56.657,\allowbreak{} 61.858] & missing bounds [59.262,\allowbreak{} 59.262] & downgrade=False & 32 \\
Flash low & D2 supplemental S minus J & 3.373 [1.137,\allowbreak{} 6.130] & direct comparison & not a format contrast & 20 \\
Flash high & D delta\_\allowbreak{}S & -\allowbreak{}1.1150 [-\allowbreak{}2.1150,\allowbreak{} -\allowbreak{}0.2150] & -\allowbreak{}0.1357 [-\allowbreak{}1.9650,\allowbreak{} 1.4325] & 0.9793 [-\allowbreak{}0.7268,\allowbreak{} 2.6888] & 20 \\
Flash high & D delta\_\allowbreak{}J & -\allowbreak{}4.4500 [-\allowbreak{}7.3250,\allowbreak{} -\allowbreak{}2.3000] & -\allowbreak{}3.6920 [-\allowbreak{}6.6698,\allowbreak{} -\allowbreak{}1.6965] & 0.7580 [-\allowbreak{}0.1690,\allowbreak{} 2.0015] & 20 \\
Flash high & E E\_\allowbreak{}status & 0.6250 [0.0000,\allowbreak{} 1.8750] & -\allowbreak{}1.5625 [-\allowbreak{}3.5766,\allowbreak{} 0.4328] & -\allowbreak{}2.1875 [-\allowbreak{}4.4375,\allowbreak{} 0.0750] & 32 \\
Flash high & E E\_\allowbreak{}default & 0.0000 [0.0000,\allowbreak{} 0.0000] & 0.1419 [-\allowbreak{}1.5772,\allowbreak{} 1.8969] & 0.1419 [-\allowbreak{}1.5772,\allowbreak{} 1.8969] & 32 \\
Flash high & E E\_\allowbreak{}recommend & 71.8750 [62.8125,\allowbreak{} 80.1562] & 58.4344 [56.1047,\allowbreak{} 60.6156] & -\allowbreak{}13.4406 [-\allowbreak{}21.4906,\allowbreak{} -\allowbreak{}4.4484] & 32 \\
Flash high & E neutral direct & 36.875 & n=32 & descriptive & 32 \\
Flash high & E advice control direct & 71.875 [62.812,\allowbreak{} 80.156] & missing bounds [71.875,\allowbreak{} 71.875] & downgrade=False & 32 \\
Flash high & E neutral multi & 53.511 & n=32 & descriptive & 32 \\
Flash high & E advice control multi & 58.434 [56.105,\allowbreak{} 60.616] & missing bounds [58.434,\allowbreak{} 58.434] & downgrade=False & 32 \\
Flash high & D2 supplemental S minus J & 3.335 [1.245,\allowbreak{} 5.865] & direct comparison & not a format contrast & 20 \\
\end{longtable}
\endgroup

% END A:558

% BEGIN A:621 text

\textbf{Table H5. Repeat consistency:\allowbreak{} 484 calls and 520 planned numeric pairs per configuration}\par

% END A:621

% BEGIN A:623 text

D joint calls each supply four dependent numeric statement pairs; they are not four independent calls. A SDs are on raw quantity scales,\allowbreak{} incomparable across items. Qwen3 loses one E multi repeat pair because its r1 unit is missing; a missing primary C cell can also remove a C pair.\par

% END A:623

% BEGIN A:625 table

\begingroup\scriptsize\setlength{\tabcolsep}{1.3mm}\setlength{\LTpre}{4pt}\setlength{\LTpost}{6pt}
\begin{longtable}{>{\raggedright\arraybackslash}p{19.266mm}>{\raggedright\arraybackslash}p{7.410mm}>{\raggedright\arraybackslash}p{8.892mm}>{\raggedright\arraybackslash}p{19.266mm}>{\raggedright\arraybackslash}p{19.266mm}>{\raggedright\arraybackslash}p{22.230mm}>{\raggedright\arraybackslash}p{23.712mm}>{\raggedright\arraybackslash}p{28.158mm}}
\toprule
Configuration & Exp & Format & Planned calls & Planned pairs & Available pairs & Zero differences & Median cell call SD \\
\midrule\endfirsthead
\toprule
Configuration & Exp & Format & Planned calls & Planned pairs & Available pairs & Zero differences & Median cell call SD \\
\midrule\endhead
\bottomrule\endfoot
codex & A & direct & 40 & 40 & 40 & 37 & 0.000 \\
codex & A & multi & 40 & 40 & 40 & 11 & 4687.502 \\
codex & B & direct & 96 & 96 & 96 & 70 & 3.125 \\
codex & B & multi & 96 & 96 & 96 & 32 & 1.994 \\
codex & C & direct & 48 & 48 & 48 & 46 & 0.000 \\
codex & C & multi & 48 & 48 & 48 & 36 & 0.000 \\
codex & D & direct & 30 & 48 & 48 & 26 & 2.189 \\
codex & D & multi & 30 & 48 & 48 & 7 & 2.303 \\
codex & E & direct & 28 & 28 & 28 & 24 & 0.000 \\
codex & E & multi & 28 & 28 & 28 & 13 & 2.129 \\
Sonnet 5.5 & A & direct & 40 & 40 & 40 & 24 & 12.787 \\
Sonnet 5.5 & A & multi & 40 & 40 & 40 & 1 & 11222500.000 \\
Sonnet 5.5 & B & direct & 96 & 96 & 96 & 45 & 3.206 \\
Sonnet 5.5 & B & multi & 96 & 96 & 96 & 6 & 3.889 \\
Sonnet 5.5 & C & direct & 48 & 48 & 48 & 41 & 0.002 \\
Sonnet 5.5 & C & multi & 48 & 48 & 48 & 4 & 0.715 \\
Sonnet 5.5 & D & direct & 30 & 48 & 48 & 17 & 1.800 \\
Sonnet 5.5 & D & multi & 30 & 48 & 48 & 0 & 1.367 \\
Sonnet 5.5 & E & direct & 28 & 28 & 28 & 9 & 2.761 \\
Sonnet 5.5 & E & multi & 28 & 28 & 28 & 1 & 1.823 \\
Qwen3 & A & direct & 40 & 40 & 40 & 23 & 1305041.211 \\
Qwen3 & A & multi & 40 & 40 & 40 & 3 & 55001.998 \\
Qwen3 & B & direct & 96 & 96 & 96 & 56 & 3.506 \\
Qwen3 & B & multi & 96 & 96 & 96 & 9 & 13.987 \\
Qwen3 & C & direct & 48 & 48 & 48 & 44 & 0.041 \\
Qwen3 & C & multi & 48 & 48 & 48 & 0 & 0.517 \\
Qwen3 & D & direct & 30 & 48 & 48 & 22 & 6.887 \\
Qwen3 & D & multi & 30 & 48 & 48 & 8 & 6.219 \\
Qwen3 & E & direct & 28 & 28 & 28 & 10 & 22.638 \\
Qwen3 & E & multi & 28 & 28 & 27 & 2 & 6.051 \\
gpt-\allowbreak{}oss-\allowbreak{}120b & A & direct & 40 & 40 & 40 & 14 & 1711905.080 \\
gpt-\allowbreak{}oss-\allowbreak{}120b & A & multi & 40 & 40 & 40 & 0 & 50000.020 \\
gpt-\allowbreak{}oss-\allowbreak{}120b & B & direct & 96 & 96 & 96 & 36 & 8.536 \\
gpt-\allowbreak{}oss-\allowbreak{}120b & B & multi & 96 & 96 & 96 & 1 & 9.240 \\
gpt-\allowbreak{}oss-\allowbreak{}120b & C & direct & 48 & 48 & 48 & 45 & 0.000 \\
gpt-\allowbreak{}oss-\allowbreak{}120b & C & multi & 48 & 48 & 48 & 6 & 0.211 \\
gpt-\allowbreak{}oss-\allowbreak{}120b & D & direct & 30 & 48 & 48 & 11 & 5.017 \\
gpt-\allowbreak{}oss-\allowbreak{}120b & D & multi & 30 & 48 & 48 & 0 & 6.247 \\
gpt-\allowbreak{}oss-\allowbreak{}120b & E & direct & 28 & 28 & 28 & 12 & 10.452 \\
gpt-\allowbreak{}oss-\allowbreak{}120b & E & multi & 28 & 28 & 28 & 1 & 6.424 \\
Flash low & A & direct & 40 & 40 & 40 & 28 & 10.335 \\
Flash low & A & multi & 40 & 40 & 40 & 5 & 22500.002 \\
Flash low & B & direct & 96 & 96 & 96 & 62 & 4.812 \\
Flash low & B & multi & 96 & 96 & 96 & 13 & 3.803 \\
Flash low & C & direct & 48 & 48 & 48 & 48 & 0.000 \\
Flash low & C & multi & 48 & 48 & 48 & 16 & 0.005 \\
Flash low & D & direct & 30 & 48 & 48 & 17 & 2.889 \\
Flash low & D & multi & 30 & 48 & 48 & 0 & 1.694 \\
Flash low & E & direct & 28 & 28 & 28 & 26 & 0.000 \\
Flash low & E & multi & 28 & 28 & 28 & 1 & 4.058 \\
Flash high & A & direct & 40 & 40 & 40 & 32 & 10.078 \\
Flash high & A & multi & 40 & 40 & 40 & 2 & 12500.010 \\
Flash high & B & direct & 96 & 96 & 96 & 60 & 3.858 \\
Flash high & B & multi & 96 & 96 & 96 & 2 & 2.928 \\
Flash high & C & direct & 48 & 48 & 48 & 48 & 0.000 \\
Flash high & C & multi & 48 & 48 & 48 & 17 & 0.011 \\
Flash high & D & direct & 30 & 48 & 48 & 21 & 2.196 \\
Flash high & D & multi & 30 & 48 & 48 & 1 & 1.998 \\
Flash high & E & direct & 28 & 28 & 28 & 25 & 0.000 \\
Flash high & E & multi & 28 & 28 & 28 & 1 & 2.808 \\
\end{longtable}
\endgroup

% END A:625

% BEGIN A:688 text

\textbf{Table H6. External delta-\allowbreak{}NLI stated-\allowbreak{}presentation H4b comparison}\par

% END A:688

% BEGIN A:690 text

Cached results from a separate frozen study,\allowbreak{} not six new Psych3 confirmation tests. codex CLI versions differ between the experiments; cross-\allowbreak{}task comparisons are descriptive.\par

% END A:690

% BEGIN A:692 table

\begingroup\scriptsize\setlength{\tabcolsep}{1.3mm}\setlength{\LTpre}{4pt}\setlength{\LTpost}{6pt}
\begin{longtable}{>{\raggedright\arraybackslash}p{25.037mm}>{\raggedright\arraybackslash}p{23.111mm}>{\raggedright\arraybackslash}p{25.037mm}>{\raggedright\arraybackslash}p{23.111mm}>{\raggedright\arraybackslash}p{59.704mm}}
\toprule
Configuration & Paired items & Direct mean s & Multi mean s & Multi minus direct [raw 95\% CI] \\
\midrule\endfirsthead
\toprule
Configuration & Paired items & Direct mean s & Multi mean s & Multi minus direct [raw 95\% CI] \\
\midrule\endhead
\bottomrule\endfoot
codex & 120 & 25.283 & 22.801 & -\allowbreak{}2.482 [-\allowbreak{}3.181,\allowbreak{} -\allowbreak{}1.797] \\
Sonnet 5.5 & 120 & 22.008 & 22.479 & 0.471 [-\allowbreak{}0.342,\allowbreak{} 1.32] \\
Qwen3 & 240 & 18.594 & 10.655 & -\allowbreak{}7.939 [-\allowbreak{}9.716,\allowbreak{} -\allowbreak{}6.245] \\
gpt-\allowbreak{}oss-\allowbreak{}120b & 120 & 17.592 & 15.296 & -\allowbreak{}2.295 [-\allowbreak{}4.081,\allowbreak{} -\allowbreak{}0.561] \\
Flash low & 120 & 26.641 & 24.163 & -\allowbreak{}2.478 [-\allowbreak{}3.677,\allowbreak{} -\allowbreak{}1.358] \\
Flash high & 120 & 28.296 & 24.563 & -\allowbreak{}3.733 [-\allowbreak{}5.051,\allowbreak{} -\allowbreak{}2.52] \\
\end{longtable}
\endgroup

% END A:692

% BEGIN A:701 text

\textbf{Table H7. Candidate spread versus absolute response-\allowbreak{}format difference}\par

% END A:701

% BEGIN A:703 text

Post-\allowbreak{}hoc,\allowbreak{} in-\allowbreak{}sample,\allowbreak{} whole-\allowbreak{}cluster bootstrap (5,\allowbreak{}000 draws; seed 20261029). Undefined correlations are left undefined; zero valid resamples are not failed experiments. No multiplicity-\allowbreak{}adjusted or held-\allowbreak{}out predictive conclusion is drawn.\par

% END A:703

% BEGIN A:705 table

\begingroup\scriptsize\setlength{\tabcolsep}{1.3mm}\setlength{\LTpre}{4pt}\setlength{\LTpost}{6pt}
\begin{longtable}{>{\raggedright\arraybackslash}p{22.533mm}>{\raggedright\arraybackslash}p{8.667mm}>{\raggedright\arraybackslash}p{8.667mm}>{\raggedright\arraybackslash}p{13.867mm}>{\raggedright\arraybackslash}p{36.400mm}>{\raggedright\arraybackslash}p{26.000mm}>{\raggedright\arraybackslash}p{34.667mm}}
\toprule
Configuration & Exp & Items & Clusters & Spearman rho [95\% CI] & Valid resamples & Degenerate resamples \\
\midrule\endfirsthead
\toprule
Configuration & Exp & Items & Clusters & Spearman rho [95\% CI] & Valid resamples & Degenerate resamples \\
\midrule\endhead
\bottomrule\endfoot
codex & A & 50 & 50 & 0.728 [0.525,\allowbreak{} 0.869] & 5000 & 0 \\
codex & B & 90 & 90 & 0.379 [0.174,\allowbreak{} 0.557] & 5000 & 0 \\
codex & C & 40 & 20 & undefined & 0 & 0 \\
codex & D & 35 & 35 & 0.473 [0.170,\allowbreak{} 0.710] & 5000 & 0 \\
codex & E & 32 & 32 & -\allowbreak{}0.256 [-\allowbreak{}0.609,\allowbreak{} 0.155] & 5000 & 0 \\
Sonnet 5.5 & A & 50 & 50 & 0.877 [0.760,\allowbreak{} 0.934] & 5000 & 0 \\
Sonnet 5.5 & B & 90 & 90 & 0.504 [0.309,\allowbreak{} 0.670] & 5000 & 0 \\
Sonnet 5.5 & C & 40 & 20 & 0.627 [0.400,\allowbreak{} 0.786] & 5000 & 0 \\
Sonnet 5.5 & D & 35 & 35 & 0.506 [0.191,\allowbreak{} 0.745] & 5000 & 0 \\
Sonnet 5.5 & E & 32 & 32 & -\allowbreak{}0.162 [-\allowbreak{}0.485,\allowbreak{} 0.184] & 5000 & 0 \\
Qwen3 & A & 50 & 50 & 0.501 [0.250,\allowbreak{} 0.694] & 5000 & 0 \\
Qwen3 & B & 90 & 90 & 0.373 [0.174,\allowbreak{} 0.543] & 5000 & 0 \\
Qwen3 & C & 40 & 20 & 0.398 [0.162,\allowbreak{} 0.573] & 5000 & 0 \\
Qwen3 & D & 35 & 35 & 0.527 [0.220,\allowbreak{} 0.735] & 5000 & 0 \\
Qwen3 & E & 32 & 32 & -\allowbreak{}0.156 [-\allowbreak{}0.481,\allowbreak{} 0.222] & 5000 & 0 \\
gpt-\allowbreak{}oss-\allowbreak{}120b & A & 50 & 50 & 0.792 [0.620,\allowbreak{} 0.893] & 5000 & 0 \\
gpt-\allowbreak{}oss-\allowbreak{}120b & B & 90 & 90 & 0.543 [0.367,\allowbreak{} 0.683] & 5000 & 0 \\
gpt-\allowbreak{}oss-\allowbreak{}120b & C & 40 & 20 & 0.579 [0.342,\allowbreak{} 0.764] & 5000 & 0 \\
gpt-\allowbreak{}oss-\allowbreak{}120b & D & 35 & 35 & 0.608 [0.321,\allowbreak{} 0.789] & 5000 & 0 \\
gpt-\allowbreak{}oss-\allowbreak{}120b & E & 32 & 32 & 0.111 [-\allowbreak{}0.288,\allowbreak{} 0.504] & 5000 & 0 \\
Flash low & A & 50 & 50 & 0.941 [0.876,\allowbreak{} 0.968] & 5000 & 0 \\
Flash low & B & 90 & 90 & 0.571 [0.384,\allowbreak{} 0.722] & 5000 & 0 \\
Flash low & C & 40 & 20 & 0.134 [-\allowbreak{}0.204,\allowbreak{} 0.450] & 5000 & 0 \\
Flash low & D & 35 & 35 & 0.911 [0.823,\allowbreak{} 0.948] & 5000 & 0 \\
Flash low & E & 32 & 32 & 0.087 [-\allowbreak{}0.304,\allowbreak{} 0.469] & 5000 & 0 \\
Flash high & A & 50 & 50 & 0.955 [0.901,\allowbreak{} 0.975] & 5000 & 0 \\
Flash high & B & 90 & 90 & 0.642 [0.477,\allowbreak{} 0.767] & 5000 & 0 \\
Flash high & C & 40 & 20 & 0.433 [0.193,\allowbreak{} 0.644] & 5000 & 0 \\
Flash high & D & 35 & 35 & 0.669 [0.432,\allowbreak{} 0.822] & 5000 & 0 \\
Flash high & E & 32 & 32 & 0.049 [-\allowbreak{}0.337,\allowbreak{} 0.427] & 5000 & 0 \\
\end{longtable}
\endgroup

% END A:705

% BEGIN A:738 text

\textbf{Table H8. Candidate spread versus absolute multi-\allowbreak{}estimate log error in A}\par

% END A:738

% BEGIN A:740 table

\begingroup\scriptsize\setlength{\tabcolsep}{1.3mm}\setlength{\LTpre}{4pt}\setlength{\LTpost}{6pt}
\begin{longtable}{>{\raggedright\arraybackslash}p{28.087mm}>{\raggedright\arraybackslash}p{19.445mm}>{\raggedright\arraybackslash}p{10.803mm}>{\raggedright\arraybackslash}p{17.285mm}>{\raggedright\arraybackslash}p{45.372mm}>{\raggedright\arraybackslash}p{32.408mm}}
\toprule
Configuration & A framing & Items & Clusters & Spearman rho [95\% CI] & Valid resamples \\
\midrule\endfirsthead
\toprule
Configuration & A framing & Items & Clusters & Spearman rho [95\% CI] & Valid resamples \\
\midrule\endhead
\bottomrule\endfoot
codex & none & 49 & 49 & 0.836 [0.697,\allowbreak{} 0.929] & 5000 \\
codex & anchored & 50 & 50 & 0.829 [0.683,\allowbreak{} 0.910] & 5000 \\
Sonnet 5.5 & none & 50 & 50 & 0.902 [0.810,\allowbreak{} 0.944] & 5000 \\
Sonnet 5.5 & anchored & 50 & 50 & 0.876 [0.768,\allowbreak{} 0.932] & 5000 \\
Qwen3 & none & 50 & 50 & 0.566 [0.308,\allowbreak{} 0.763] & 5000 \\
Qwen3 & anchored & 50 & 50 & 0.598 [0.378,\allowbreak{} 0.754] & 5000 \\
gpt-\allowbreak{}oss-\allowbreak{}120b & none & 49 & 49 & 0.692 [0.469,\allowbreak{} 0.840] & 5000 \\
gpt-\allowbreak{}oss-\allowbreak{}120b & anchored & 50 & 50 & 0.733 [0.528,\allowbreak{} 0.856] & 5000 \\
Flash low & none & 50 & 50 & 0.798 [0.647,\allowbreak{} 0.895] & 5000 \\
Flash low & anchored & 50 & 50 & 0.900 [0.806,\allowbreak{} 0.945] & 5000 \\
Flash high & none & 50 & 50 & 0.884 [0.756,\allowbreak{} 0.947] & 5000 \\
Flash high & anchored & 50 & 50 & 0.881 [0.753,\allowbreak{} 0.946] & 5000 \\
\end{longtable}
\endgroup

% END A:740

% BEGIN A:755 text

\textbf{Table H9. Exploratory A error split at the within-\allowbreak{}sample median spread}\par

% END A:755

% BEGIN A:757 text

Lower group:\allowbreak{} spread at or below the median; upper:\allowbreak{} strictly above. Ties are not split. These are not validated deployment thresholds.\par

% END A:757

% BEGIN A:759 table

\begingroup\scriptsize\setlength{\tabcolsep}{1.3mm}\setlength{\LTpre}{4pt}\setlength{\LTpost}{6pt}
\begin{longtable}{>{\raggedright\arraybackslash}p{16.756mm}>{\raggedright\arraybackslash}p{11.600mm}>{\raggedright\arraybackslash}p{16.756mm}>{\raggedright\arraybackslash}p{21.911mm}>{\raggedright\arraybackslash}p{30.933mm}>{\raggedright\arraybackslash}p{21.911mm}>{\raggedright\arraybackslash}p{30.933mm}}
\toprule
Configuration & A framing & Median cutoff & Lower-\allowbreak{}group items & Lower mean abs log error & Upper-\allowbreak{}group items & Upper mean abs log error \\
\midrule\endfirsthead
\toprule
Configuration & A framing & Median cutoff & Lower-\allowbreak{}group items & Lower mean abs log error & Upper-\allowbreak{}group items & Upper mean abs log error \\
\midrule\endhead
\bottomrule\endfoot
codex & none & 0.0390 & 25 & 0.0266 & 24 & 0.4022 \\
codex & anchored & 0.0579 & 25 & 0.0425 & 25 & 0.3594 \\
Sonnet 5.5 & none & 0.0985 & 25 & 0.0291 & 25 & 0.5500 \\
Sonnet 5.5 & anchored & 0.1448 & 25 & 0.0588 & 25 & 0.4559 \\
Qwen3 & none & 0.0929 & 25 & 0.7175 & 25 & 0.8857 \\
Qwen3 & anchored & 0.1107 & 25 & 0.4215 & 25 & 0.9914 \\
gpt-\allowbreak{}oss-\allowbreak{}120b & none & 0.0729 & 25 & 0.1360 & 24 & 1.4604 \\
gpt-\allowbreak{}oss-\allowbreak{}120b & anchored & 0.0781 & 25 & 0.1118 & 25 & 0.3939 \\
Flash low & none & 0.0415 & 25 & 0.0276 & 25 & 0.3532 \\
Flash low & anchored & 0.0521 & 25 & 0.0139 & 25 & 0.3996 \\
Flash high & none & 0.0559 & 25 & 0.0283 & 25 & 0.3859 \\
Flash high & anchored & 0.0446 & 25 & 0.0353 & 25 & 0.3314 \\
\end{longtable}
\endgroup

% END A:759

% BEGIN A:774 text

\textbf{Table H10. D spread associations separated by person and event items}\par

% END A:774

% BEGIN A:776 text

P:\allowbreak{} 20 person items; W:\allowbreak{} 15 event pairs. Pooling these strata does not identify an association in each stratum.\par

% END A:776

% BEGIN A:778 table

\begingroup\scriptsize\setlength{\tabcolsep}{1.3mm}\setlength{\LTpre}{4pt}\setlength{\LTpost}{6pt}
\begin{longtable}{>{\raggedright\arraybackslash}p{29.764mm}>{\raggedright\arraybackslash}p{11.448mm}>{\raggedright\arraybackslash}p{11.448mm}>{\raggedright\arraybackslash}p{18.316mm}>{\raggedright\arraybackslash}p{48.081mm}>{\raggedright\arraybackslash}p{34.343mm}}
\toprule
Configuration & Type & Items & Clusters & Spearman rho [95\% CI] & Valid resamples \\
\midrule\endfirsthead
\toprule
Configuration & Type & Items & Clusters & Spearman rho [95\% CI] & Valid resamples \\
\midrule\endhead
\bottomrule\endfoot
codex & P & 20 & 20 & 0.346 [-\allowbreak{}0.148,\allowbreak{} 0.734] & 5000 \\
codex & W & 15 & 15 & 0.796 [0.462,\allowbreak{} 0.944] & 5000 \\
Sonnet 5.5 & P & 20 & 20 & 0.296 [-\allowbreak{}0.116,\allowbreak{} 0.657] & 5000 \\
Sonnet 5.5 & W & 15 & 15 & 0.582 [0.011,\allowbreak{} 0.905] & 5000 \\
Qwen3 & P & 20 & 20 & 0.570 [0.160,\allowbreak{} 0.837] & 5000 \\
Qwen3 & W & 15 & 15 & 0.679 [0.176,\allowbreak{} 0.922] & 5000 \\
gpt-\allowbreak{}oss-\allowbreak{}120b & P & 20 & 20 & 0.611 [0.208,\allowbreak{} 0.840] & 5000 \\
gpt-\allowbreak{}oss-\allowbreak{}120b & W & 15 & 15 & 0.654 [0.154,\allowbreak{} 0.924] & 5000 \\
Flash low & P & 20 & 20 & 0.862 [0.647,\allowbreak{} 0.941] & 5000 \\
Flash low & W & 15 & 15 & 0.864 [0.609,\allowbreak{} 0.955] & 5000 \\
Flash high & P & 20 & 20 & 0.549 [0.153,\allowbreak{} 0.820] & 5000 \\
Flash high & W & 15 & 15 & 0.754 [0.370,\allowbreak{} 0.908] & 5000 \\
\end{longtable}
\endgroup

% END A:778

% BEGIN A:793 text

\textbf{Table H11. Exploratory Flash-\allowbreak{}high execution-\allowbreak{}window partition}\par

% END A:793

% BEGIN A:795 text

F0 and F1 are non-\allowbreak{}randomized execution windows confounded with item schedule order and collection time. Differences are exploratory and do not identify a causal execution-\allowbreak{}window or model effect. Cluster D /\allowbreak{} w01 spans F0 (7 r0 calls) and F1 (3 r0 calls) across both response formats. There are zero cross-\allowbreak{}batch r0/\allowbreak{}r1 repeat pairs:\allowbreak{} every repeated unit has its r0 and r1 calls in the same batch. Full item/\allowbreak{}cluster partitions and the empty cross-\allowbreak{}batch repeat-\allowbreak{}pair list remain in high\_\allowbreak{}batch\_\allowbreak{}diagnostics\_\allowbreak{}exploratory.json; assignments to the other batch are not missing outcomes.\par

% END A:795

% BEGIN A:797 table

\begingroup\scriptsize\setlength{\tabcolsep}{1.3mm}\setlength{\LTpre}{4pt}\setlength{\LTpost}{6pt}
\begin{longtable}{>{\raggedright\arraybackslash}p{9.833mm}>{\raggedright\arraybackslash}p{9.833mm}>{\raggedright\arraybackslash}p{19.667mm}>{\raggedright\arraybackslash}p{45.233mm}>{\raggedright\arraybackslash}p{23.600mm}>{\raggedright\arraybackslash}p{45.233mm}}
\toprule
Batch & Round & Experiment & Assigned admitted units & True missing & Assigned to other batch \\
\midrule\endfirsthead
\toprule
Batch & Round & Experiment & Assigned admitted units & True missing & Assigned to other batch \\
\midrule\endhead
\bottomrule\endfoot
F0 & r0 & A & 430 & 0 & 653 \\
F0 & r0 & B & 1216 & 1 &  \\
F0 & r0 & C & 408 & 0 &  \\
F0 & r0 & D & 297 & 0 &  \\
F0 & r0 & E & 336 & 0 &  \\
F0 & r1 & A & 60 & 0 &  \\
F0 & r1 & B & 128 & 0 &  \\
F0 & r1 & C & 72 & 0 &  \\
F0 & r1 & D & 60 & 0 &  \\
F0 & r1 & E & 42 & 0 &  \\
F1 & r0 & A & 70 & 0 & 3049 \\
F1 & r0 & B & 224 & 0 &  \\
F1 & r0 & C & 72 & 0 &  \\
F1 & r0 & D & 53 & 0 &  \\
F1 & r0 & E & 112 & 0 &  \\
F1 & r1 & A & 20 & 0 &  \\
F1 & r1 & B & 64 & 0 &  \\
F1 & r1 & C & 24 & 0 &  \\
F1 & r1 & D & 0 & 0 &  \\
F1 & r1 & E & 14 & 0 &  \\
\end{longtable}
\endgroup

% END A:797

% BEGIN A:821 text

\clearpage\section*{Appendix I. Post-\allowbreak{}Cohort Stage 2 Reproduction Details and Preregistered Supplemental Suites (P1 and P2)}

% END A:821

% BEGIN A:823 text

\subsection*{I.1 Scope and execution accounting}

% END A:823

% BEGIN A:825 text

This appendix extends the unchanged Appendices A--H. The Stage 1 frozen cohort,\allowbreak{} exploratory Stage 2B/\allowbreak{}2C,\allowbreak{} and independently preregistered formal P1/\allowbreak{}P2 have distinct inferential status. Reproduce upstream collection only from its frozen code and sampling banks; the present verification script reads existing aggregates without API requests. All calls in this ledger are planned/\allowbreak{}admitted calls,\allowbreak{} excluding retries.\par

% END A:825

% BEGIN A:827 table

\begingroup\scriptsize\setlength{\tabcolsep}{1.3mm}\setlength{\LTpre}{4pt}\setlength{\LTpost}{6pt}
\begin{longtable}{>{\raggedright\arraybackslash}p{39.079mm}>{\raggedright\arraybackslash}p{53.733mm}>{\raggedright\arraybackslash}p{68.388mm}}
\toprule
Stage /\allowbreak{} accounting label & Planned/\allowbreak{}admitted evaluation units & Planned/\allowbreak{}admitted calls (excluding retries) \\
\midrule\endfirsthead
\toprule
Stage /\allowbreak{} accounting label & Planned/\allowbreak{}admitted evaluation units & Planned/\allowbreak{}admitted calls (excluding retries) \\
\midrule\endhead
\bottomrule\endfoot
Stage 1 Dev & 3708 & 3708 \\
Stage 1 Formal & 22212 & 22212 \\
Stage 2B & 384 & 384 \\
Stage 2C & 1400 & 1400 \\
P1 Block A & 1152 & 1152 \\
P1 Block B & 768 & 768 \\
P1 total & 1920 & 1920 \\
P2-\allowbreak{}A & 1280 & 1280 \\
P2-\allowbreak{}B & 768 & 1280 \\
P2 total & 2048 & 2560 \\
Entire study & 31672 & 32184 \\
P2 first-\allowbreak{}attempt parsed calls & 2556 & 99.84\% \\
\end{longtable}
\endgroup

% END A:827

% BEGIN A:842 text

\subsection*{I.2 Stage 2B/\allowbreak{}2C protocol and parsing}
The protocol and parsing evidence appears in \hyperref[sec:5.2]{Section 5.2}, Figure 4, Table 3 and Figures 5--6; source excerpts appear in \hyperref[sec:I.11]{Section I.11}.

% END A:842

% BEGIN A:950 text

\subsection*{I.3 P1 protocol,\allowbreak{} pooled effects and secondary findings}
The protocol and pooled effects appear in \hyperref[sec:5.3.1]{Section 5.3.1} and Table 4; see also Sections I.6, I.8 and I.10.

% END A:950

% BEGIN A:994 text

P1 ITT assigns Bayesian failures the prespecified 50\% fallback; framing failures receive the polarity-\allowbreak{}gap upper bound,\allowbreak{} binary failures 100 pp. Average repeats within item,\allowbreak{} then polarity,\allowbreak{} scenario cluster and model with equal weights. PP uses common successful repeats and common clusters across cells/\allowbreak{}models. Confirmatory success requires ITT and PP direction,\allowbreak{} SESOI,\allowbreak{} adjusted p and simultaneous interval criteria; operational matching additionally requires Manski robustness. Model-\allowbreak{}specific analyses are secondary. Zero final missingness makes the missingness alternatives coincide in this dataset; it does not establish general robustness to future missingness.\par

% END A:994

% BEGIN A:996 text

\subsection*{I.4 P2 protocol and all pooled results}
The protocol and pooled results appear in \hyperref[sec:5.3.2]{Section 5.3.2} and Table 5; see also Sections I.5, I.7 and I.9.

% END A:996

% BEGIN A:1047 text

\subsection*{I.5 Frozen fingerprints and exact execution parameters}

% END A:1047

% BEGIN A:1049 text

The tables below transcribe freeze records using repository-\allowbreak{}relative names. Each freeze record is historical provenance,\allowbreak{} not an authorization by this package. P1's historical stage field is stage1\_\allowbreak{}pending\_\allowbreak{}independent\_\allowbreak{}c\_\allowbreak{}audit and formal\_\allowbreak{}authorized=false; formal release evidence is maintained separately upstream. Both suites use R=2,\allowbreak{} B=10000,\allowbreak{} seed=20261110,\allowbreak{} max\_\allowbreak{}tokens=2048,\allowbreak{} max\_\allowbreak{}attempts=4,\allowbreak{} temperature=0.7,\allowbreak{} top\_\allowbreak{}p=0.95; Python 3.11.2 and NumPy 1.24.2 are recorded. P1 Holm family sizes are 3/\allowbreak{}3/\allowbreak{}6/\allowbreak{}6/\allowbreak{}6; P2-\allowbreak{}A uses 3 per endpoint,\allowbreak{} P2-\allowbreak{}B uses 6 over two endpoints. All bootstrap quantiles use linear interpolation. Raw data archives remain upstream under their Drive manifests.\par

% END A:1049

% BEGIN A:1052 text

\subsubsection*{factorial-\allowbreak{}pf-\allowbreak{}asc-\allowbreak{}20261110}

% END A:1052

% BEGIN A:1054 text

Freeze record SHA-\allowbreak{}256:\allowbreak{} \texttt{\seqsplit{c12f7269d30eb7bff88f39c7a4dd59a64e19e9c760ebb18244f516cad5c391bd}}.\par

% END A:1054

% BEGIN A:1056 table

\begingroup\scriptsize\setlength{\tabcolsep}{1.3mm}\setlength{\LTpre}{4pt}\setlength{\LTpost}{6pt}
\begin{longtable}{>{\raggedright\arraybackslash}p{22.372mm}>{\raggedright\arraybackslash}p{74.907mm}>{\raggedright\arraybackslash}p{63.921mm}}
\toprule
Manifest group & Repository-\allowbreak{}relative file & Frozen SHA-\allowbreak{}256 \\
\midrule\endfirsthead
\toprule
Manifest group & Repository-\allowbreak{}relative file & Frozen SHA-\allowbreak{}256 \\
\midrule\endhead
\bottomrule\endfoot
execution\_\allowbreak{}dependencies & connecting-\allowbreak{}dots/\allowbreak{}psych3/\allowbreak{}factorial-\allowbreak{}pf-\allowbreak{}asc-\allowbreak{}20261110/\allowbreak{}build\_\allowbreak{}banks.py & \seqsplit{414f57a83e6246f8694a75b7d76ed243ab480ad8c7b563bf31b2741285d1cf54} \\
execution\_\allowbreak{}dependencies & connecting-\allowbreak{}dots/\allowbreak{}psych3/\allowbreak{}factorial-\allowbreak{}pf-\allowbreak{}asc-\allowbreak{}20261110/\allowbreak{}run\_\allowbreak{}p1\_\allowbreak{}factorial.py & \seqsplit{0cb0baeafd46f5cf99f2feb2448e3573ad071b3864e0686900af4df64f46a3ae} \\
execution\_\allowbreak{}dependencies & connecting-\allowbreak{}dots/\allowbreak{}psych3/\allowbreak{}factorial-\allowbreak{}pf-\allowbreak{}asc-\allowbreak{}20261110/\allowbreak{}analyze\_\allowbreak{}p1\_\allowbreak{}factorial.py & \seqsplit{68663fc3a128332ea69f860010cd26d6717d7c2b9a6a46bcf87a23ba97504db9} \\
execution\_\allowbreak{}dependencies & connecting-\allowbreak{}dots/\allowbreak{}psych3/\allowbreak{}factorial-\allowbreak{}pf-\allowbreak{}asc-\allowbreak{}20261110/\allowbreak{}bank\_\allowbreak{}probe.json & \seqsplit{653ba498ee553fb8641fc74be5ca6367f658a3cce758065d0b21443dd8ee0501} \\
execution\_\allowbreak{}dependencies & connecting-\allowbreak{}dots/\allowbreak{}psych3/\allowbreak{}factorial-\allowbreak{}pf-\allowbreak{}asc-\allowbreak{}20261110/\allowbreak{}bank\_\allowbreak{}pilot.json & \seqsplit{d55e61d2bd13b447ef393948da705ca61846c74a9be8fe311d4b6d9df5799ae4} \\
execution\_\allowbreak{}dependencies & connecting-\allowbreak{}dots/\allowbreak{}psych3/\allowbreak{}factorial-\allowbreak{}pf-\allowbreak{}asc-\allowbreak{}20261110/\allowbreak{}bank\_\allowbreak{}formal.json & \seqsplit{8446069dd0beacc7ac2d6b10bacc12a8275c3ec66ea4d04a72b2d0019fbe25ef} \\
\end{longtable}
\endgroup

% END A:1056

% BEGIN A:1079 text

\subsubsection*{p2-\allowbreak{}supplemental-\allowbreak{}20261110}

% END A:1079

% BEGIN A:1081 text

Freeze record SHA-\allowbreak{}256:\allowbreak{} \texttt{\seqsplit{30db675b3e9e99f030376051aeee10c34c44fdbecc5cf9566c3b246f7d2e6cad}}.\par

% END A:1081

% BEGIN A:1083 table

\begingroup\scriptsize\setlength{\tabcolsep}{1.3mm}\setlength{\LTpre}{4pt}\setlength{\LTpost}{6pt}
\begin{longtable}{>{\raggedright\arraybackslash}p{23.087mm}>{\raggedright\arraybackslash}p{72.148mm}>{\raggedright\arraybackslash}p{65.964mm}}
\toprule
Manifest group & Repository-\allowbreak{}relative file & Frozen SHA-\allowbreak{}256 \\
\midrule\endfirsthead
\toprule
Manifest group & Repository-\allowbreak{}relative file & Frozen SHA-\allowbreak{}256 \\
\midrule\endhead
\bottomrule\endfoot
execution\_\allowbreak{}dependencies & connecting-\allowbreak{}dots/\allowbreak{}psych3/\allowbreak{}p2-\allowbreak{}supplemental-\allowbreak{}20261110/\allowbreak{}build\_\allowbreak{}p2\_\allowbreak{}banks.py & \seqsplit{a1e57de8b7c3fc2aa120f585ff406d9d88de04a5e623f891076abccab03d3b14} \\
execution\_\allowbreak{}dependencies & connecting-\allowbreak{}dots/\allowbreak{}psych3/\allowbreak{}p2-\allowbreak{}supplemental-\allowbreak{}20261110/\allowbreak{}run\_\allowbreak{}p2\_\allowbreak{}supplemental.py & \seqsplit{682e0fe8272dba6c0ecc10d243b8610ac0d96f1576a5e6c0dbed1776a8fa7c0f} \\
execution\_\allowbreak{}dependencies & connecting-\allowbreak{}dots/\allowbreak{}psych3/\allowbreak{}p2-\allowbreak{}supplemental-\allowbreak{}20261110/\allowbreak{}analyze\_\allowbreak{}p2\_\allowbreak{}supplemental.py & \seqsplit{d8d9490b8cf99ae57e0ff215140c77ca790b3d82d7f03204c45370718fc60aee} \\
execution\_\allowbreak{}dependencies & connecting-\allowbreak{}dots/\allowbreak{}psych3/\allowbreak{}p2-\allowbreak{}supplemental-\allowbreak{}20261110/\allowbreak{}archive\_\allowbreak{}p2\_\allowbreak{}pilot.py & \seqsplit{be716489ffbad86da322bf9b0109e627834f52926e08d2863b81b7cec24c804c} \\
execution\_\allowbreak{}dependencies & connecting-\allowbreak{}dots/\allowbreak{}psych3/\allowbreak{}p2-\allowbreak{}supplemental-\allowbreak{}20261110/\allowbreak{}freeze\_\allowbreak{}p2\_\allowbreak{}stage1.py & \seqsplit{c536dbe84412c212e619e1a6294d31a7adfc011b8e83ebe76f745708879cd921} \\
execution\_\allowbreak{}dependencies & connecting-\allowbreak{}dots/\allowbreak{}psych3/\allowbreak{}p2-\allowbreak{}supplemental-\allowbreak{}20261110/\allowbreak{}finish\_\allowbreak{}p2\_\allowbreak{}stage1.py & \seqsplit{765ac62f87899fa5731876c69d9c4712120154658f06a9446c12df2455b67506} \\
\end{longtable}
\endgroup

% END A:1083

% BEGIN A:1102 text

\subsection*{I.6 P1 complete model-\allowbreak{}specific secondary effects}

% END A:1102

% BEGIN A:1104 text

Selected model rows appear once in Table 4 above; the remaining rows follow. Together these cover all 120 pooled/\allowbreak{}model effects with 95\% simultaneous CIs. Models are not random population draws.\par

% END A:1104

% BEGIN A:1106 table

\begingroup\scriptsize\setlength{\tabcolsep}{1.3mm}\setlength{\LTpre}{4pt}\setlength{\LTpost}{6pt}
\begin{longtable}{>{\raggedright\arraybackslash}p{47.200mm}>{\raggedright\arraybackslash}p{20.060mm}>{\raggedright\arraybackslash}p{20.060mm}>{\raggedright\arraybackslash}p{18.880mm}>{\raggedright\arraybackslash}p{35.872mm}>{\raggedright\arraybackslash}p{11.328mm}}
\toprule
Population /\allowbreak{} family & Endpoint & Contrast & Delta (pp) & 95\% simultaneous CI & p\_\allowbreak{}Holm \\
\midrule\endfirsthead
\toprule
Population /\allowbreak{} family & Endpoint & Contrast & Delta (pp) & 95\% simultaneous CI & p\_\allowbreak{}Holm \\
\midrule\endhead
\bottomrule\endfoot
A\_\allowbreak{}Bayesian; sf:\allowbreak{}qwen3-\allowbreak{}8b & MAE & Delta\_\allowbreak{}P & -\allowbreak{}2.71 & [-\allowbreak{}5.859,\allowbreak{} +0.530] & 0.1434 \\
A\_\allowbreak{}Bayesian; sf:\allowbreak{}qwen3-\allowbreak{}8b & MAE & Delta\_\allowbreak{}A & +0.23 & [-\allowbreak{}2.886,\allowbreak{} +3.137] & 1.0000 \\
A\_\allowbreak{}Bayesian; sf:\allowbreak{}qwen3-\allowbreak{}8b & MAE & interaction & -\allowbreak{}0.22 & [-\allowbreak{}4.406,\allowbreak{} +3.757] & 1.0000 \\
A\_\allowbreak{}Bayesian; aistudio:\allowbreak{}gemma-\allowbreak{}4-\allowbreak{}26b-\allowbreak{}a4b-\allowbreak{}it & MAE & Delta\_\allowbreak{}P & +0.28 & [-\allowbreak{}0.561,\allowbreak{} +1.081] & 0.4060 \\
A\_\allowbreak{}Bayesian; aistudio:\allowbreak{}gemma-\allowbreak{}4-\allowbreak{}26b-\allowbreak{}a4b-\allowbreak{}it & MAE & interaction & +1.08 & [-\allowbreak{}0.389,\allowbreak{} +2.867] & 0.2308 \\
A\_\allowbreak{}Bayesian; nv:\allowbreak{}nemotron-\allowbreak{}3.5-\allowbreak{}super-\allowbreak{}vl-\allowbreak{}120b & MAE & Delta\_\allowbreak{}P & -\allowbreak{}0.26 & [-\allowbreak{}1.500,\allowbreak{} +0.977] & 1.0000 \\
A\_\allowbreak{}Bayesian; nv:\allowbreak{}nemotron-\allowbreak{}3.5-\allowbreak{}super-\allowbreak{}vl-\allowbreak{}120b & MAE & Delta\_\allowbreak{}A & -\allowbreak{}0.23 & [-\allowbreak{}2.090,\allowbreak{} +1.780] & 1.0000 \\
A\_\allowbreak{}Bayesian; xf:\allowbreak{}spark-\allowbreak{}x2.6 & MAE & Delta\_\allowbreak{}P & +2.00 & [-\allowbreak{}1.171,\allowbreak{} +5.401] & 0.2756 \\
\end{longtable}
\endgroup

% END A:1106

% BEGIN A:1201 text

\subsection*{I.7 P2 complete per-\allowbreak{}model,\allowbreak{} leave-\allowbreak{}one-\allowbreak{}model-\allowbreak{}out and quality details}

% END A:1201

% BEGIN A:1203 text

The following numeric field paths retain the summary schema; per\_\allowbreak{}model and leave\_\allowbreak{}one\_\allowbreak{}model\_\allowbreak{}out rows are secondary. ci95\_\allowbreak{}raw denotes a marginal interval; ci95\_\allowbreak{}simultaneous denotes a simultaneous interval. Historical MDE\_\allowbreak{}pilot\_\allowbreak{}R1\_\allowbreak{}80 fields in Formal describe post-\allowbreak{}hoc precision,\allowbreak{} not newly prospective power.\par

% END A:1203

% BEGIN A:1206 text

\subsubsection*{P2 quality}

% END A:1206

% BEGIN A:1208 table

\begingroup\scriptsize\setlength{\tabcolsep}{1.3mm}\setlength{\LTpre}{4pt}\setlength{\LTpost}{6pt}
\begin{longtable}{>{\raggedright\arraybackslash}p{122.850mm}>{\raggedright\arraybackslash}p{40.950mm}}
\toprule
JSON field path & Exact value \\
\midrule\endfirsthead
\toprule
JSON field path & Exact value \\
\midrule\endhead
\bottomrule\endfoot
quality/\allowbreak{}p2a/\allowbreak{}sf:\allowbreak{}qwen3-\allowbreak{}8b/\allowbreak{}E0\_\allowbreak{}direct/\allowbreak{}planned\_\allowbreak{}calls & 80 \\
quality/\allowbreak{}p2a/\allowbreak{}sf:\allowbreak{}qwen3-\allowbreak{}8b/\allowbreak{}E0\_\allowbreak{}direct/\allowbreak{}first\_\allowbreak{}records & 80 \\
quality/\allowbreak{}p2a/\allowbreak{}sf:\allowbreak{}qwen3-\allowbreak{}8b/\allowbreak{}E0\_\allowbreak{}direct/\allowbreak{}first\_\allowbreak{}parse\_\allowbreak{}rate & 1.0 \\
quality/\allowbreak{}p2a/\allowbreak{}sf:\allowbreak{}qwen3-\allowbreak{}8b/\allowbreak{}E0\_\allowbreak{}direct/\allowbreak{}final\_\allowbreak{}missing\_\allowbreak{}rate & 0.0 \\
quality/\allowbreak{}p2a/\allowbreak{}sf:\allowbreak{}qwen3-\allowbreak{}8b/\allowbreak{}E0\_\allowbreak{}direct/\allowbreak{}unit\_\allowbreak{}missing\_\allowbreak{}rate & 0.0 \\
quality/\allowbreak{}p2a/\allowbreak{}sf:\allowbreak{}qwen3-\allowbreak{}8b/\allowbreak{}E0\_\allowbreak{}direct/\allowbreak{}truncation\_\allowbreak{}rate & 0.0 \\
quality/\allowbreak{}p2a/\allowbreak{}sf:\allowbreak{}qwen3-\allowbreak{}8b/\allowbreak{}E0\_\allowbreak{}direct/\allowbreak{}unknown\_\allowbreak{}finish\_\allowbreak{}rate & 0.0 \\
quality/\allowbreak{}p2a/\allowbreak{}sf:\allowbreak{}qwen3-\allowbreak{}8b/\allowbreak{}E0\_\allowbreak{}direct/\allowbreak{}completion\_\allowbreak{}tokens & 11.35 \\
\end{longtable}
\endgroup

% END A:1208

% BEGIN A:1547 text

\subsubsection*{P2 replicate\_\allowbreak{}variance}

% END A:1547

% BEGIN A:1549 table

\begingroup\scriptsize\setlength{\tabcolsep}{1.3mm}\setlength{\LTpre}{4pt}\setlength{\LTpost}{6pt}
\begin{longtable}{>{\raggedright\arraybackslash}p{122.850mm}>{\raggedright\arraybackslash}p{40.950mm}}
\toprule
JSON field path & Exact value \\
\midrule\endfirsthead
\toprule
JSON field path & Exact value \\
\midrule\endhead
\bottomrule\endfoot
replicate\_\allowbreak{}variance/\allowbreak{}p2a/\allowbreak{}sf:\allowbreak{}qwen3-\allowbreak{}8b/\allowbreak{}E0\_\allowbreak{}direct/\allowbreak{}sigma2\_\allowbreak{}rep & 16.724666250000002 \\
replicate\_\allowbreak{}variance/\allowbreak{}p2a/\allowbreak{}sf:\allowbreak{}qwen3-\allowbreak{}8b/\allowbreak{}E0\_\allowbreak{}direct/\allowbreak{}sigma2\_\allowbreak{}cluster & 332.64643518092106 \\
replicate\_\allowbreak{}variance/\allowbreak{}p2a/\allowbreak{}sf:\allowbreak{}qwen3-\allowbreak{}8b/\allowbreak{}E0\_\allowbreak{}direct/\allowbreak{}rho\_\allowbreak{}rep & 0.9698223648556566 \\
replicate\_\allowbreak{}variance/\allowbreak{}p2a/\allowbreak{}sf:\allowbreak{}qwen3-\allowbreak{}8b/\allowbreak{}E0\_\allowbreak{}direct/\allowbreak{}complete\_\allowbreak{}repeat\_\allowbreak{}pairs & 40 \\
replicate\_\allowbreak{}variance/\allowbreak{}p2a/\allowbreak{}sf:\allowbreak{}qwen3-\allowbreak{}8b/\allowbreak{}E0\_\allowbreak{}direct/\allowbreak{}effective\_\allowbreak{}replications & 1.015319977924281 \\
replicate\_\allowbreak{}variance/\allowbreak{}p2a/\allowbreak{}sf:\allowbreak{}qwen3-\allowbreak{}8b/\allowbreak{}E1\_\allowbreak{}generic\_\allowbreak{}cot/\allowbreak{}sigma2\_\allowbreak{}rep & 20.672393749999998 \\
replicate\_\allowbreak{}variance/\allowbreak{}p2a/\allowbreak{}sf:\allowbreak{}qwen3-\allowbreak{}8b/\allowbreak{}E1\_\allowbreak{}generic\_\allowbreak{}cot/\allowbreak{}sigma2\_\allowbreak{}cluster & 361.2649753125 \\
replicate\_\allowbreak{}variance/\allowbreak{}p2a/\allowbreak{}sf:\allowbreak{}qwen3-\allowbreak{}8b/\allowbreak{}E1\_\allowbreak{}generic\_\allowbreak{}cot/\allowbreak{}rho\_\allowbreak{}rep & 0.963709511320886 \\
\end{longtable}
\endgroup

% END A:1549

% BEGIN A:1692 text

\subsubsection*{P2B per\_\allowbreak{}model}

% END A:1692

% BEGIN A:1694 table

\begingroup\scriptsize\setlength{\tabcolsep}{1.3mm}\setlength{\LTpre}{4pt}\setlength{\LTpost}{6pt}
\begin{longtable}{>{\raggedright\arraybackslash}p{119.289mm}>{\raggedright\arraybackslash}p{44.511mm}}
\toprule
JSON field path & Exact value \\
\midrule\endfirsthead
\toprule
JSON field path & Exact value \\
\midrule\endhead
\bottomrule\endfoot
P2B/\allowbreak{}per\_\allowbreak{}model/\allowbreak{}sf:\allowbreak{}qwen3-\allowbreak{}8b/\allowbreak{}MAE\_\allowbreak{}norm/\allowbreak{}0/\allowbreak{}estimate & -\allowbreak{}3.3774043371847458 \\
P2B/\allowbreak{}per\_\allowbreak{}model/\allowbreak{}sf:\allowbreak{}qwen3-\allowbreak{}8b/\allowbreak{}MAE\_\allowbreak{}norm/\allowbreak{}0/\allowbreak{}ci95\_\allowbreak{}raw/\allowbreak{}0 & -\allowbreak{}6.117637521792359 \\
P2B/\allowbreak{}per\_\allowbreak{}model/\allowbreak{}sf:\allowbreak{}qwen3-\allowbreak{}8b/\allowbreak{}MAE\_\allowbreak{}norm/\allowbreak{}0/\allowbreak{}ci95\_\allowbreak{}raw/\allowbreak{}1 & -\allowbreak{}0.9367613742873503 \\
P2B/\allowbreak{}per\_\allowbreak{}model/\allowbreak{}sf:\allowbreak{}qwen3-\allowbreak{}8b/\allowbreak{}MAE\_\allowbreak{}norm/\allowbreak{}0/\allowbreak{}p\_\allowbreak{}raw & 0.0037996200379962005 \\
P2B/\allowbreak{}per\_\allowbreak{}model/\allowbreak{}sf:\allowbreak{}qwen3-\allowbreak{}8b/\allowbreak{}MAE\_\allowbreak{}norm/\allowbreak{}0/\allowbreak{}ci95\_\allowbreak{}simultaneous/\allowbreak{}0 & -\allowbreak{}7.022991325317349 \\
P2B/\allowbreak{}per\_\allowbreak{}model/\allowbreak{}sf:\allowbreak{}qwen3-\allowbreak{}8b/\allowbreak{}MAE\_\allowbreak{}norm/\allowbreak{}0/\allowbreak{}ci95\_\allowbreak{}simultaneous/\allowbreak{}1 & -\allowbreak{}0.20062221404679056 \\
P2B/\allowbreak{}per\_\allowbreak{}model/\allowbreak{}sf:\allowbreak{}qwen3-\allowbreak{}8b/\allowbreak{}MAE\_\allowbreak{}norm/\allowbreak{}0/\allowbreak{}cluster\_\allowbreak{}sd & 5.488529696141225 \\
P2B/\allowbreak{}per\_\allowbreak{}model/\allowbreak{}sf:\allowbreak{}qwen3-\allowbreak{}8b/\allowbreak{}MAE\_\allowbreak{}norm/\allowbreak{}0/\allowbreak{}MDE\_\allowbreak{}pilot\_\allowbreak{}R1\_\allowbreak{}80 & 3.844145617190954 \\
\end{longtable}
\endgroup

% END A:1694

% BEGIN A:1937 text

\subsubsection*{P2B leave\_\allowbreak{}one\_\allowbreak{}model\_\allowbreak{}out}

% END A:1937

% BEGIN A:1939 table

\begingroup\scriptsize\setlength{\tabcolsep}{1.3mm}\setlength{\LTpre}{4pt}\setlength{\LTpost}{6pt}
\begin{longtable}{>{\raggedright\arraybackslash}p{122.850mm}>{\raggedright\arraybackslash}p{40.950mm}}
\toprule
JSON field path & Exact value \\
\midrule\endfirsthead
\toprule
JSON field path & Exact value \\
\midrule\endhead
\bottomrule\endfoot
P2B/\allowbreak{}leave\_\allowbreak{}one\_\allowbreak{}model\_\allowbreak{}out/\allowbreak{}aistudio:\allowbreak{}gemma-\allowbreak{}4-\allowbreak{}26b-\allowbreak{}a4b-\allowbreak{}it,\allowbreak{}nv:\allowbreak{}nemotron-\allowbreak{}3.5-\allowbreak{}super-\allowbreak{}vl-\allowbreak{}120b,\allowbreak{}xf:\allowbreak{}spark-\allowbreak{}x2.6/\allowbreak{}MAE\_\allowbreak{}norm/\allowbreak{}0/\allowbreak{}estimate & -\allowbreak{}0.5675800003865324 \\
P2B/\allowbreak{}leave\_\allowbreak{}one\_\allowbreak{}model\_\allowbreak{}out/\allowbreak{}aistudio:\allowbreak{}gemma-\allowbreak{}4-\allowbreak{}26b-\allowbreak{}a4b-\allowbreak{}it,\allowbreak{}nv:\allowbreak{}nemotron-\allowbreak{}3.5-\allowbreak{}super-\allowbreak{}vl-\allowbreak{}120b,\allowbreak{}xf:\allowbreak{}spark-\allowbreak{}x2.6/\allowbreak{}MAE\_\allowbreak{}norm/\allowbreak{}0/\allowbreak{}ci95\_\allowbreak{}raw/\allowbreak{}0 & -\allowbreak{}1.4999531066457532 \\
P2B/\allowbreak{}leave\_\allowbreak{}one\_\allowbreak{}model\_\allowbreak{}out/\allowbreak{}aistudio:\allowbreak{}gemma-\allowbreak{}4-\allowbreak{}26b-\allowbreak{}a4b-\allowbreak{}it,\allowbreak{}nv:\allowbreak{}nemotron-\allowbreak{}3.5-\allowbreak{}super-\allowbreak{}vl-\allowbreak{}120b,\allowbreak{}xf:\allowbreak{}spark-\allowbreak{}x2.6/\allowbreak{}MAE\_\allowbreak{}norm/\allowbreak{}0/\allowbreak{}ci95\_\allowbreak{}raw/\allowbreak{}1 & 0.3581610757948663 \\
P2B/\allowbreak{}leave\_\allowbreak{}one\_\allowbreak{}model\_\allowbreak{}out/\allowbreak{}aistudio:\allowbreak{}gemma-\allowbreak{}4-\allowbreak{}26b-\allowbreak{}a4b-\allowbreak{}it,\allowbreak{}nv:\allowbreak{}nemotron-\allowbreak{}3.5-\allowbreak{}super-\allowbreak{}vl-\allowbreak{}120b,\allowbreak{}xf:\allowbreak{}spark-\allowbreak{}x2.6/\allowbreak{}MAE\_\allowbreak{}norm/\allowbreak{}0/\allowbreak{}p\_\allowbreak{}raw & 0.23457654234576542 \\
\end{longtable}
\endgroup

% END A:1939

% BEGIN A:2182 text

\subsection*{I.8 P1 gates and diagnostic details}

% END A:2182

% BEGIN A:2184 table

\begingroup\scriptsize\setlength{\tabcolsep}{1.3mm}\setlength{\LTpre}{4pt}\setlength{\LTpost}{6pt}
\begin{longtable}{>{\raggedright\arraybackslash}p{118.800mm}>{\raggedright\arraybackslash}p{45.000mm}}
\toprule
JSON field path & Exact value \\
\midrule\endfirsthead
\toprule
JSON field path & Exact value \\
\midrule\endhead
\bottomrule\endfoot
gates/\allowbreak{}sf:\allowbreak{}qwen3-\allowbreak{}8b/\allowbreak{}A00/\allowbreak{}planned & 72 \\
gates/\allowbreak{}sf:\allowbreak{}qwen3-\allowbreak{}8b/\allowbreak{}A00/\allowbreak{}recorded & 72 \\
gates/\allowbreak{}sf:\allowbreak{}qwen3-\allowbreak{}8b/\allowbreak{}A00/\allowbreak{}first\_\allowbreak{}attempt\_\allowbreak{}parse\_\allowbreak{}rate & 1.0 \\
gates/\allowbreak{}sf:\allowbreak{}qwen3-\allowbreak{}8b/\allowbreak{}A00/\allowbreak{}length\_\allowbreak{}truncated\_\allowbreak{}rate & 0.0 \\
gates/\allowbreak{}sf:\allowbreak{}qwen3-\allowbreak{}8b/\allowbreak{}A00/\allowbreak{}unknown\_\allowbreak{}stop\_\allowbreak{}reason\_\allowbreak{}rate & 0.0 \\
gates/\allowbreak{}sf:\allowbreak{}qwen3-\allowbreak{}8b/\allowbreak{}A00/\allowbreak{}final\_\allowbreak{}missing\_\allowbreak{}rate & 0.0 \\
gates/\allowbreak{}sf:\allowbreak{}qwen3-\allowbreak{}8b/\allowbreak{}A00/\allowbreak{}all\_\allowbreak{}attempts & 72 \\
gates/\allowbreak{}sf:\allowbreak{}qwen3-\allowbreak{}8b/\allowbreak{}A00/\allowbreak{}pass\_\allowbreak{}gate & true \\
\end{longtable}
\endgroup

% END A:2184

% BEGIN A:4147 text

\subsection*{I.9 P2-\allowbreak{}A complete model-\allowbreak{}specific effects and sensitivity ranges}

% END A:4147

% BEGIN A:4149 table

\begingroup\scriptsize\setlength{\tabcolsep}{1.3mm}\setlength{\LTpre}{4pt}\setlength{\LTpost}{6pt}
\begin{longtable}{>{\raggedright\arraybackslash}p{122.850mm}>{\raggedright\arraybackslash}p{40.950mm}}
\toprule
JSON field path & Exact value \\
\midrule\endfirsthead
\toprule
JSON field path & Exact value \\
\midrule\endhead
\bottomrule\endfoot
stage2\_\allowbreak{}supplement/\allowbreak{}P2A\_\allowbreak{}per\_\allowbreak{}model/\allowbreak{}sf:\allowbreak{}qwen3-\allowbreak{}8b/\allowbreak{}bayes\_\allowbreak{}MAE/\allowbreak{}0/\allowbreak{}estimate & 3.9439984709480123 \\
stage2\_\allowbreak{}supplement/\allowbreak{}P2A\_\allowbreak{}per\_\allowbreak{}model/\allowbreak{}sf:\allowbreak{}qwen3-\allowbreak{}8b/\allowbreak{}bayes\_\allowbreak{}MAE/\allowbreak{}0/\allowbreak{}ci95\_\allowbreak{}raw/\allowbreak{}0 & 2.5897735091743126 \\
stage2\_\allowbreak{}supplement/\allowbreak{}P2A\_\allowbreak{}per\_\allowbreak{}model/\allowbreak{}sf:\allowbreak{}qwen3-\allowbreak{}8b/\allowbreak{}bayes\_\allowbreak{}MAE/\allowbreak{}0/\allowbreak{}ci95\_\allowbreak{}raw/\allowbreak{}1 & 5.292255351681956 \\
stage2\_\allowbreak{}supplement/\allowbreak{}P2A\_\allowbreak{}per\_\allowbreak{}model/\allowbreak{}sf:\allowbreak{}qwen3-\allowbreak{}8b/\allowbreak{}bayes\_\allowbreak{}MAE/\allowbreak{}0/\allowbreak{}p\_\allowbreak{}raw & 0.00019998000199980003 \\
stage2\_\allowbreak{}supplement/\allowbreak{}P2A\_\allowbreak{}per\_\allowbreak{}model/\allowbreak{}sf:\allowbreak{}qwen3-\allowbreak{}8b/\allowbreak{}bayes\_\allowbreak{}MAE/\allowbreak{}0/\allowbreak{}ci95\_\allowbreak{}simultaneous/\allowbreak{}0 & 2.2596875 \\
stage2\_\allowbreak{}supplement/\allowbreak{}P2A\_\allowbreak{}per\_\allowbreak{}model/\allowbreak{}sf:\allowbreak{}qwen3-\allowbreak{}8b/\allowbreak{}bayes\_\allowbreak{}MAE/\allowbreak{}0/\allowbreak{}ci95\_\allowbreak{}simultaneous/\allowbreak{}1 & 5.627435970948026 \\
stage2\_\allowbreak{}supplement/\allowbreak{}P2A\_\allowbreak{}per\_\allowbreak{}model/\allowbreak{}sf:\allowbreak{}qwen3-\allowbreak{}8b/\allowbreak{}bayes\_\allowbreak{}MAE/\allowbreak{}0/\allowbreak{}cluster\_\allowbreak{}sd & 2.4911759588749307 \\
\end{longtable}
\endgroup

% END A:4149

% BEGIN A:4917 text

\subsection*{I.10 Operational matching counts}

% END A:4917

% BEGIN A:4919 table

\begingroup\scriptsize\setlength{\tabcolsep}{1.3mm}\setlength{\LTpre}{4pt}\setlength{\LTpost}{6pt}
\begin{longtable}{>{\raggedright\arraybackslash}p{57.902mm}>{\raggedright\arraybackslash}p{20.140mm}>{\raggedright\arraybackslash}p{60.419mm}>{\raggedright\arraybackslash}p{20.140mm}}
\toprule
Measure & Cell & Matched /\allowbreak{} valid & Rate \\
\midrule\endfirsthead
\toprule
Measure & Cell & Matched /\allowbreak{} valid & Rate \\
\midrule\endhead
\bottomrule\endfoot
P1 operational matching & M100\_\allowbreak{}pct & 40/\allowbreak{}128 & 31.25\% \\
P1 operational matching & M13\_\allowbreak{}pct & 1/\allowbreak{}128 & 0.78\% \\
P1 operational matching & M15\_\allowbreak{}pct & 2/\allowbreak{}128 & 1.56\% \\
P1 operational matching & M100\_\allowbreak{}freq & 7/\allowbreak{}128 & 5.47\% \\
P1 operational matching & M13\_\allowbreak{}freq & 0/\allowbreak{}128 & 0.00\% \\
P1 operational matching & M15\_\allowbreak{}freq & 0/\allowbreak{}128 & 0.00\% \\
\end{longtable}
\endgroup

% END A:4919

% BEGIN A:4928 text

\subsection*{I.11 Exact Stage 2 prompt and parser specifications}\label{sec:I.11}

% END A:4928

% BEGIN A:4930 text

Stage 2B uses temperature 0.6,\allowbreak{} top\_\allowbreak{}p 0.95,\allowbreak{} max\_\allowbreak{}tokens 256,\allowbreak{} timeout 60 s,\allowbreak{} up to five attempts and default eight worker threads. Stage 2C uses temperature 0.6,\allowbreak{} top\_\allowbreak{}p 0.95,\allowbreak{} timeout 60 s,\allowbreak{} up to five attempts and default two worker threads per model. OpenAI-\allowbreak{}compatible outputs default to 512 tokens,\allowbreak{} with GPT-\allowbreak{}OSS configured to 2048 and reasoning\_\allowbreak{}effort low; Gemma native calls use maxOutputTokens 2048 and topK 20. These are source-\allowbreak{}code defaults,\allowbreak{} not reconstructed per-\allowbreak{}request token logs. GPT-\allowbreak{}OSS may fail over to Cloudflare after retry; routing and quantization were not logged per unit. Resuming skips recorded unit keys. Neither exploratory suite contributes new Stage 1 confirmatory tests.\par

% END A:4930

% BEGIN A:4932 text

The frozen prompt builders and parsers below specify label matching,\allowbreak{} decimal handling,\allowbreak{} candidate count,\allowbreak{} weight normalization,\allowbreak{} ordering tolerance,\allowbreak{} grid violations and final-\allowbreak{}estimate failures. They are documentary excerpts,\allowbreak{} not commands to call APIs. Stage 2B ascending/\allowbreak{}descending checks are strict in the indicated arms; Stage 2C's loose check uses a numerical tolerance and does not itself require all five candidates. The 329 loose ascending successes and 324 strict five-\allowbreak{}candidate successes therefore have different definitions. Invalid and noncompliant outputs retain their original status and are not repaired after error inspection.\par

% END A:4932

% BEGIN A:4935 text

\subsubsection*{stage2b-\allowbreak{}pilot-\allowbreak{}20261105/\allowbreak{}run\_\allowbreak{}stage2b\_\allowbreak{}suite.py}

% END A:4935

% BEGIN A:4937 text

Source SHA-\allowbreak{}256:\allowbreak{} \texttt{\seqsplit{d15b881a4b3f36f955bd08fb0ca653fc1431dddd25f4b06032b2bfc93356f1ed}}.\par

% END A:4937

% BEGIN A:4939 code

\begin{Verbatim}[fontsize=\scriptsize,breaklines=true,breakanywhere=true]
def build_block_a_prompt(story: str, arm: str) -> str:
    if arm == "direct":
        ask = (
            f"{STRICT_NO_THINK_NOTE}\n"
            "Output exactly one line in this format:\n"
            "Probability: NN%"
        )
    elif arm == "multi_free":
        ask = (
            f"{STRICT_NO_THINK_NOTE}\n"
            "Give 5 plausible numerical estimates (as percentages from 0% to 100%) for this posterior probability, "
            "and assign a nonnegative weight to each estimate so that the 5 weights sum to 1.00. "
            "The five numerical estimates must be distinct.\n"
            "Output exactly 5 lines in this format:\n"
            "Estimate 1: NN% | Weight: 0.XX\n"
            "Estimate 2: NN% | Weight: 0.XX\n"
            "Estimate 3: NN% | Weight: 0.XX\n"
            "Estimate 4: NN% | Weight: 0.XX\n"
            "Estimate 5: NN% | Weight: 0.XX"
        )
    elif arm == "multi_posterior_focus":
        ask = (
            f"{STRICT_NO_THINK_NOTE}\n"
            f"{POSTERIOR_FOCUS_NOTE}\n"
            "Give 5 plausible numerical estimates (as percentages from 0% to 100%) for this posterior probability, "
            "and assign a nonnegative weight to each estimate so that the 5 weights sum to 1.00. "
            "The five numerical estimates must be distinct.\n"
            "Output exactly 5 lines in this format:\n"
            "Estimate 1: NN% | Weight: 0.XX\n"
            "Estimate 2: NN% | Weight: 0.XX\n"
            "Estimate 3: NN% | Weight: 0.XX\n"
            "Estimate 4: NN% | Weight: 0.XX\n"
            "Estimate 5: NN% | Weight: 0.XX"
        )
    elif arm in ("1p5_asc", "1p5_desc"):
        order_rule = (
            "The 5 numerical estimates must be listed in strictly increasing order (Estimate 1 < Estimate 2 < Estimate 3 < Estimate 4 < Estimate 5)."
            if arm == "1p5_asc"
            else "The 5 numerical estimates must be listed in strictly decreasing order (Estimate 1 > Estimate 2 > Estimate 3 > Estimate 4 > Estimate 5)."
        )
        ask = (
            f"{STRICT_NO_THINK_NOTE}\n"
            "First, on Line 1, give your single best point estimate of this posterior probability.\n"
            "Then, on Lines 2 to 6, give 5 distinct plausible numerical estimates (from 0% to 100%) and assign a nonnegative weight to each so the 5 weights sum to 1.00. "
            f"{order_rule}\n"
            "Output exactly 6 lines in this format:\n"
            "Probability: NN%\n"
            "Estimate 1: NN% | Weight: 0.XX\n"
            "Estimate 2: NN% | Weight: 0.XX\n"
            "Estimate 3: NN% | Weight: 0.XX\n"
            "Estimate 4: NN% | Weight: 0.XX\n"
            "Estimate 5: NN% | Weight: 0.XX"
        )
    elif arm in ("5p1_asc", "5p1_desc"):
        order_rule = (
            "The 5 numerical estimates must be listed in strictly increasing order (Estimate 1 < Estimate 2 < Estimate 3 < Estimate 4 < Estimate 5)."
            if arm == "5p1_asc"
            else "The 5 numerical estimates must be listed in strictly decreasing order (Estimate 1 > Estimate 2 > Estimate 3 > Estimate 4 > Estimate 5)."
        )
        ask = (
            f"{STRICT_NO_THINK_NOTE}\n"
            "First, on Lines 1 to 5, give 5 distinct plausible numerical estimates (from 0% to 100%) for this posterior probability and assign a nonnegative weight to each so the 5 weights sum to 1.00. "
            f"{order_rule}\n"
            "Then, on Line 6, give your final synthesized single point estimate of this posterior probability.\n"
            "Output exactly 6 lines in this format:\n"
            "Estimate 1: NN% | Weight: 0.XX\n"
            "Estimate 2: NN% | Weight: 0.XX\n"
            "Estimate 3: NN% | Weight: 0.XX\n"
            "Estimate 4: NN% | Weight: 0.XX\n"
            "Estimate 5: NN% | Weight: 0.XX\n"
            "Final Estimate: NN%"
        )
    else:
        raise ValueError(f"Unknown Block A arm: {arm}")
    return f"{story}\n\n{ask}"


def build_scale_prompt(story: str, scale_M: int, fmt: str, target_desc: str) -> str:
    header = (
        f"{story}\n\n"
        f"Estimate the probability that {target_desc}.\n"
        f"Report the probability on a linear scale from 0 to {scale_M}.\n"
        f"0 means probability 0, and {scale_M} means probability 1.\n"
        f"A value x represents probability x/{scale_M}.\n"
        "Decimals are allowed.\n"
        f"{STRICT_NO_THINK_NOTE}"
    )
    if fmt == "direct":
        body = (
            "Output exactly one line:\n"
            f"estimate: <number between 0 and {scale_M}>"
        )
    elif fmt == "multi":
        body = (
            "Give five distinct plausible estimates of the same probability.\n"
            f"Use the 0-to-{scale_M} scale for every estimate.\n"
            "Give a nonnegative weight for each estimate; the weights must sum to 1.\n"
            "The weights express relative plausibility across these estimates.\n"
            "Output exactly five lines:\n"
            "candidate: <number>, weight: <number>"
        )
    elif fmt == "multi_repeat":
        body = (
            "Give five plausible estimates of the same probability (estimates may repeat; you may reuse the same number across multiple lines).\n"
            f"Use the 0-to-{scale_M} scale for every estimate.\n"
            "Give a nonnegative weight for each estimate; the weights must sum to 1.\n"
            "The weights express relative plausibility across these estimates.\n"
            "Output exactly five lines:\n"
            "candidate: <number>, weight: <number>"
        )
    else:
        raise ValueError(fmt)
    return f"{header}\n{body}"


def build_grid7_prompt(story: str, grid_label: str, fmt: str, target_desc: str) -> str:
    if grid_label == "int_0_7":
        grid_desc = (
            "Report the probability on an 8-point integer grid from 0 to 7: {0, 1, 2, 3, 4, 5, 6, 7}.\n"
            "0 means probability 0, and 7 means probability 1 (each integer k represents probability k/7).\n"
            "Only integers in {0, 1, 2, 3, 4, 5, 6, 7} are allowed (no decimals)."
        )
    elif grid_label == "pct_grid_7":
        grid_desc = (
            "Report the probability by choosing only from this exact 8-point percentage grid:\n"
            "{0.00%, 14.29%, 28.57%, 42.86%, 57.14%, 71.43%, 85.71%, 100.00%}.\n"
            "Only values from these 8 grid points are allowed."
        )
    else:
        raise ValueError(grid_label)

    if fmt == "direct":
        body = "Output exactly one line:\nestimate: <grid value>"
    elif fmt == "multi":
        body = (
            "Give five distinct plausible estimates of the same probability chosen from the allowed 8-point grid.\n"
            "Give a nonnegative weight for each estimate; the weights must sum to 1.\n"
            "Output exactly five lines:\n"
            "candidate: <grid value>, weight: <number>"
        )
    elif fmt == "multi_repeat":
        body = (
            "Give five plausible estimates of the same probability chosen from the allowed 8-point grid (grid values may repeat; you may reuse the same grid value across multiple lines).\n"
            "Give a nonnegative weight for each estimate; the weights must sum to 1.\n"
            "Output exactly five lines:\n"
            "candidate: <grid value>, weight: <number>"
        )
    else:
        raise ValueError(fmt)
    return f"{story}\n\nEstimate the probability that {target_desc}.\n{grid_desc}\n{STRICT_NO_THINK_NOTE}\n{body}"


\end{Verbatim}

% END A:4939

% BEGIN A:5094 text

\subsubsection*{stage2b-\allowbreak{}pilot-\allowbreak{}20261105/\allowbreak{}run\_\allowbreak{}stage2b\_\allowbreak{}suite.py}

% END A:5094

% BEGIN A:5096 text

Source SHA-\allowbreak{}256:\allowbreak{} \texttt{\seqsplit{d15b881a4b3f36f955bd08fb0ca653fc1431dddd25f4b06032b2bfc93356f1ed}}.\par

% END A:5096

% BEGIN A:5098 code

\begin{Verbatim}[fontsize=\scriptsize,breaklines=true,breakanywhere=true]
def parse_2b_response(text: str, unit: dict) -> dict:
    # Strip any <think>...</think> block if present
    cleaned = re.sub(r"<think>.*?</think>", "", text or "", flags=re.DOTALL).strip()
    lines = [ln.strip() for ln in cleaned.splitlines() if ln.strip()]
    block = unit["block"]
    scale_M = float(unit["scale_M"])

    if block == "A":
        arm = unit["arm"]
        if arm == "direct":
            for ln in reversed(lines):
                m = re.search(r"(?:Probability|Estimate)\s*:\s*([-+]?\d+(?:\.\d+)?)\s*%", ln, re.I)
                if m:
                    val = float(m.group(1))
                    if 0.0 <= val <= 100.0:
                        return {"ok": True, "value_pct": val}
            return {"ok": False, "error": "direct_parse_failed"}

        # Extract candidate lines and optional point lines
        cands = []
        wts = []
        first_point = None
        final_point = None
        for idx, ln in enumerate(lines):
            m_cand = re.search(
                r"Estimate\s*\d+\s*:\s*([-+]?\d+(?:\.\d+)?)\s*%?\s*\|\s*Weight\s*:\s*([-+]?\d+(?:\.\d+)?)",
                ln,
                re.I,
            )
            if m_cand:
                cands.append(float(m_cand.group(1)))
                wts.append(float(m_cand.group(2)))
                continue
            m_fin = re.search(r"Final\s+Estimate\s*:\s*([-+]?\d+(?:\.\d+)?)\s*%?", ln, re.I)
            if m_fin:
                final_point = float(m_fin.group(1))
                continue
            m_pt = re.search(r"Probability\s*:\s*([-+]?\d+(?:\.\d+)?)\s*%?", ln, re.I)
            if m_pt and len(cands) == 0 and first_point is None:
                first_point = float(m_pt.group(1))

        if len(cands) != 5 or len(wts) != 5:
            return {"ok": False, "error": f"expected_5_cands_got_{len(cands)}"}
        if any(c < 0.0 or c > 100.0 for c in cands) or any(w < 0.0 for w in wts):
            return {"ok": False, "error": "out_of_bounds"}
        w_sum = sum(wts)
        if w_sum <= 0:
            return {"ok": False, "error": "zero_weight_sum"}
        w_norm = [w / w_sum for w in wts]
        vs_pct = sum(c * w for c, w in zip(cands, w_norm))
        is_asc = all(cands[i] < cands[i + 1] for i in range(4))
        is_desc = all(cands[i] > cands[i + 1] for i in range(4))
        is_distinct = len(set(round(c, 4) for c in cands)) == 5

        order_compliant = True
        if arm.endswith("_asc") and not is_asc:
            order_compliant = False
        if arm.endswith("_desc") and not is_desc:
            order_compliant = False

        max_w = max(w_norm)
        top_slots = [i for i, w in enumerate(w_norm) if abs(w - max_w) < 1e-6]
        slot1_unique_max = (top_slots == [0])
        eff_k = 1.0 / sum(w * w for w in w_norm)

        res = {
            "ok": True,
            "value_pct": vs_pct,
            "candidates_pct": cands,
            "weights_raw": wts,
            "weights_norm": w_norm,
            "weight_sum_ok": abs(w_sum - 1.0) <= 0.05,
            "is_asc": is_asc,
            "is_desc": is_desc,
            "is_distinct": is_distinct,
            "order_compliant": order_compliant,
            "slot1_weight": w_norm[0],
            "slot1_unique_max": slot1_unique_max,
            "effective_k": eff_k,
            "d_first_pp": abs(vs_pct - cands[0]),
        }
        if arm.startswith("1p5_"):
            if first_point is None:
                return {"ok": False, "error": "missing_initial_point"}
            res["initial_point_pct"] = first_point
            res["d_point_pp"] = abs(vs_pct - first_point)
        elif arm.startswith("5p1_"):
            if final_point is None:
                return {"ok": False, "error": "missing_final_point"}
            res["final_point_pct"] = final_point
            res["d_point_pp"] = abs(vs_pct - final_point)
        return res

    # Blocks B, C, D use `estimate: <val>` or `candidate: <val>, weight: <val>`
    fmt = unit["fmt"]
    if fmt == "direct":
        for ln in reversed(lines):
            m = re.search(r"(?:estimate|probability)\s*:\s*([-+]?\d+(?:\.\d+)?)", ln, re.I)
            if m:
                raw_x = float(m.group(1))
                break
        else:
            # Fallback if model outputs just a single number line
            if len(lines) == 1 and _NUM_RE.fullmatch(lines[0].replace("%", "").strip()):
                raw_x = float(lines[0].replace("%", "").strip())
            else:
                return {"ok": False, "error": "scale_direct_parse_failed"}

        if block == "D":
            grid_label = unit["grid_label"]
            if grid_label == "int_0_7":
                k_round = round(raw_x)
                grid_compliant = (abs(raw_x - k_round) <= 1e-4 and 0 <= k_round <= 7)
                if not (0.0 <= raw_x <= 7.0):
                    return {"ok": False, "error": "out_of_bounds_0_7"}
                val_pct = (k_round * 100.0 / 7.0) if grid_compliant else (raw_x * 100.0 / 7.0)
                return {"ok": True, "raw_x": raw_x, "value_pct": val_pct, "grid_compliant": grid_compliant}
            else:
                if not (0.0 <= raw_x <= 100.0):
                    return {"ok": False, "error": "out_of_bounds_pct"}
                k_nearest = min(range(8), key=lambda k: abs(raw_x - (k * 100.0 / 7.0)))
                dist_grid = abs(raw_x - (k_nearest * 100.0 / 7.0))
                grid_compliant = dist_grid <= 0.10
                val_pct = (k_nearest * 100.0 / 7.0) if grid_compliant else raw_x
                return {"ok": True, "raw_x": raw_x, "value_pct": val_pct, "grid_compliant": grid_compliant}

        # Block B or C direct
        out_of_range = not (0.0 <= raw_x <= scale_M)
        if out_of_range:
            return {"ok": False, "error": f"out_of_bounds_0_{int(scale_M)}", "raw_x": raw_x}
        val_pct = (raw_x / scale_M) * 100.0
        return {"ok": True, "raw_x": raw_x, "value_pct": val_pct}

    # fmt == "multi" for Blocks B, C, D
    cands_raw = []
    wts_raw = []
    for ln in lines:
        m = re.search(
            r"(?:candidate|estimate)\s*\d*\s*:\s*([-+]?\d+(?:\.\d+)?)\s*%?\s*(?:,|\|)\s*weight\s*:\s*([-+]?\d+(?:\.\d+)?)",
            ln,
            re.I,
        )
        if m:
            cands_raw.append(float(m.group(1)))
            wts_raw.append(float(m.group(2)))
    if len(cands_raw) != 5:
        return {"ok": False, "error": f"scale_multi_expected_5_got_{len(cands_raw)}"}
    if any(w < 0.0 for w in wts_raw) or sum(wts_raw) <= 0.0:
        return {"ok": False, "error": "invalid_weights"}
    w_sum = sum(wts_raw)
    w_norm = [w / w_sum for w in wts_raw]

    if block == "D":
        grid_label = unit["grid_label"]
        cands_pct = []
        all_compliant = True
        for c in cands_raw:
            if grid_label == "int_0_7":
                if not (0.0 <= c <= 7.0):
                    return {"ok": False, "error": "cand_out_of_bounds_0_7"}
                kr = round(c)
                if abs(c - kr) <= 1e-4 and 0 <= kr <= 7:
                    cands_pct.append(kr * 100.0 / 7.0)
                else:
                    all_compliant = False
                    cands_pct.append(c * 100.0 / 7.0)
            else:
                if not (0.0 <= c <= 100.0):
                    return {"ok": False, "error": "cand_out_of_bounds_pct"}
                kn = min(range(8), key=lambda k: abs(c - (k * 100.0 / 7.0)))
                if abs(c - (kn * 100.0 / 7.0)) <= 0.10:
                    cands_pct.append(kn * 100.0 / 7.0)
                else:
                    all_compliant = False
                    cands_pct.append(c)
        val_pct = sum(cp * w for cp, w in zip(cands_pct, w_norm))
        return {
            "ok": True,
            "candidates_raw": cands_raw,
            "candidates_pct": cands_pct,
            "weights_norm": w_norm,
            "value_pct": val_pct,
            "grid_compliant": all_compliant,
        }

    # Block B or C multi
    if any(c < 0.0 or c > scale_M for c in cands_raw):
        return {"ok": False, "error": f"cand_out_of_bounds_0_{int(scale_M)}", "candidates_raw": cands_raw}
    cands_pct = [(c / scale_M) * 100.0 for c in cands_raw]
    val_pct = sum(cp * w for cp, w in zip(cands_pct, w_norm))
    return {
        "ok": True,
        "candidates_raw": cands_raw,
        "candidates_pct": cands_pct,
        "weights_norm": w_norm,
        "value_pct": val_pct,
        "slot1_weight": w_norm[0],
        "d_first_pp": abs(val_pct - cands_pct[0]),
    }


# --- API Runner ---
\end{Verbatim}

% END A:5098

% BEGIN A:5303 text

\subsubsection*{stage2c-\allowbreak{}crossover-\allowbreak{}20261105/\allowbreak{}run\_\allowbreak{}stage2c\_\allowbreak{}suite.py}

% END A:5303

% BEGIN A:5305 text

Source SHA-\allowbreak{}256:\allowbreak{} \texttt{\seqsplit{b6aa474db19c2d1d1cf55f42e693d57c53a73c507ca7b8fc71d4fd7aa274e75c}}.\par

% END A:5305

% BEGIN A:5307 code

\begin{Verbatim}[fontsize=\scriptsize,breaklines=true,breakanywhere=true]
def build_prompt(item: Dict[str, Any], scale_M: int, org: str) -> Tuple[str, str]:
    system_prompt = (
        "You are a precise quantitative judgment assistant. "
        "Follow the exact output format requested with no extra commentary or markdown."
    )
    scale_desc = (
        f"Report all estimates on a 0 to {scale_M} scale, where 0 means completely impossible (0%) "
        f"and {scale_M} means completely certain (100%). Decimals are allowed."
    )

    if org == "single":
        format_inst = (
            f"{scale_desc}\n"
            f"Respond with exactly one line in this format:\n"
            f"estimate: <number between 0 and {scale_M}>"
        )
    elif org == "multi_repeat":
        format_inst = (
            f"{scale_desc}\n"
            f"Provide 5 candidate estimates on the 0 to {scale_M} scale and their relative weights "
            f"(weights must be between 0 and 1 and sum to 1). Repeated candidate values are allowed.\n"
            f"Respond with exactly 5 lines in this format:\n"
            f"candidate: <number between 0 and {scale_M}>, weight: <number between 0 and 1>"
        )
    elif org == "1p5_asc_repeat":
        format_inst = (
            f"{scale_desc}\n"
            f"First provide your single best point estimate on line 1, and then provide 5 candidate estimates "
            f"sorted in non-decreasing order (c1 <= c2 <= c3 <= c4 <= c5, repeated values allowed) along with "
            f"their weights summing to 1.\n"
            f"Respond with exactly 6 lines in this format:\n"
            f"estimate: <number between 0 and {scale_M}>\n"
            f"candidate: <c1>, weight: <w1>\n"
            f"candidate: <c2>, weight: <w2>\n"
            f"candidate: <c3>, weight: <w3>\n"
            f"candidate: <c4>, weight: <w4>\n"
            f"candidate: <c5>, weight: <w5>"
        )
    elif org == "posterior_focus":
        format_inst = (
            f"{scale_desc}\n"
            f"{POSTERIOR_FOCUS_NOTE}\n"
            f"Respond with exactly 3 lines in this format:\n"
            f"prior_anchor: <initial baseline anchor on the 0 to {scale_M} scale>\n"
            f"evidence_shift: <signed shift on the -{scale_M} to +{scale_M} scale>\n"
            f"final_estimate: <integrated final estimate between 0 and {scale_M}>"
        )
    else:
        raise ValueError(f"Unknown organization: {org}")

    user_prompt = (
        f"Scenario:\n{item['narrative']}\n\n"
        f"Question:\n{item['question']}\n\n"
        f"Instructions:\n{format_inst}"
    )
    return system_prompt, user_prompt


def parse_response(text: str, scale_M: int, org: str) -> Dict[str, Any]:
    if not text or not text.strip():
        return {"parse_ok": False, "error": "empty_response"}

    cleaned = re.sub(r"<think>.*?</think>", "", text, flags=re.DOTALL).strip()
    if not cleaned:
        cleaned = text.strip()

    def to_pct(val_on_M: float) -> float:
        return (val_on_M / float(scale_M)) * 100.0

    def in_bounds(val_on_M: float) -> bool:
        return -1e-6 <= val_on_M <= float(scale_M) + 1e-6

    if org == "single":
        m = re.search(r"estimate\s*[:=]\s*([-+]?\d*\.?\d+)", cleaned, re.IGNORECASE)
        if not m:
            nums = re.findall(r"[-+]?\d*\.?\d+", cleaned)
            if not nums:
                return {"parse_ok": False, "error": "no_number_found"}
            raw_val = float(nums[-1])
        else:
            raw_val = float(m.group(1))
        if not in_bounds(raw_val):
            return {"parse_ok": False, "error": f"out_of_bounds({raw_val})"}
        pct = to_pct(raw_val)
        return {
            "parse_ok": True,
            "raw_scale_value": round(raw_val, 4),
            "value_pct": round(pct, 4),
            "initial_point_pct": round(pct, 4),
            "n_unique_candidates": 1,
        }

    elif org in ("multi_repeat", "1p5_asc_repeat"):
        initial_pct: Optional[float] = None
        if org == "1p5_asc_repeat":
            m_est = re.search(r"estimate\s*[:=]\s*([-+]?\d*\.?\d+)", cleaned, re.IGNORECASE)
            if m_est:
                est_val = float(m_est.group(1))
                if in_bounds(est_val):
                    initial_pct = round(to_pct(est_val), 4)

        cand_matches = re.findall(
            r"candidate\s*[:=]\s*([-+]?\d*\.?\d+)\s*[,;]\s*weight\s*[:=]\s*([-+]?\d*\.?\d+)",
            cleaned,
            re.IGNORECASE,
        )
        if len(cand_matches) < 1:
            return {"parse_ok": False, "error": f"too_few_candidates({len(cand_matches)})"}

        cands: List[float] = []
        weights: List[float] = []
        for c_str, w_str in cand_matches[:5]:
            c_val = float(c_str)
            w_val = float(w_str)
            if not in_bounds(c_val) or w_val < 0:
                return {"parse_ok": False, "error": f"invalid_cand_or_weight({c_val},{w_val})"}
            cands.append(c_val)
            weights.append(w_val)

        w_sum = sum(weights)
        if w_sum <= 1e-6:
            return {"parse_ok": False, "error": "zero_weight_sum"}
        norm_w = [w / w_sum for w in weights]
        weighted_M = sum(c * w for c, w in zip(cands, norm_w))
        pct = to_pct(weighted_M)
        if initial_pct is None:
            initial_pct = round(pct, 4)
        is_non_decreasing = all(cands[i] <= cands[i + 1] + 1e-6 for i in range(len(cands) - 1))
        return {
            "parse_ok": True,
            "raw_scale_value": round(weighted_M, 4),
            "value_pct": round(pct, 4),
            "initial_point_pct": initial_pct,
            "candidates_M": [round(c, 4) for c in cands],
            "weights_norm": [round(w, 4) for w in norm_w],
            "n_unique_candidates": len(set(round(c, 4) for c in cands)),
            "is_non_decreasing": is_non_decreasing,
        }

    elif org == "posterior_focus":
        m_final = re.search(r"final_estimate\s*[:=]\s*([-+]?\d*\.?\d+)", cleaned, re.IGNORECASE)
        m_prior = re.search(r"prior_anchor\s*[:=]\s*([-+]?\d*\.?\d+)", cleaned, re.IGNORECASE)
        m_shift = re.search(r"evidence_shift\s*[:=]\s*([-+]?\d*\.?\d+)", cleaned, re.IGNORECASE)
        if not m_final:
            return {"parse_ok": False, "error": "no_final_estimate_label"}
        final_val = float(m_final.group(1))

        if not in_bounds(final_val):
            return {"parse_ok": False, "error": f"out_of_bounds({final_val})"}

        prior_val = float(m_prior.group(1)) if m_prior else None
        shift_val = float(m_shift.group(1)) if m_shift else None
        pct = to_pct(final_val)
        prior_pct = round(to_pct(prior_val), 4) if (prior_val is not None and in_bounds(prior_val)) else round(pct, 4)
        return {
            "parse_ok": True,
            "raw_scale_value": round(final_val, 4),
            "value_pct": round(pct, 4),
            "initial_point_pct": prior_pct,
            "prior_anchor_M": round(prior_val, 4) if prior_val is not None else None,
            "evidence_shift_M": round(shift_val, 4) if shift_val is not None else None,
            "n_unique_candidates": 1,
        }

    return {"parse_ok": False, "error": "unknown_org"}


def check_input_number_echo(item: Dict[str, Any], value_pct: float, raw_scale_value: float, scale_M: int) -> bool:
    """Checks whether the reported estimate directly copies one of the raw percentages in the narrative."""
    input_pcts = [item["prior_pct"], item["hit_pct"], item["false_alarm_pct"]]
    for ip in input_pcts:
        if abs(value_pct - ip) <= 0.75:
            return True
        if scale_M != 100 and abs(raw_scale_value - ip) <= 0.25 and ip <= scale_M:
            return True
    return False


# ---------------------------------------------------------------------------
\end{Verbatim}

% END A:5307

\end{document}